\documentclass[3p]{elsarticle}

\usepackage{lineno,hyperref}

\usepackage[labelfont=bf]{caption}
\usepackage{caption}
\DeclareCaptionLabelSeparator{point}{.~}
\usepackage{amsmath}
\usepackage{algorithm}
\usepackage[noend]{algpseudocode}
\usepackage{booktabs}
\usepackage[skip=0.1\baselineskip]{caption}
\usepackage{subcaption}
\usepackage{multirow}
\usepackage{tabu}
\usepackage{xcolor}
\usepackage{graphicx}
\usepackage{nomencl}
\usepackage{multicol}
\usepackage{etoolbox}

\journal{arXiv.}

\begin{document}
\begin{frontmatter}

\title{Fault Diagnosis
using eXplainable AI: a Transfer Learning-based Approach for Rotating Machinery exploiting Augmented Synthetic Data}

\author[mymainaddress]{Lucas Costa Brito\corref{mycorrespondingauthor}}
\cortext[mycorrespondingauthor]{Corresponding author}
\ead{lucas.brito@ufu.br | brito.lcb@gmail.com}
\author[mysecondaryaddress]{Gian Antonio Susto}
\author[mythirdaryaddress]{Jorge Nei Brito}
\author[mymainaddress]{Marcus Antonio Viana Duarte}

\address[mymainaddress]{School of Mechanical Engineering, Federal University of Uberlândia, Av. João N. Ávila, 2121, Uberlândia, Brazil}
\address[mysecondaryaddress]{Department of Information Engineering, University of Padova, Via Gradenigo 6/B, 35131, Padova, Italy}
\address[mythirdaryaddress]{Department of Mechanical Engineering, Federal University of São João del Rei, P.Orlando, 170, São João del Rei, Brazil}

\begin{abstract}
Due to the growing interest for increasing productivity and cost reduction in industrial environment, new techniques for monitoring rotating machinery are emerging. Artificial Intelligence (AI) is one of the approaches that has been proposed to analyze the collected data (e.g., vibration signals) providing a diagnosis of the asset's operating condition. It is known that models trained with labeled data (supervised) achieve excellent results, but two main problems make their application in production processes difficult: (i) impossibility or long time to obtain a sample of all operational conditions (since faults seldom happen) and (ii) high cost of experts to label all acquired data. Another limitating factor for the applicability of AI approaches in this context is the lack of interpretability of the models (black-boxes), which reduces the confidence of the diagnosis and trust/adoption from users. To overcome these problems, a new generic and interpretable approach for classifying faults in rotating machinery based on transfer learning from augmented synthetic data to real rotating machinery is here proposed, namelly FaultD-XAI (Fault Diagnosis using eXplainable AI). To provide scalability using transfer learning, synthetic vibration signals are created mimicking the characteristic behavior of failures in operation. The application of Gradient-weighted Class Activation Mapping (Grad-CAM) with 1D Convolutional Neural Network (1D CNN) allows the interpretation of results, supporting the user in decision making and increasing diagnostic confidence. The proposed approach not only obtained promising diagnostic performance, but was also able to learn characteristics used by experts to identify conditions in a source domain and apply them in another target domain. The experimental results obtained on three datasets containing different mechanical faults suggest the method offers a promising approach on exploiting transfer learning, synthetic data and explainable artificial intelligence for fault diagnosis. Lastly, to guarantee reproducibility and foster research in the field, the developed dataset is made publicly available.

\end{abstract}

\begin{keyword}
\texttt Data Augmentation \sep Explainable Artificial Intelligence \sep Fault Diagnosis \sep Rotating Machinery \sep Synthetic Data \sep Transfer Learning 
\end{keyword}
\end{frontmatter}

\section{Introduction}

Currently, with the great need to increase the amount of final product manufactured, the industry has been looking for ways to monitor its assets in order to avoid unexpected breaks that can directly impact production \cite{xai_lucas}. Due to the growing demand, searches for alternatives in the monitoring of rotating machinery have been commonplace, leading to a large amount of information and research being generated: from the definition of informative signals to the development of smart data processing techniques, from new sensors to new best practice in data acquisition and monitoring.

The increase in information generates a need for experts to achieve quick analysis and effective diagnosis. Thus, artificial intelligence models have been proposed to assist in the diagnosis and monitoring of assets.

\cite{Kumar} present a review of the main Machine Learning (ML) and Deep Learning (DL) techniques applied in the monitoring of induction motors, aiming to detect faults such as: broken bars, bearings failures, stator failures and eccentricity. \cite{LIU201833} also present a review on artificial intelligence for fault detection in rotating machinery, in which more than 100 cited references refer mostly to fault classification (fault diagnosis). More recently, \cite{LEI2020106587} presented a comprehensive review with more than 400 citations, focused on Artificial Intelligence (AI) applications for fault diagnosis, providing a historical overview, in addition to current developments and future prospects. Further details on the state-of-the-art in Artificial Intelligence (AI) for fault diagnosis in rotating machinery can be found in \cite{i10,i11,i12,i13,ix1}.

By analysing the aforementioned studies, it can be noted that, for the task of monitoring rotating machinery, AI models have high accuracy rates when trained in supervised condition, ie. in the presence of labeled data \cite{LEI2020106587} representing both normal and faulty conditions. Unfortunately, the supervised scenario  is difficult to be applied in real world industrial settings, for two main reasons \cite{BYang}: 

i) \emph{Faults rarely happen in real world}: by design, equipment are made to be ideally operating all the time in a normal condition, therefore not providing sample labels for faulty conditions. When a fault manifests itself, the MCP (Maintenance Planning and Control) team schedules as soon as possible the corrective maintenance for repair or replacement, which results in a low number of acquired signals of fault conditions. In most conventional AI models, unbalanced datasets are associated with many issues, like for example low capability of prediction the least represented class, as the model tends to learn more about the condition that is most presented to it;

ii) \emph{High cost to obtain and label data}: with the development of AI approaches in the industry, more complex and sophisticated models are being used, like DL approaches; this is due to the fact that DL techniques can, among other things, automatically extract features from raw signals and can achieve high performance. This reduces the expert's intervention in feature engineering, a fundamental process for using conventional machine learning models. However, for successful training, such models require great amount of data, which is time consuming to obtain or not always feasible. To label the acquired data, experts are typically required: although the vast majority of faults have their characteristic behaviors known, the analysis requires specific knowledge that is not always available within the team.

Fortunately, \emph{transfer learning} \cite{transferlearning1} approaches can overcome such weaknesses by applying the knowledge learned from one task (or multiple tasks) to new, related, ones. \cite{i27, LEI2020106587, transferlearning2} present some approaches developed with a focus on transfer learning, showing that transfer-learning theories can alleviated the lacking labeled samples and enhancing the applications of Intelligent Fault Diagnosis (IFD) in industry, in such works several approaches are considered, namely: feature-based approaches \cite{BYang}, generative adversarial network (GAN) based approaches \cite{8511076}, instance-based approaches \cite{i36}, and parameter-based approaches \cite{i37}. Despite the promising studies, the vast majority of approaches are based on obtaining at least the Source Domain (samples on which the models will be trained) with real and labeled signals, which does not fully solve the aforementioned problems.

In order to reduce the need for labeled real data, a few recent studies propose the use of synthetically generated data from numerical and physical models \cite{I.Simulation1, I.Simulation2, I.Simulation3, I.Simulation4,ci}. Among the studies, stands out the methodology presented in \cite{ci}; in the study, a simulation-driven machine learning for bearing classification is proposed where training signals were generated from high resolution simulations in Siemens LMS Imagine. Lab Amesim simulation software using the one-dimensional 3-DOF model \cite{12i}. As with the other works, despite the satisfactory results obtained, the approaches are limited in terms of adjustment of boundary conditions, element properties, definition of solvers, model complexity, and application to only one type of failure (in this case, bearing fault), which can limit large-scale industrial applications.

Similarly to the problem of obtaining a complete and labeled training set, the adoption of AI models in the industry comes up against its interpretability \cite{xaidl7}. Called Explainable Artificial Intelligence (XAI), the area has been showing great interest by researchers, and is identified as one of the solutions to bridge the gap between AI researches and engineering applications \cite{LEI2020106587}. Many models are considered as black boxes, that is, the user responsible for receiving the diagnosis does not know how the model reached the final conclusion. By not knowing exactly how or which features of the signal the model was based on for decision making, the reliability of the diagnosis is compromised, implying even the non-use of the model. To solve this problem, studies have been developed to explain why the model obtained the final classification. Among the techniques that provide visual explanation, the state-of-the-art technique in post-hoc methods is Gradient-weighted Class Activation Mapping (Grad-CAM). As faults in IFD using vibration are generally identified through a visual analysis of the signal in the frequency domain, it is interesting to provide a heatmap overlaid on the input signal, identifying the most relevant frequencies for the classification. Despite the wide application in other areas, at present, Grad-CAM algorithm is seldom used in AI models for fault diagnosis \cite{xai2, xai3, xai4, Saeki}. 

To overcome the problems presented, a new  interpretable approach for classifying faults in rotating machinery based on transfer learning from augmented synthetic data to real rotating machinery is proposed, namelly FaultD-XAI (Fault Diagnosis using eXplainable AI). The approach starts from the concept where, transfer learning models focus on storing knowledge gained while solving one problem and applying it to a different but related problem. However, instead of working on fine-tuning, or retraining the model, the approach focuses on synthetic creation of the source domain: in this way the trained model learns features, which allow the transfer knowledge through the shared features of the source and target domain.

In FaultD-XAI, we propose to train the AI model with synthetic signals. The signals are generated from the knowledge about different faults in rotating machinery combined with the original signal of the machine under analysis, avoiding the use of complex mechanical models. The original signal is used only as a reference to create the synthetic signals, not requiring samples of all possible failures, which solves the problem of lack of labeled data in training set and enables engineering applications. To increase the variability of the training dataset, and consequently the robustness of the AI model, signals are generated using data augmentation techniques. To avoid the need to perform featuring engineering, reduce the computational cost, and enable the model's interpretability through the use of Grad-CAM, 1D Convolutional Neural Network (1D CNN) is used. Vibration signals in frequency domain are used, bringing the analysis as close as possible to that performed by experts. Due to the importance of rotors, bearings and gears for rotating machinery and especially their respective faults, three datasets are used for validation of the methodology, each one being related to a respective component and its possible faults.

The use of synthetic data makes it possible to apply the AI model in different rotating machinery even without having real signals in all operating conditions. Such ability can be seen as a transfer learning approach, since the proposal can be used in any type of machine, just by modifying the reference signal. In addition, the model is trained with a dataset belonging to the source domain, which, in turn, presents a difference from the signals present in the target domain. The data augmentation of the data allows the model to be implemented quickly, since few collections are enough to generate the training data. Finally, the interpretability of the FaultD-XAI allows the user to have reliability and confirm the diagnosis, enabling its implementation, and even, for the person responsible for the development of the model, to verify the coherence of the learning during the training phase, allowing adjustments if necessary.

The main contributions of the proposed approach are: i) a new transfer learning classification approach based on a synthetic dataset, without the need to have signals of real fault conditions; ii) possibility of interpreting the way in which the final result was obtained by the model, supporting decision making (a new contribution to the study of XAI in fault diagnosis); iii) a generic and simple way to generate synthetic fault data for training, based on the knowledge available in vibration analysis, without the need for complex models; iv) possibility of generating varied training datasets of different sizes (data augmentation - targeting deep learning applications); v) new dataset, publicly available at the link: \textit{https://data.mendeley.com/datasets/zx8pfhdtnb/1}, to study failures such as: unbalance, misalignment and looseness; vi) possibility of application in different types of faults; vii) faster deployment of the model in production; viii) industrial application on real world datasets.

The remainder of this paper starts with a brief explanation about the 1D CNN and Grad-CAM. The proposed approach is presented in Section 3. Experimental procedure is shown in Section 4. Results and discussion are given in Section 5. Finally, Section 6 concludes this paper by drawing conclusions and discussing potential future works.

\section{Background}

\subsection{1D Convolutional Neural Network (1D CNN)} \label{sec:ad}

Convolutional Neural Networks (CNNs) have become the de facto standard for various Computer Vision and Machine Learning operations. CNNs are feed-forward Artificial Neural Networks (ANNs) with alternating convolutional and subsampling layers \cite{x3}. The high efficiency of CNNs and the ability to extract features from the raw signal without the need for feature extraction or engineering make them basic tools in many 2D applications such as images and videos. The fact of being able to extract features automatically from the raw signal, avoids a lot of analysis and signal processing that need to be done when working with other techniques, such as those shown in \cite{infomartics_brito}.

Recently, several research works involving vibration signals and deep 2D CNNs were carried out, where researchers adopted signal processing strategies to convert the original 1D signal to 2D, and thus be able to use the architecture \cite{x3-34,x3-35,x3-36,x3-40,x3-41}. Despite the good results obtained, such techniques need to perform a dimension modification and can considerably increase the computational cost, in addition to the one already present in 2D CNN, making the applications sometimes unfeasible.

To solve this problem, 1D CNNs were proposed \cite{x3-46, x3-48, x3-52, x3-53, x3-54}, and quickly became state-of-the-art in applications such as: biomedical data classification and early diagnosis, structural health monitoring, anomaly detection and identification in power electronics and electrical motor fault detection. In the area of fault detection in rotating machines, some studies involving 1D CNN are \cite{xaidl2, 2.1x3, 2.1x4, 2.1x5, 2.1x6, 2.1x7}.

Among the main advantages of using 1D CNNs, the following stand out: i) significant computational cost reduction; ii) more compact networks (with 1-2 hidden CNN layers and configurations with networks having less than 10K parameters \cite{x3}); iii) possibility of training on any CPU; iv) possibility of being used for real-time and low-cost applications due to its low computational cost.

Basically, the configuration of a 1D CNN consists of: i) Hidden CNN and Multilayer perceptron (MLP) layers/neurons; ii) Filter (kernel); iii) Subsampling factor; iv) Pooling and activation functions.

The input layer receives the raw 1D signal, while the output layer is an MLP layer with a number of neurons equal to the number of classes. A kernel function moves in one direction: it first performs a sequence of convolutions, the sum of which is passed through the activation function, followed by the sub-sampling operation. The features are extracted, and used by the MLP layer to perform the classification. For more details on how 1D CNN works, please refer to \cite{x3}.

\subsection{Gradient-weighted Class Activation Mapping (Grad-CAM)} \label{sec:ad}

Grad-CAM \cite{x4} is an improvement on traditional CAM \cite{2.2x1}. Studies show the use of the method in the monitoring of rotating machinery, as in \cite{xai3} that used 1D CNN and 1D LeNet-5\footnote{LeNet is classic CNN architecture \cite{lenet}} (I1DLeNet) to prove the efficiency of the proposed method SBDS (smart bearing diagnosis system), which uses a gradient diagnosis -weighted class activation mapping (Grad-CAM) - based convolutional neuro-fuzzy network (GC-CNFN). Proving that the method is not only capable of correctly classifying bearing faults, but also helping the user to understand the result.

\cite{2.2x3} used Grad-CAM and Acoustic Emission Signals to detect bearing faults, again providing the user interpretability regarding the frequencies used by the model to obtain the result. \cite{xai2} applied the methodology in the monitoring of bearings and validated it in a gear fault dataset, showing that it is possible to use the methodology to identify the most attentive part of the model in relation to each type of fault. \cite{xai4} used Grad-CAM and eigenvector-based class activation map (Eigen-CAM) to interpret the ResNet06 (a popular CNN architecture \cite{resnet} in 4 databases, 3 bearing and 1 gearbox dataset. In order to interpret the effectiveness of the method, Grad-CAM is applied to localize the regions in the input that contribute the most to the network’s prediction. The proposed method is validated by a motor bearing dataset and an industrial hydro turbine dataset \cite{2.2x6}.

Grad-CAM uses the gradients of any target concept flowing into the final convolutional layer to produce a coarse localization map highlighting the important regions in the image for predicting the concept \cite{x4}. The signal is forward propagated by the CNN part of the model and then processed to obtain the raw score for the category. Gradients are reset for all classes except the class under analysis, set to 1. The signal is then backpropagated to the rectified convolutional feature maps of interest, which is combined to generate the heatmap. The highlighted points (i.e., the highest gradient value) are the most relevant regions that the model uses to make the decision. For more details about Grad-CAM please refer to \cite{x4}.

\section{Proposed approach: FaultD-XAI}

\begin{figure*}[ht]
  \centering
  \includegraphics[scale=0.5]{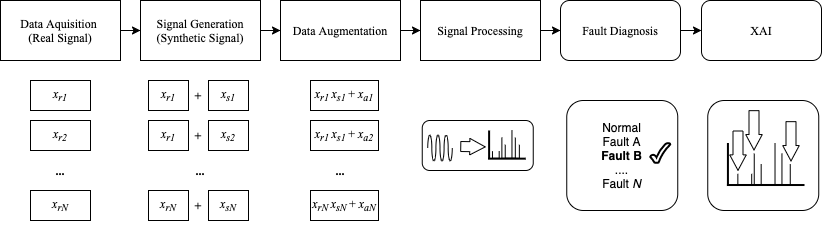}
  \caption{FaultD-XAI: General framework of the proposed methodology.}
  \label{fig:Cap_Classification}
\end{figure*}

The proposed approach is presented in Fig. \ref{fig:Cap_Classification}, and consists of six parts: i) Data Acquisition; ii) Signal Generation; iii) Data Augmentation; iv) Signal Processing; v) Fault Diagnosis; vi) Explainable Artificial Intelligence (XAI). Initially the real signals ($x_{r}$) are collected from the machine. To create different operating and fault conditions, synthetic signals ($x_{s}$) imitating fault characteristics are generated and then combined with real signals ($x_{r}$). To reduce the amount of original data needed for training the model, the signals are augmented ($x_{r}$$x_{s}$+$x_{a}$). The fourth step consists of the last phase of data preparation, where the signals are converted from the time domain to frequency domain. After pre-processing the data, the model is created and trained, enabling the diagnosis of the machine's condition (fault diagnosis). Finally, the model is explained, identifying the most relevant frequencies used for classification, supporting the user's decision making.

To the best of our knowledge, FaultD-XAI is the first approach that combined synthetic data with data augmentation to avoid the need of real data by exploiting XAI and transfer learning for fault diagnosis in rotating machinery.

\subsection{Data Acquisition}

Among the sensors used for monitoring rotating machinery, the vibration-based diagnostic method is the most popular and researched. The interest is justified by the fact that the vibration signals directly represent the dynamic behavior of the equipment \cite{feature01, feature02, feature03, feature04} and are a non-invasive technique \cite{xai_lucas}. Therefore, vibration signals were used. 

The real signal is fundamental to generate the synthetic dataset, enabling transfer learning between the source domain (synthetic signals) and target domain (real fault signals). Through the real signal of the machine under analysis, it is possible to generate a dataset that presents similar characteristics, favoring the learning of the model. It is worth mentioning that the real signal used is the current operating condition of the machine, not requiring a sample of each condition.

\subsection{Signal Generation}

Engineering applications involve a major limitation regarding the data available for training the AI model. In the vast majority of cases, we do not have real and labeled samples of all possible operating conditions, since faults seldom happen. In applications where data is available, it takes a lot of time and cost with experts to label and analyze all signals. A strategy to overcome this problem is to create the training dataset synthetically, since the vast majority of faults in rotating machines have known characteristic patterns.

Different approaches can be used to model faults in rotating machinery, such as: Digital Twin, where through equations of motion, numerical models, optimization and support software, the machine is created virtually, and through modifications in its conditions it is possible to verify the signal response, identifying a fault or a new operational condition.

Despite the available approaches, modeling a rotating machinery is not always an easy task or a low computational cost, which can make some industrial applications unfeasible. Therefore, we chose to model the faults, in a simple way, through the equation of a waveform, with a sine function, considering that the faults are mostly related to deterministic frequencies or characteristic behaviors in the signal.

In general, the most common faults found in the industry are related to bearing defects, classic problems such as unbalance, misalignment, looseness, as well as problems in gears, and for this reason they were chosen for analysis of the proposed approach. It is worth mentioning that the way the FaultD-XAI is proposed, other faults can be modeled and added, without changing its functioning (e.g., cavitation, electrical defects, oil whirl and oil whip etc). It is up to the specialist to determine the synthetic faults that will be created referring to the machine under analysis. It is also important to highlight that the methodology applies to stationary operating conditions, considering that the faults are created as a function of certain frequencies, which are mostly associated with the rotation frequency of the machine.

The representation of the oscillatory signal {\it{x(t)}} used to generate the faults (unbalance, misalignment, looseness and gear fault) is defined as:

\begin{flalign}
    & x(t) = A \times sin(2\pi \times f\times t + \theta\,) & \label{eq:sinal_oscil}
\end{flalign}

Where {\it{A}} indicates the peak amplitude of the signal, {\it{f}} frequency [Hz], {\it{t}} time vector [s] and $\theta$ phase [rad]. As accelerometers and no tachometer are used, the phase reference was not applied, although it can help in the identification of some faults.

The amplitudes were randomly generated so that the synthetic fault signal presented an approximate variation of at least 3 dB in the characteristic frequencies of the fault in relation to the normal signal. Since, according to \cite{10816-3:2009} and the experience of experts in vibration analysis, a variation of less than 3 dB does not represent a significant change in the behavior of the machine in terms of predictive maintenance or failure. The amplitudes are randomly generated to represent different stages of fault, and improve the robustness of the AI model (greater variability in the training group). It is worth mentioning that, depending on the fault, minor variations can be significant, and it is up to the specialist to determine when assembling the model.

The unbalance occurs predominantly at 1 x fr (Rotation Frequency), so a signal was generated with a frequency equal to the rotation frequency. It is worth mentioning again that the phase is not being taken into account to differentiate the faults, only the characteristic frequencies of the signal.

Misalignment presents high vibration values at 1x, 2x and 3x rpm, and can be angular, parallel or mixed (when the two are combined). In that case, Eq. \eqref{eq:sinal_oscil} was used three times, being the final signal the sum of each equation with a respective frequency (1x, 2x and 3x rpm).

The looseness caused by loose pillow block bolts, cracks in frame structure or in bearing pedestal, or by improper fit between components parts, causes harmonics and sub-harmonics of the rotation frequency. The signal will present multiples of the rotation frequency as: 1x, 2x, 3x, ... {\it{N}}x and sub-harmonics: 0.5x, 1.5x, 2.5x, ... {\it{N}}/2x, where {\it{N}} is the multiple of the rotation. As well as the misalignment, the final signal was composed of the sum of harmonics and sub-harmonics.

Gear faults are usually related to Gear Mesh Frequency (GMF) and its harmonics. GMF is the product of the number of teeth on the gear ($n\textsubscript{teeth}$) multiplied by the running speed of the gear ($g\textsubscript{fr}$), described as:

\begin{flalign}
    & GMF = n\textsubscript{teeth} \times g\textsubscript{fr} & \label{eq:gmf}
\end{flalign}

Defects such as wear, misalignment or eccentricity and backlash in gears may not only cause a significant increase in 1x GMF. In many cases, there is an increase in the amplitude of its harmonics as 2x GMF and 3x GMF. Such defects not only cause variation in the GMF and its harmonics, but in general, also present sidebands of the rotating speed of the defective gear. Therefore, the faults were simulated using Eq.\eqref{eq:gmf} where, the frequency for each  signal is given by GMF and its harmonics (1x, 2x, 3x GMF) and by the sidebands (GMF +/- rotating speed of the defective gear). The final signal was composed by the sum of each signal generated.

The faults in rolling element bearings are related to their components: inner race, balls or rollers, cage and outer race. Deterioration will cause characteristic frequencies in the signal allowing fault identification. The four frequencies are: i) BPFO (Ball Pass Frequency Outer); ii) BPFI (Ball Pass Frequency Inner); BSF (Ball Spin Frequency); iv) FTF (Fundamental Train Frequency). The calculation basically takes into account: the number of elements, internal and external diameter, contact angle, and its equation can be easily found in vibration analysis software, supplier catalogs and existing literature.

High frequency resonances between the bearing and the response transducer are excited when the rolling elements strike a local fault on the outer or inner race, or a fault on a rolling element strikes the outer or inner race \cite{x1}. The impact frequency depends directly on the machine components and their natural frequencies, however, as the interest is to evaluate the characteristic fault frequencies, this value can be arbitrarily defined in the signal modeling. The impact signal can be modelled using Eq.\eqref{eq:sinal_oscil} where {\it{A}} indicates the peak amplitude of the signal, {\it{f}} impact frequency [Hz], {\it{t}} time vector of impact [s]. To make the impact periodic, it is convoluted with a Comb function, where the non-null values (periodic peaks) correspond to the fault frequencies of the elements. The Comb is a sum of time shifted Dirac Delta, that is, it is defined to be zero at alternate time points.

After creating the synthetic faults, the original signal is added to represent the dynamic behavior of the machine, allowing the transfer knowledge. Due to the types of faults under study, at the end of the process, each original signal will result in 7 signals, being: normal condition, outer race fault (BPFO), inner race fault (BPFI), unbalanced, misaligned, mechanical looseness and gear fault (due to their common usage, the terms BPFO and BPFI will also be used to identify the type of fault under analysis, that is, fault in the outer and inner races, respectively.). Although the inner race fault (BPFI) was not present in any of the cases studied, it was introduced to increase the complexity of the training and also validating the FaultD-XAI.

\subsection{Data Augmentation}

Data augmentation is a widely used practice in ML when it is desired to increase the relevance of the dataset under study, mainly for image classification, natural language understanding, semantic segmentation and also in fault diagnosis \cite{I.Simulation2, x2, Aug.68, Aug.69}. In addition to increasing the amount of data for training, it is possible to insert small variations that, do not change the general context of the sample (although they exist). Some popular techniques are: flip, rotation, scale, crop, translation, gaussian noise. This makes the model able to correctly classify the sample, even if it presents noise, translations, different size, lighting etc. Performing augmentation can prevent the model from learning irrelevant patterns and consequently improve overall performance.

Just as the image of an object can be obtained from different perspectives by varying the context, but not the object under analysis, the same can be observed in the vibration signal of a rotating machinery. In an industrial plant, machines are subject to process variations, human interference, disturbances from nearby machines that can introduce extra information into the signal. Therefore, a robust classification model should be irrelevant to such variations, focusing only on identifying the operating condition of the asset under analysis.

In addition to improving data variability, data augmentation contributes to increasing data in the training set, enabling the implementation of more complex and deep learning models.

For the proposed approach, each synthetic signal is augmented, randomly varying the method variables. The methods presented in \cite{x2} were used. It is worth mentioning that other methodologies can also be used. The proposed methods are:

i) Gaussian Noise: apply random noise to the raw signal, as follows:

\begin{flalign}
    & \bar{x} = x + \alpha \textsubscript{gauss} \times G, \, G \, \sim \, N(0,1) & \label{eq:gauss_noise}
\end{flalign}

where $\alpha \textsubscript{gauss} >$ 0 is the Gaussian noise coefficient, $\bar{x}$ is the augmented data, x is the raw signal and G the gaussian noise.

ii) Masking Noise: an alternative way to apply random noise, where a fraction $\alpha \textsubscript{mask}$ of the elements in x are randomly selected and set to 0.

iii) Signal Translation: the signal is randomly shifted forward and backward, with the data gap resulting from the translation complete with zeros, maintaining the original size of the signal.

iv) Amplitude Shifting: Moderate variations in signal amplitude are common, and do not significantly alter the machine's operating state. Therefore, performing modifications of this nature tends to enrich the dataset. Amplitude shifting is implemented through a scaling factor ($\alpha \textsubscript{scal}$):
\begin{flalign}
    & \bar{x} = \alpha \textsubscript{scal} \times x & \label{eq:ampl_shift}
\end{flalign}

v) Time Stretching: The signal is stretched along the time axis by a stretching factor ($\alpha \textsubscript{stre}>$ 0). A sample with $N \textsubscript{aug}$ values in the range of [1, $\alpha \textsubscript{stre}$] if $\alpha \textsubscript{stre}\ge$1 or [ $\alpha \textsubscript{stre}$,1] is obtained. The sample is centered and stretching is applied. Points outside the dimensions are clipped out and gaps filled with zeros.

At the end of the data augmentation process, each signal will generate five new signals, and the process can be repeated {\it{n}} times. In summary, the entire process of creating synthetic signals and augmentation from one original signal will result in 35 new signals (7 synthetic signal * 5 augmented signal). The number of times ($q \textsubscript{aug}$) the procedure needs to be executed to obtain a total of {\it{n}} signals ($n \textsubscript{total}$) in the training group can be calculated as follows:

\begin{flalign}
    & q\textsubscript{aug} = \frac{n\textsubscript{total}}{7 \times 5 \times n \textsubscript{r}} & \label{eq:n_aug}
\end{flalign}

where, $n \textsubscript{total}$ is the desired amount of samples in the training set, 7 is the amount of synthetic signals, 5 is the amount of augmented signals and $n \textsubscript{r}$ the amount of real signals used. When the relationship does not result in an integer, the value is rounded up to the nearest integer, and the excess is randomly selected and discarded.

\subsection{Signal Processing}

DL models are able to extract features from raw signals and learn characteristic behaviors without the need for feature engineering as in classic ML models.

Although the raw signal (time domain) presents relevant information, due to the amount of frequencies present, the analysis often becomes complex. Therefore, it is common to perform analysis in the frequency domain. As one of the goals of the work is to provide interpretability of the result, the input signal was used in the frequency domain instead of the time domain. 

The signal was normalized using Z-score normalization, and finally, cut to reduce the number of points, and consequently the computational cost of the model. For the cut, the maximum frequencies at which the fault could manifest and their harmonics were analyzed, so according to the monitored asset, the size of the signal used may vary. As the dataset under analysis present constant speed, it is not necessary to perform order tracking or other method as a pre-processing step to remove the effect of shaft speed variations. 

\subsection{Fault Diagnosis}

CNN are present in several applications due to their main characteristics: i) possibility to learn features directly from raw signal in training; ii) are immune to small transformations/disturbances in the input data; iii) can adapt to different sizes; iv) excellent computational cost compared to conventional fully-connected Multi-Layer Perceptrons (MLP) networks (CNN neurons are sparsely-connected with tied weights \cite{x3}).

Since AlexNet's proposal \cite{x3-26}, when it achieved 16.4\% error rate in the ImageNet benchmark dataset (this was about 10\% lower than the second top method), several deep 2D CNNs have been used. However, there are certain drawbacks and limitations of using such deep CNNs: i) high computational cost and special hardware for training; ii) requires massive size dataset for training to achieve reasonable generalization capability \cite{x3}. Factors that are not always available when it comes to industrial applications involving vibration signals. In addition, vibration signals are in 1D, which would make it necessary to modify their dimension, further increasing the computational cost and need for pre-processing. On the other hand, 1D CNN makes it possible to use the vibration signal in its original dimension. It has low computational cost compared to 2D CNN and is well-suited for real-time and low-cost applications especially on mobile or hand-held devices \cite{x3}. For these reasons, it was used in the study.

In this part, the synthetically generated and augmented signals are used to train the model (1D CNN) and adjust the hyperparameters. It is worth mentioning that no real fault signals are used in the training. The test group consists of other real machine signals under different operating conditions. After training, the model is ready to perform the classification.

\subsection{Explainable Artificial Intelligence (XAI)}

Explaining the results obtained by the AI model is essential to ensure its use in the industry, since it is associated with increased trust. Vibration signals are preferably analyzed in the frequency domain in order to facilitate the correlation between the highlighted frequency/behavior and the fault pattern. In this way, applying a method that evaluates the points of relevance in the signal used for the classification, allows the end user to understand and validate the result obtained. For this purpose, the Gradient-weighted Class Activation Mapping (Grad-CAM) approach was used.

After classifying the sample by CNN 1D, the method is applied, returning a heatmap with the most relevant frequencies for the model. The heatmap obtained helps the specialist in decision making, allowing the validation of the classification determined by the model.

\section{Experimental procedure}
\subsection{Data description}

Three datasets were used to address different faults found in rotating machinery, two of which were publicly available \cite{QIU20061066, 8360102} and one developed by the author for the study. As shown in \cite{yleibook} even though the rotating machinery is diversified, some common essential rotating parts are: rotors, rolling element bearings, and gears \cite{xai_lucas}. Therefore, the datasets were chosen and developed to address the main faults present in rotating machinery components, namely: defects in bearing and gearbox, misalignment, unbalance and mechanical looseness. The use of different datasets aims to validate the proposed methodology in different scenarios. As they are publicly available datasets and already explored in \cite{xai_lucas}, Case 1: Bearing Dataset and Case 2: Gearbox Dataset will be briefly discussed. For more details, please refer to \cite{QIU20061066, 8360102, xai_lucas}. All datasets are vibration signals collected using accelerometers.

\subsubsection{Case 1: Bearing dataset}

The dataset \cite{QIU20061066} is composed of remaining useful life (RUL) bearing tests, each file consisting of 20,480 points with the sampling rate set at 20 kHz. As this is a run-to-failure test, no labels are available (the only information provided is the type of fault present at the end of each test). Therefore, the data were analyzed and labeled manually. For the study the bearing 01 of test 02 was used. The test presents 984 observations, with the first 531 observations labeled as normal and the last 453 as fault in the outer race (BPFO).

\subsubsection{Case 2: Gearbox dataset}

To evaluate faults in gears, the dataset \cite{8360102} was used. Nine different gear conditions were introduced to the pinion on the input shaft, including healthy condition, missing tooth, root crack, spalling, and chipping tip with five different levels of severity. For each gear condition, 104 observations were collected resulting in a total of 936 observations with sampling frequency of 20 kHz. For the test, the samples were considered as: 104 normal and 832 gear fault.

\subsubsection{Case 3: Mechanical faults dataset}

The last dataset was developed by the authors, aiming to address classic faults in rotating machines such as: unbalance, misalignment and mechanical looseness, in addition to the normal operating condition.

\begin{figure*}[ht]
  \centering
  \includegraphics[scale=0.05]{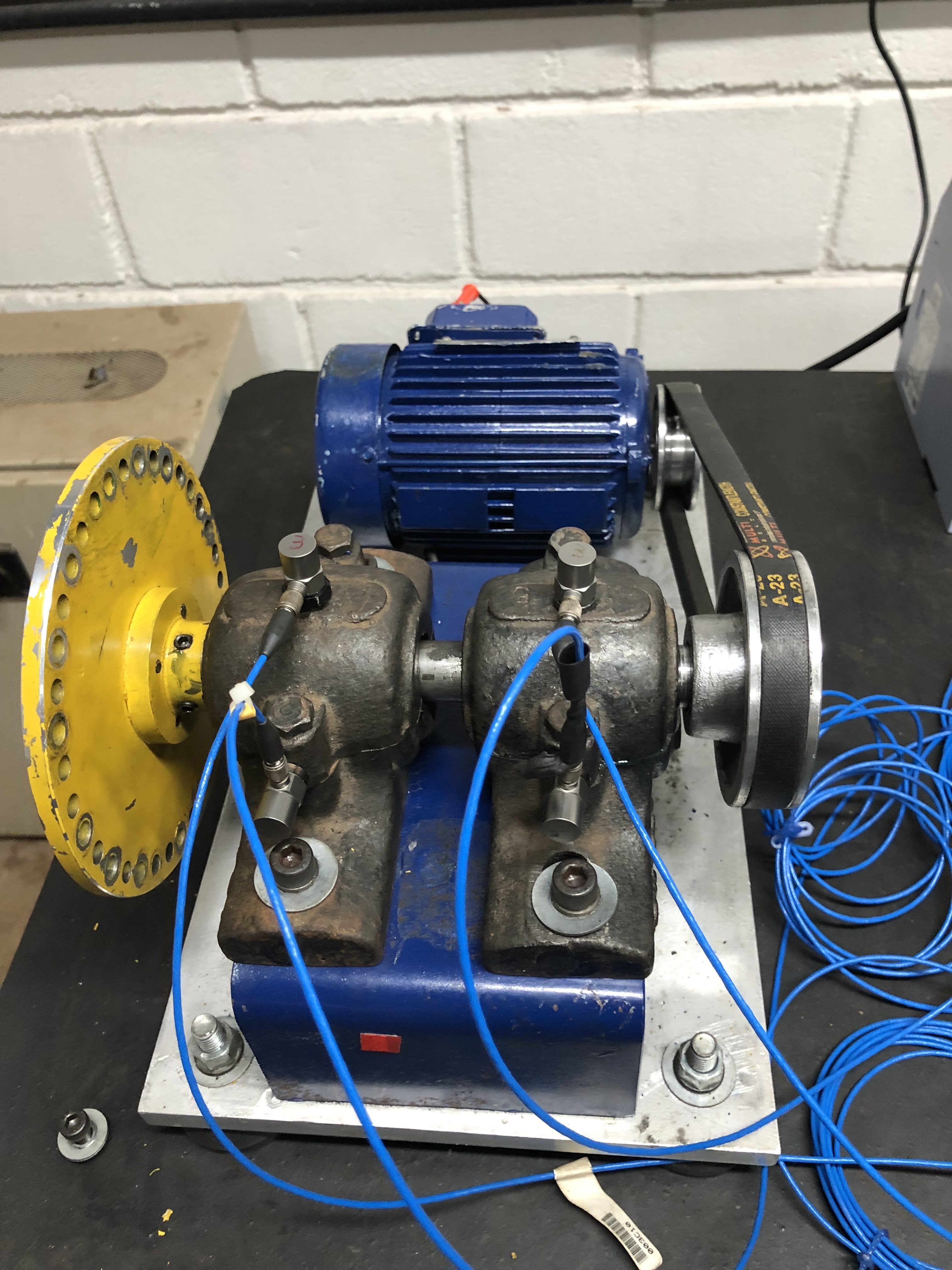}
  \caption{Bench test.}
  \label{fig:Bench}
\end{figure*}

The faults were introduced on a test bench, Fig. \ref{fig:Bench}, being: motor, frequency inverter, bearing house, two bearings, two pulleys, belt and rotor (disc).

- Motor: Three-phase induction motor, B56 B4, Manufacturer Eberle, nominal speed 1650 rpm, power 0.09 kW, voltage 220 V, current 0.70 A, bearings 6200 ZZ.

- Frequency Inverter: Vector Inverter, model CFW300, manufacturer WEG.

- Driven Pulley (Disc Side): model A80, external diameter 80 mm, internal diameter 53 mm.

- Motor Pulley (Motor Side): model A60, external diameter 60 mm, internal diameter 37 mm.

- Belt: model V - A23, top width 12.7 mm, outer perimeter 630 mm, inner perimeter 580 mm.

- Rotor (Disc): diameter 149 mm, 36 holes, thickness 6.5 mm.

- Bearings 1205, bearing house SN505, bushing H305.

For the acquisition of the signals were used: 04 accelerometers PCB 352C33 (2 in each bearing), mounted in the vertical and horizontal positions (y- and x-axes), 01 Power Supply PCB 482A20, 01 acquisition board Hi-Speed USB Carrier, NI USB-9162. A Python language program was developed to collect the data. All accelerometer sensitivities were adjusted according to the calibration chart. The rotation was kept constant with a value measured on the shaft of approximately 1238 rpm.

20 tests were performed, 5 for each condition (normal, unbalance, misaligment and looseness). Each test consists of 4 sets of 420 signals collected continuously, each file consisting of 25,000 points with the sampling rate set at 25 kHz (420 signals per accelerometer). Resulting in the end of all tests, in a total of 8400 signals per accelerometer.

The sequence of tests was randomly defined. Before starting any test, the bench was dismantled and returned to normal operating condition, to later introduce the fault. The experimental procedure allows variations to occur, making the tests closer to industrial reality. All data used in this study will be made available at the link: https://data.mendeley.com/datasets/zx8pfhdtnb/1

To increase the randomness of the tests in the fault classification (AI), and to validate the robustness of the model, only 3 tests per condition were used in the analyses, resulting in 12 tests in total (5040 signals). Since the fault classification was performed 10 times, at each new run of the model, 3 tests were randomly selected for each class of fault.

As it is a small bench test, the measurement points are close, and in order to reduce the computational cost, only the signals from the horizontal position of the accelerometer present in the bearing (near the pulley) were used in the analyses.

\subsection{Hyperparameter tuning and evaluation metrics}

The hyperparameters were adjusted based on the training set to obtain the best performance and stability in the model. The cross-validation method was used, where the signals were randomly chosen between the training and validation group, being 90\% for training and 10\% for validation. It is worth mentioning that the training group is composed only of synthetic signals, and therefore does not contain any real fault signals. The test set is composed of all signals excluding the normal signals that were used for generating the synthetic signals.

The model was composed of a 1D CNN layer, followed by a dropout layer for regularization, then a pooling layer. The objective of using only 1 layer is to reduce the complexity and computational cost of the model. As CNN has high learning capacity, the dropout layer was used to slow down the learning process and hopefully avoid overfitting. Pooling learning reduces the amount of learned features by consolidating them and keeping only the essential ones, which prevents the network from learning features uncorrelated with the condition.

The learned features are flattended into a single long vector and passed to the fully connected layer before the output layer used to make a prediction. The fully connected layer provides a buffer between learned features and the output with the intention of interpreting learned features before making a prediction.

For the model, 32 parallel feature maps and kernel size of 5 were used. ReLu activation function and Adam optimizer were also used. To automate the training, the model was trained with a variable number of epochs, based on the Early Stopping technique, with a patience of 8 and a variation of 0.001 in relation to the validation accuracy. Batch size of 32 samples was used. Categorical cross entropy loss function and Softmax (activation function) were used as it is a multi-class classification problem.

The model was trained to classify 7 conditions, regardless of whether the case under analysis presents the fault or not, being: normal condition, BPFO, BPFI, unbalance, misalignment, looseness and gear fault. This simulates a condition where different faults may appear on rotating machinery, and the model will be prepared to identify them. In addition, it increases the complexity of the analyses.

Neural networks are stochastic, so at each new training a different model will be obtained, even if using the same dataset. It is worth mentioning that, in addition to the stochasticity of the networks, there is also the randomness in the generation and augmentation of the signals. To ensure that the model is stable and robust to variations in both network training and signal generation, the analyzes were repeated 10 times. At each analysis, new synthetic signals were randomly generated and the model trained. As in all dataset, the number of samples per condition were approximately equal, accuracy was used. For Case 2 where the test data are imbalanced, despite the use of accuracy, the confusion matrix is shown. To measure the variability after the 10 analyses, the standard deviation was calculated. The tests were performed using 2.2 GHz Intel Core i7 Dual-Core, 8 GB 1600 MHz DDR3, Intel HD Graphics 6000 1536 MB.

\subsection{Analysis approaches}

The success of an AI model is directly related to the quality and quantity of training data. On the other hand, a large amount of data implies acquisition costs, as well as a low quality implies a reduction in the model's accuracy. Being able to reduce the amount of real data needed, and generate data varied enough to make the model robust, is a solution to enable the implementation of AI in the industrial application.

Besides the main objective of verifying the possibility of using a model trained with synthetic and augmented data to evaluate real signals and their explainability, other analyzes were performed.

First, the augmented data are analyzed, verifying if the behavior of the synthetic signal in the frequency domain, matches the available knowledge about fault detection in rotating machinery using vibration analysis. Subsequently, in order to reduce the amount of real data needed to generate the synthetic data, the minimum viable amount of data was analyzed to obtain a stable and robust model. This approach can be seen as an analogy of the study area called Few-Shot Learning, where few samples are used to train the model. Another analysis performed is in relation to the total amount of real samples needed in the training group to ensure the stability and robustness of the model. In addition, the possibility of working with 100\% synthetic samples was analyzed, that is, where synthetic data are generated without dependence on the real sample. For comparison between training forms, supervised training was performed, using only real signals, which were divided into 70\% for training and 30\% for testing. Finally, the data obtained through XAI are analyzed to verify if the features learned by the model are capable of providing relevant information to the expert for decision making.

\section{Results and discussion}
\subsection{Data exploration}

In this section the synthetic samples for Case 1-3 are analyzed, discussed and compared with the real samples. One sample per condition from the training database was randomly selected. Due to the randomness of FaultD-XAI, variations in signals may occur, being important to observe whether the deterministic characteristics of the conditions remain similar.

As the objective is only to visualize the signal characteristic, df = 1 hz was used, with the exception of Case 2, where df = 5.5 hz due to data limitation. For BPFO, BPFI and Gear Fault conditions, the full signal is presented to highlight the fault characteristic at high frequency. In Unbalance, Misalignment, Looseness, a zoom was performed in the range from 0 to 500 hz to facilitate the visualization of the characteristic frequencies. For the normal condition, the complete signal is displayed.

\subsubsection{Case 1 - Bearing Fault}
\begin{figure}[ht]
  \subfloat[Real - Normal]{
	\begin{minipage}[c][0.65\width]{
	   0.3\textwidth}
	   \centering
       \label{fig:bpfo_original}
	   \includegraphics[width=1\textwidth]{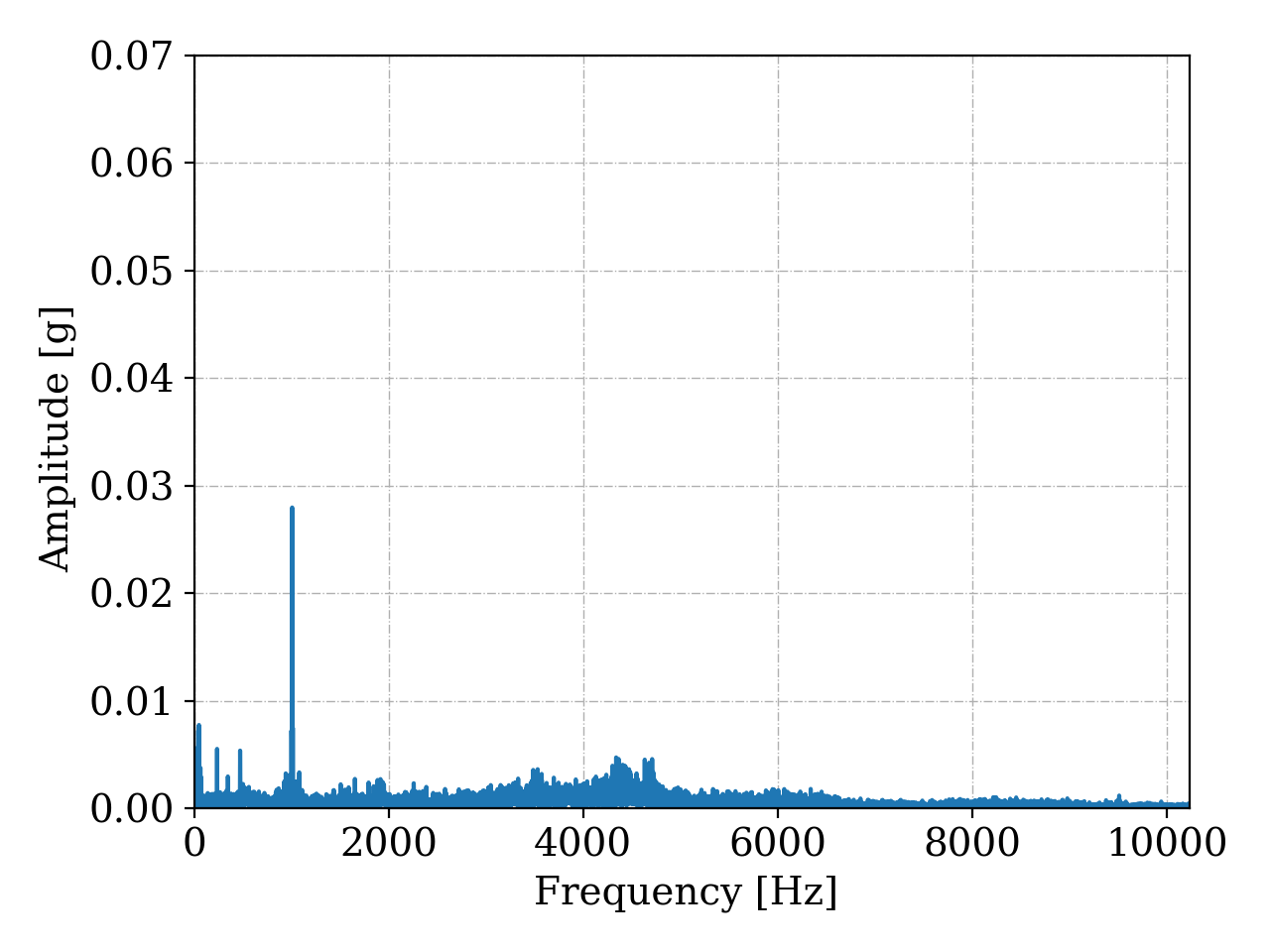}
	\end{minipage}}
 \hfill 
  \subfloat[Real - BPFO]{
	\begin{minipage}[c][0.65\width]{
	   0.3\textwidth}
	   \centering
        \label{fig:bpfo_falha_original}
	   \includegraphics[width=1\textwidth]{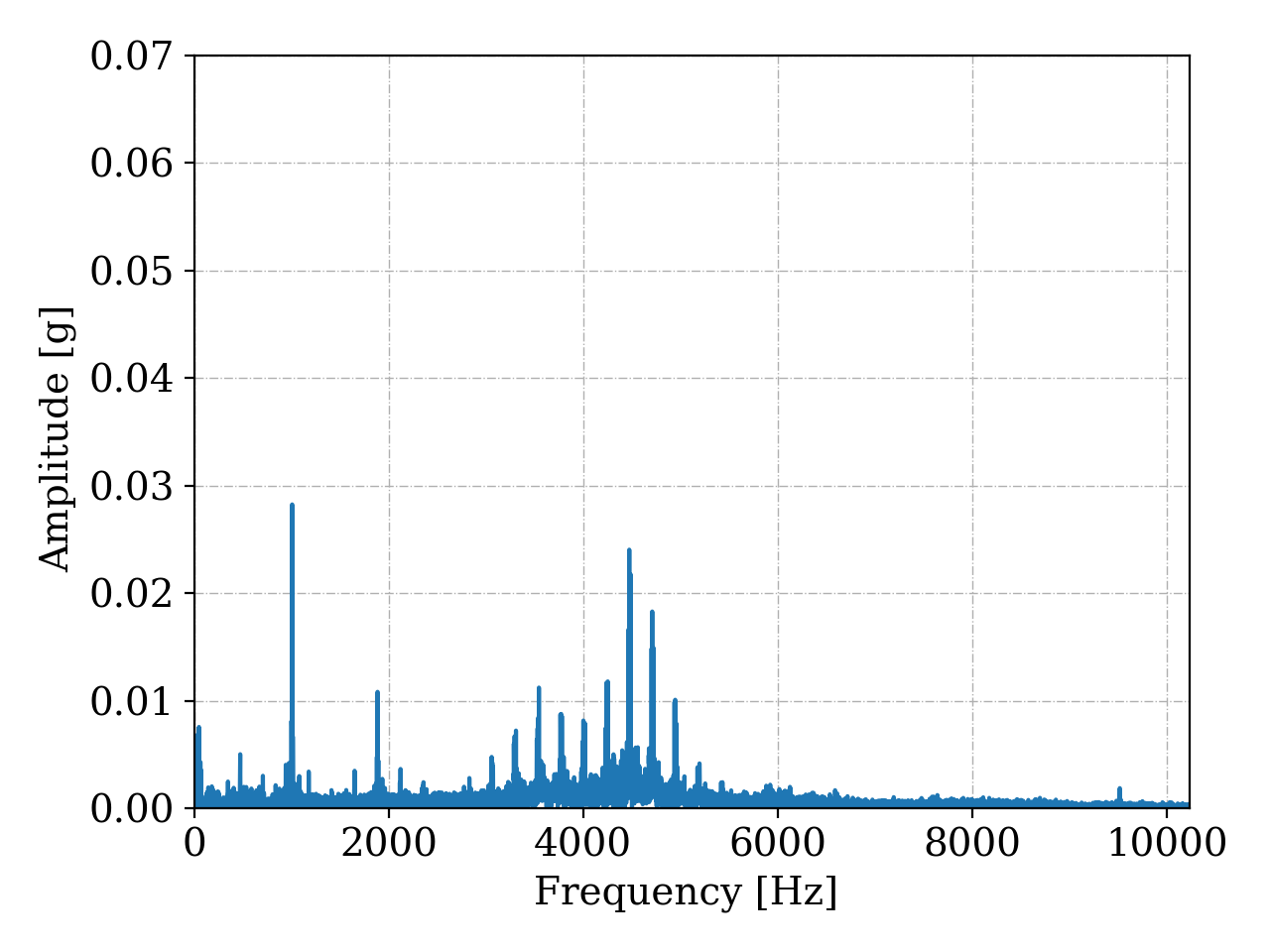}
	\end{minipage}}
 \hfill	
  \subfloat[Synthetic - Normal]{
	\begin{minipage}[c][0.65\width]{
	   0.3\textwidth}
	   \centering
	   \includegraphics[width=1\textwidth]{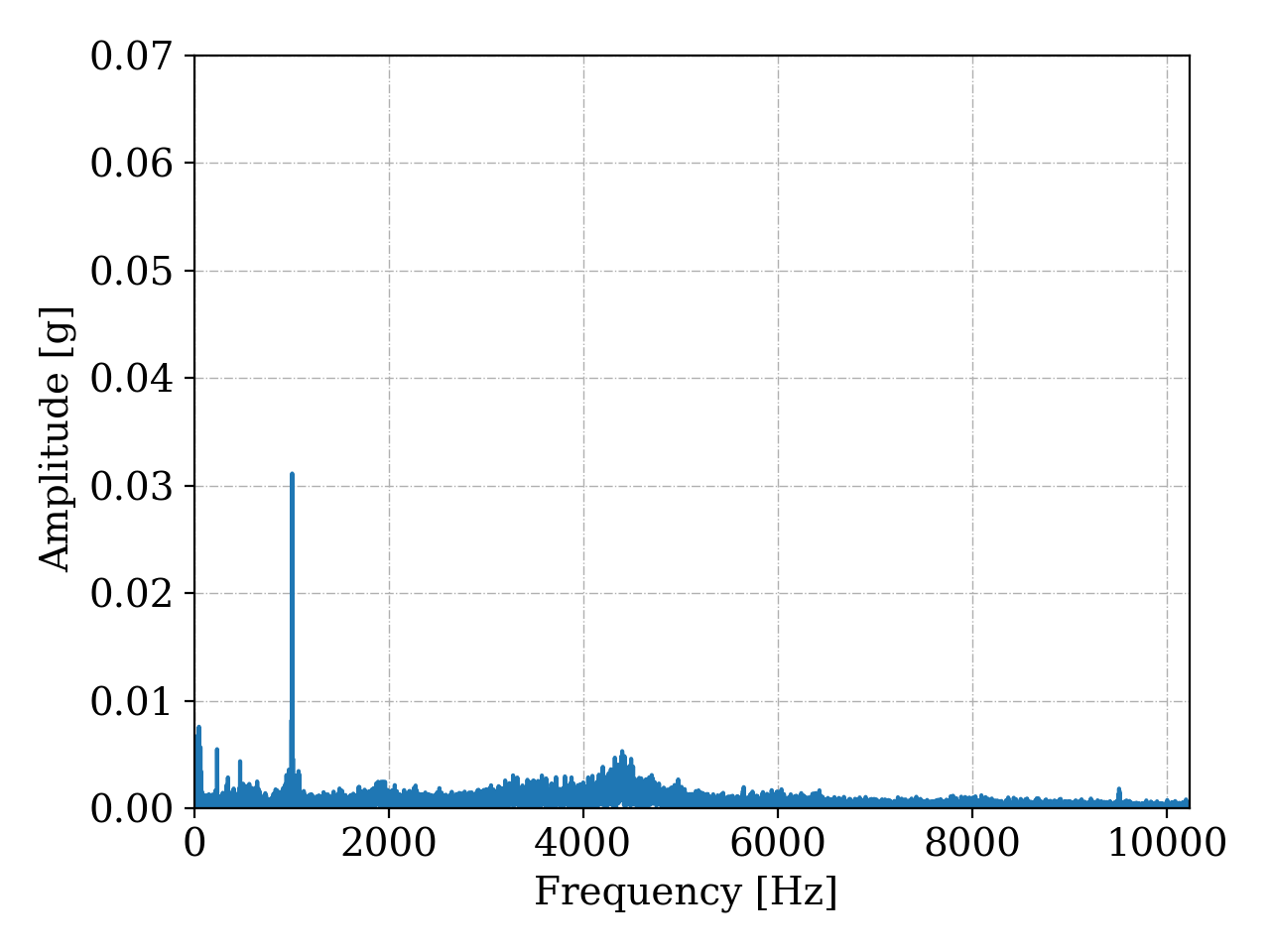}
	\end{minipage}}\\
  \subfloat[Synthetic - BPFO]{
	\begin{minipage}[c][0.65\width]{
	   0.3\textwidth}
	   \centering
       \label{fig:bpfo_syn}
	   \includegraphics[width=1\textwidth]{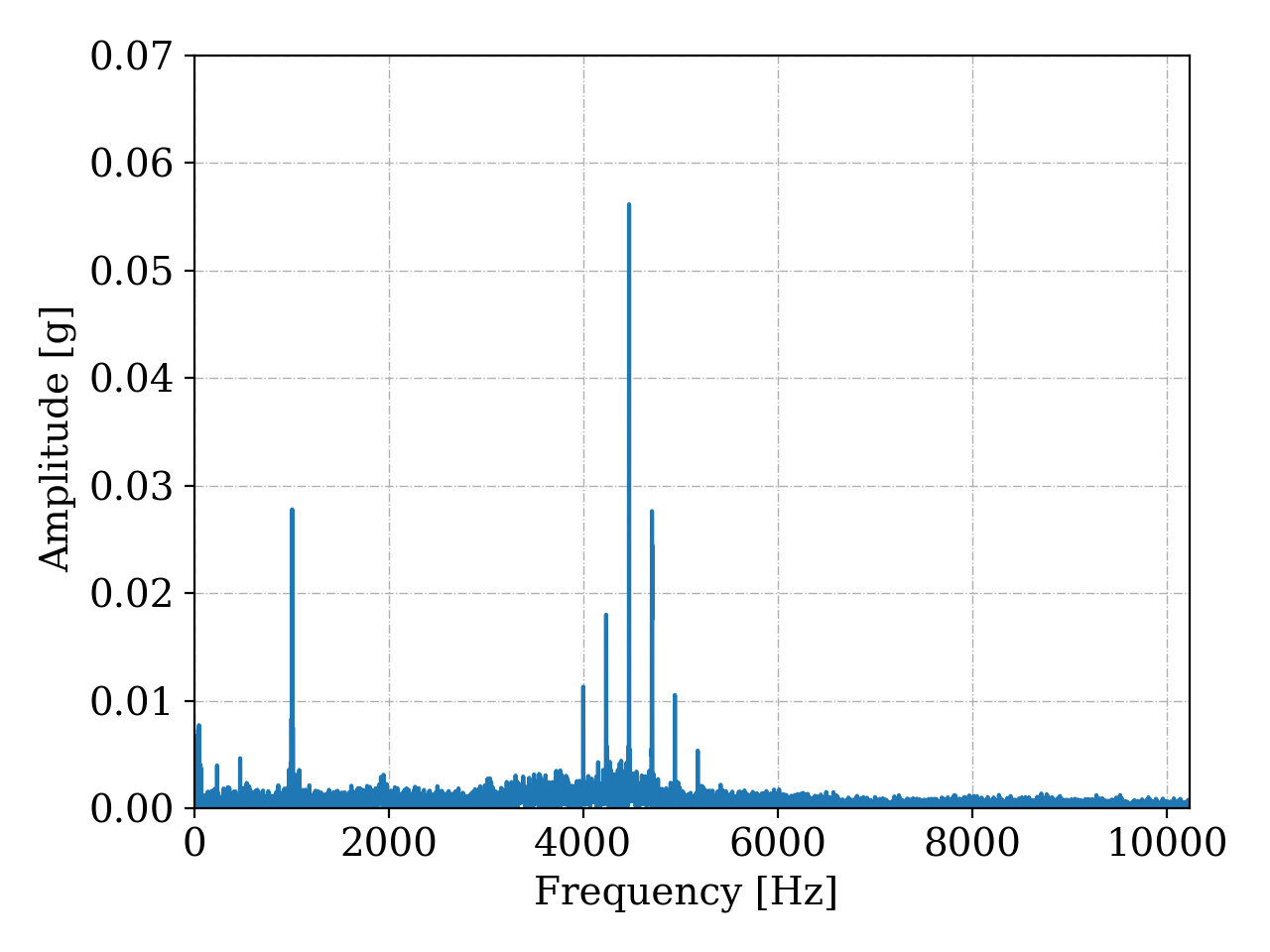}
	\end{minipage}}
 \hfill 	
  \subfloat[Synthetic - BPFI]{
	\begin{minipage}[c][0.65\width]{
	   0.3\textwidth}
	   \centering
       \label{fig:bpfi_syn}
	   \includegraphics[width=1\textwidth]{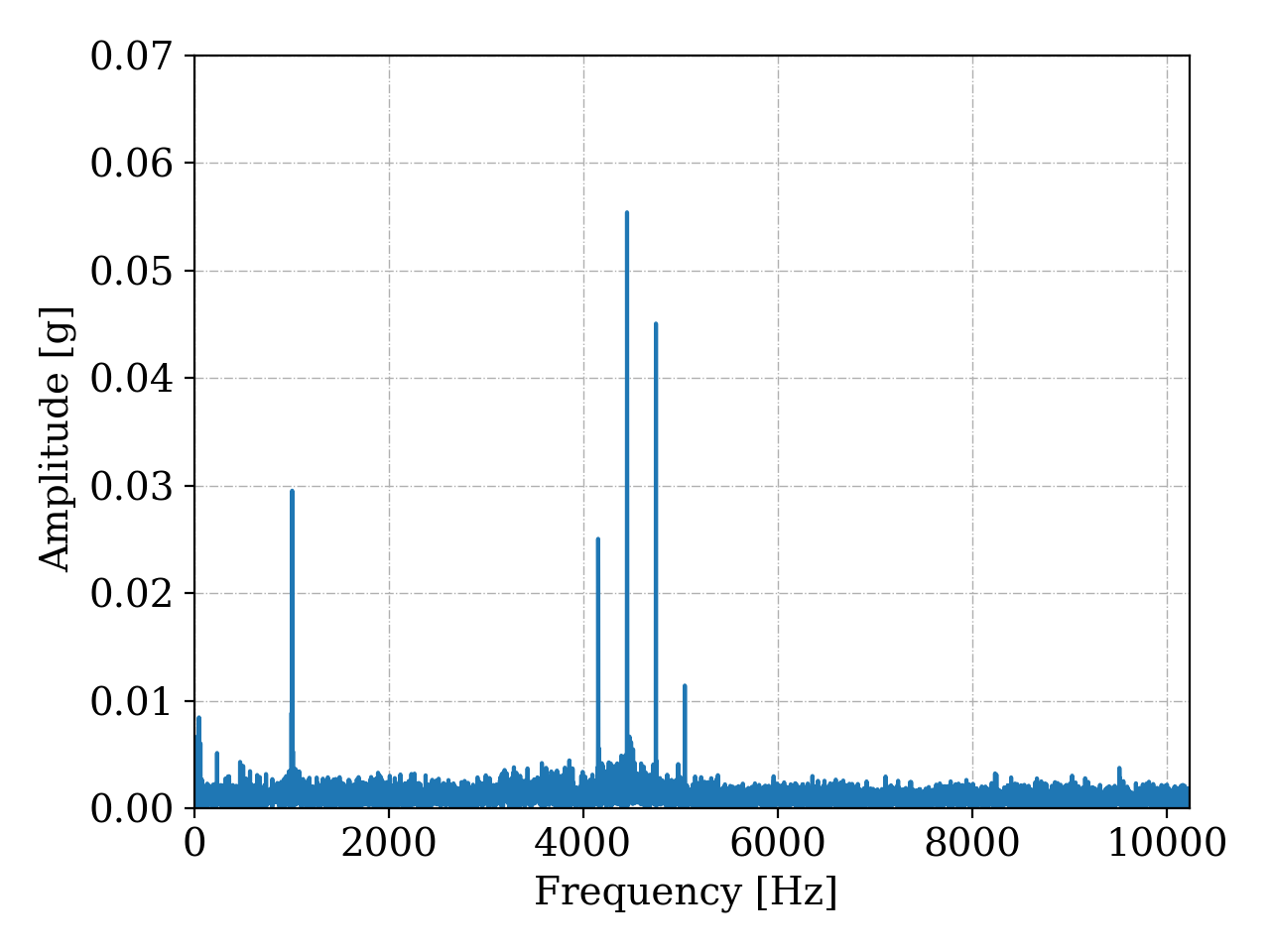}
	\end{minipage}}
  \hfill 	
  \subfloat[Synthetic - Unbalance]{
	\begin{minipage}[c][0.65\width]{
	   0.3\textwidth}
	   \centering
       \label{fig:unbalance_syn}
	   \includegraphics[width=1\textwidth]{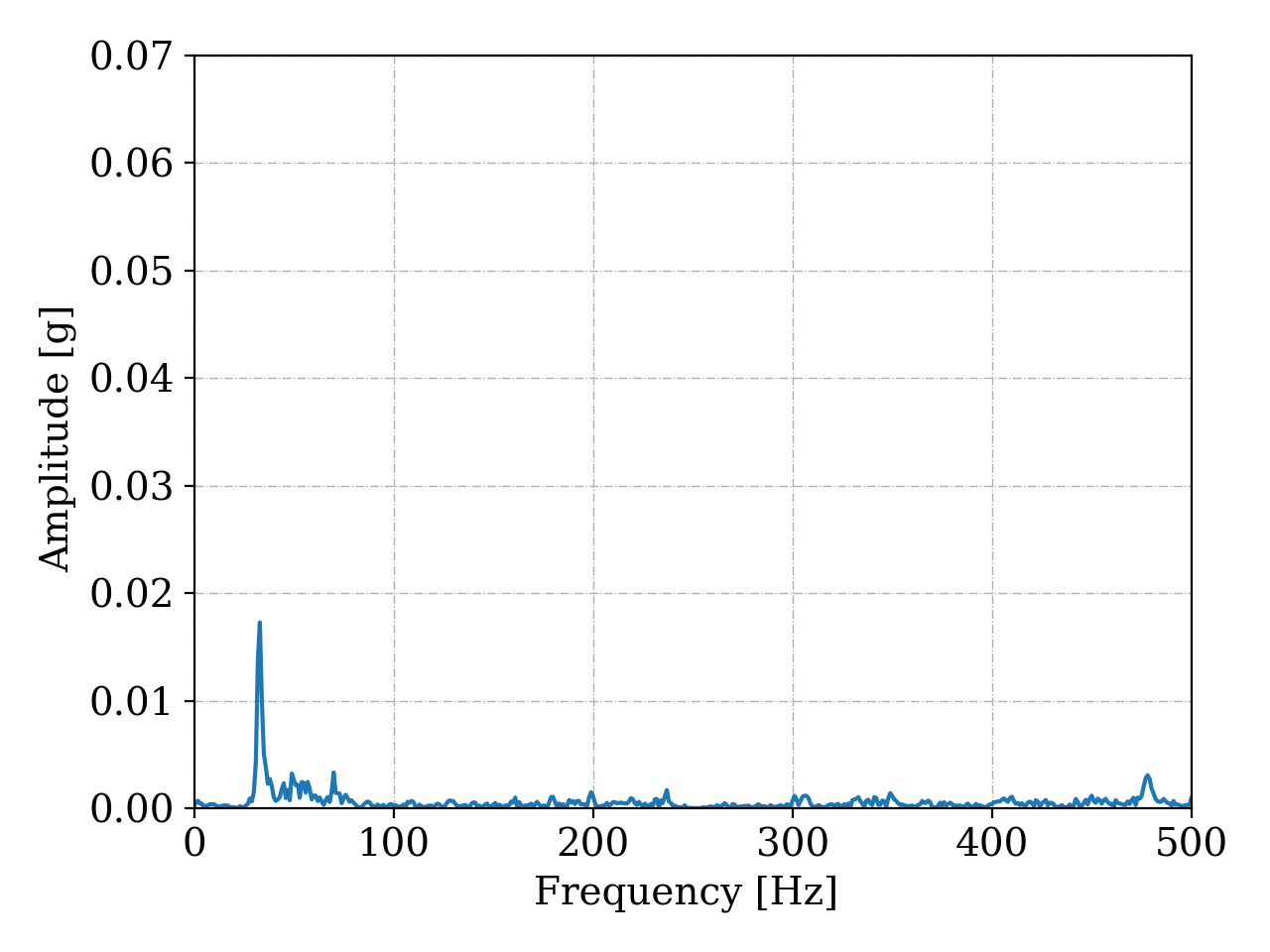}
	\end{minipage}}\\
   \hfill 	
  \subfloat[Synthetic - Misalignment]{
	\begin{minipage}[c][0.65\width]{
	   0.3\textwidth}
	   \centering
       \label{fig:mis_syn}
	   \includegraphics[width=1\textwidth]{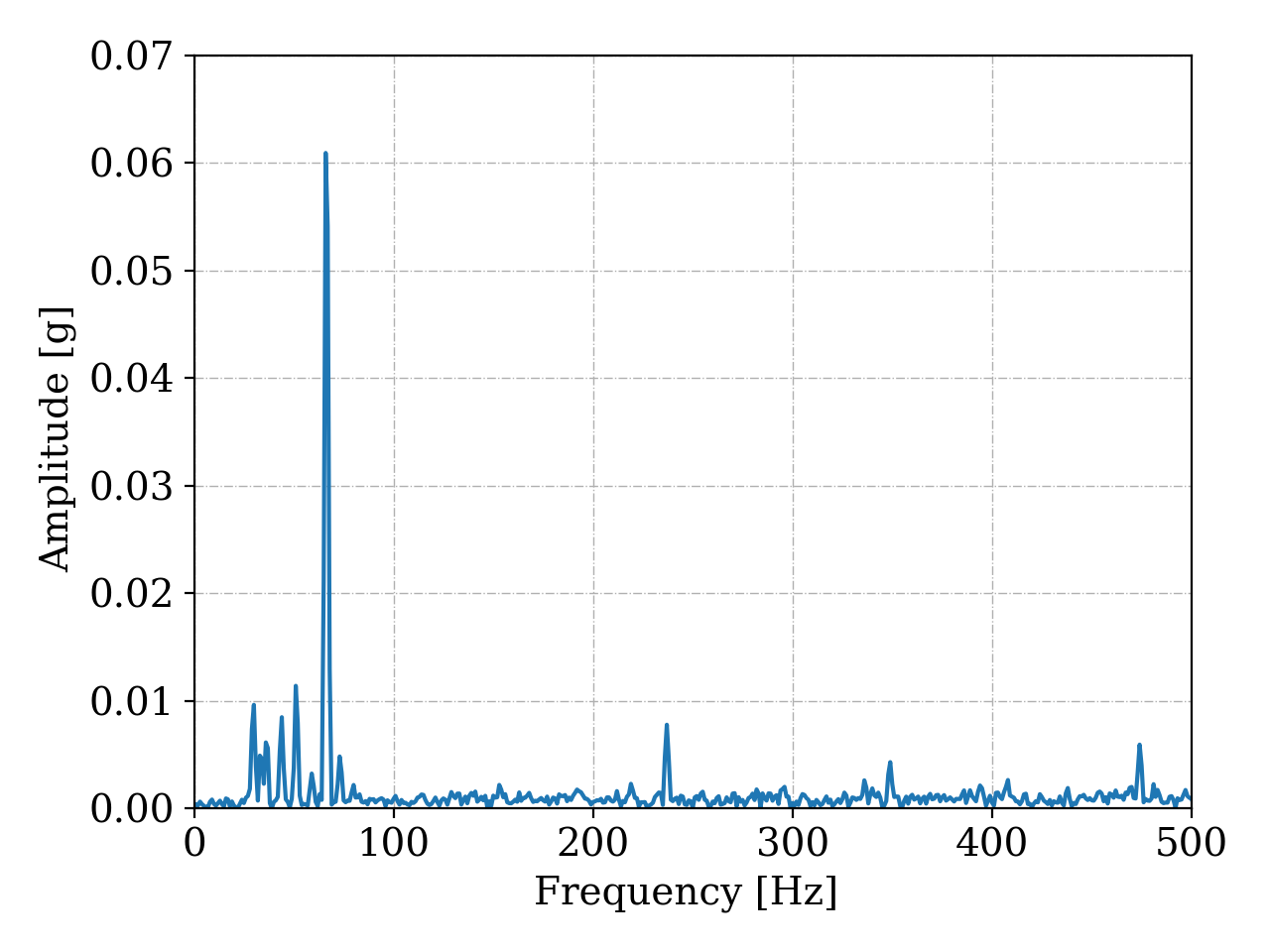}
	\end{minipage}}
   \hfill 	
  \subfloat[Synthetic - Looseness]{
	\begin{minipage}[c][0.65\width]{
	   0.3\textwidth}
	   \centering
       \label{fig:loos_syn}
	   \includegraphics[width=1\textwidth]{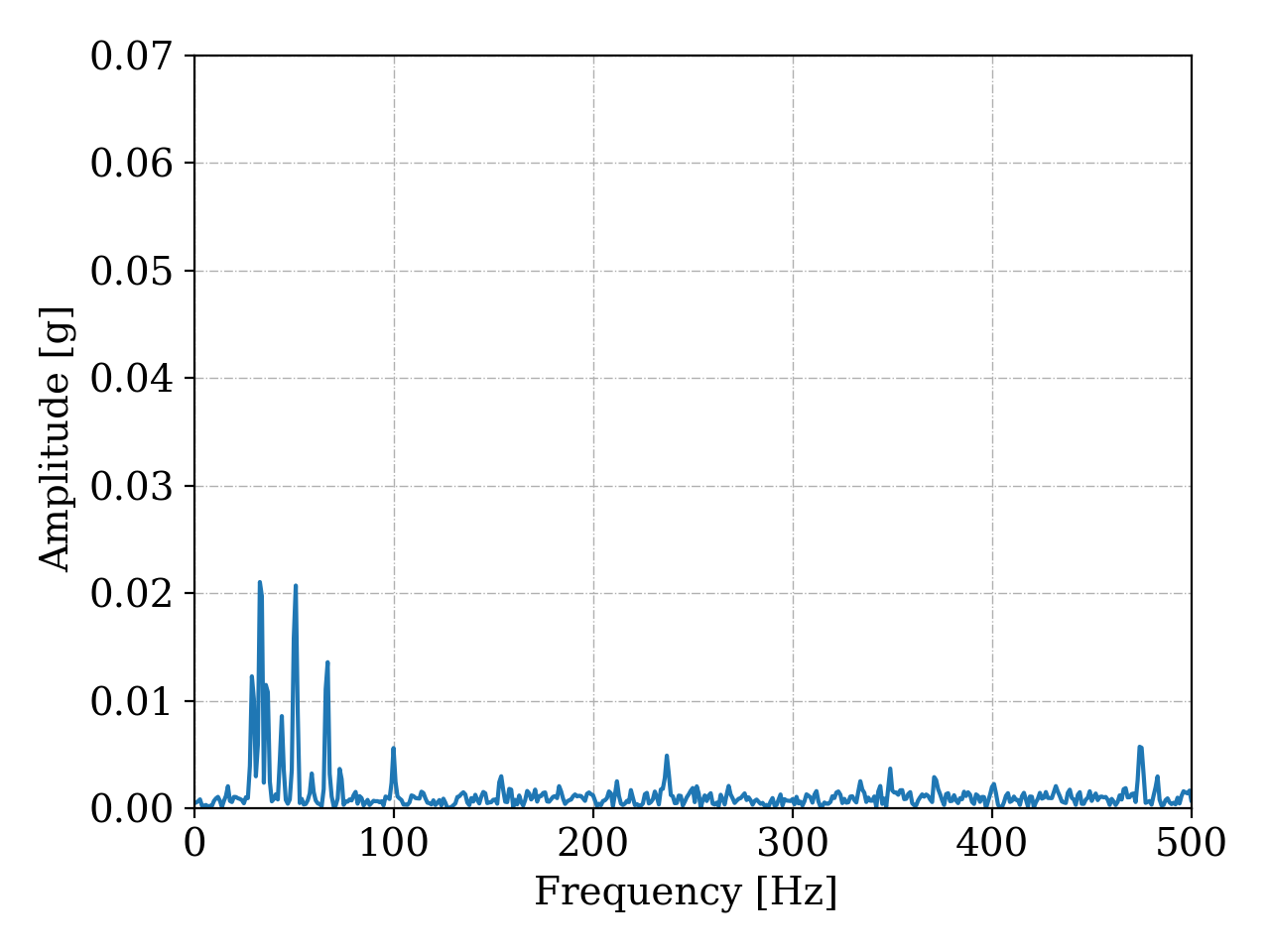}
	\end{minipage}}
   \hfill 	
  \subfloat[Synthetic - Gear Fault]{
	\begin{minipage}[c][0.65\width]{
	   0.3\textwidth}
	   \centering
       \label{fig:gear_syn}
	   \includegraphics[width=1\textwidth]{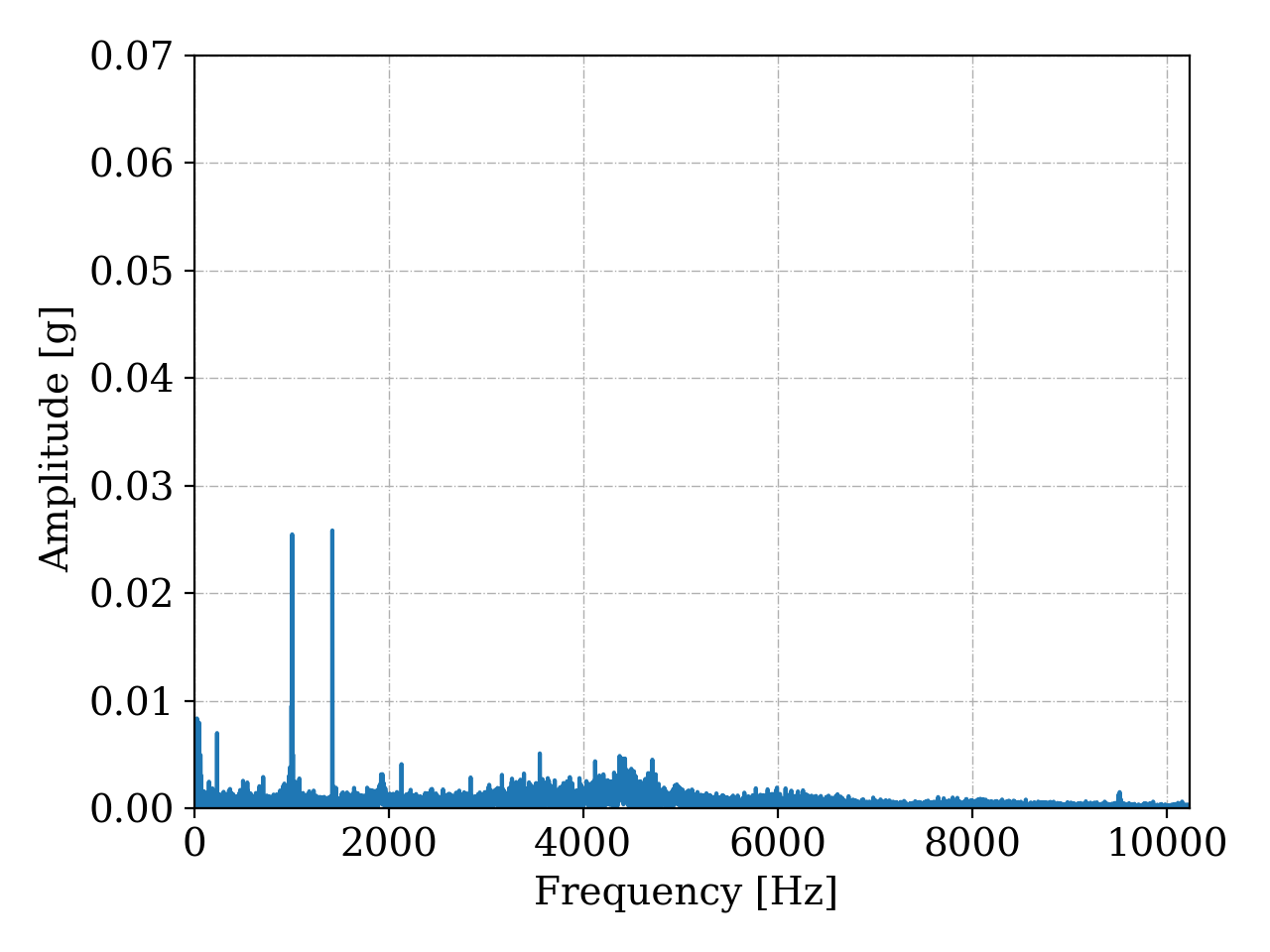}
	\end{minipage}}
    \caption{Examples of vibration signals for Case 1, real and synthetic.}
    \label{fig:sinalBPFOfaults}	
\end{figure}

In Fig. \ref{fig:sinalBPFOfaults} the original signals for the two conditions (Normal and BPFO) and the synthetic signals for the seven created conditions (Normal, BPFO, BPFI, Unbalance, Misalignment, Looseness and Gear Fault) from Case 1 are presented.

\renewcommand{\baselinestretch}{3} 
\renewcommand{\baselinestretch}{1.5} 

The real normal condition signal, Fig.\ref{fig:bpfo_original} does not present any frequency related to faults, showing only the characteristic behavior of the machine. On the other hand, in Fig. \ref{fig:bpfo_falha_original} there is the appearance of frequencies in the region of 3000 to 6000 Hz, related to excitation of the natural frequencies, caused by the fault in the outer race.

Analyzing the synthetic signals, it can be noted that for bearing faults, the impacts due to a small defect tend to excite the races natural frequencies (at high frequency) Fig.\ref{fig:bpfo_syn} e \ref{fig:bpfi_syn}, being the fundamental frequency corresponding to the damaged element, BPFO = 236.4 hz and BPFI = 296.9 hz. For the unbalance, Fig.\ref{fig:unbalance_syn}, there is a predominant increase in amplitude by 1 x fr (rotation frequency). In misalignment, there is an increase in 2 and 3 x fr, with emphasis on 2 x fr, Fig.\ref{fig:mis_syn}. Finally, for the mechanical looseness, there is an increase in harmonics and the appearance of some sub-harmonics of fr, Fig.\ref{fig:loos_syn}. Gear defect, Fig.\ref{fig:gear_syn} show an appearance of frequencies related to GMF, hypothetically placed at 711 hz and its harmonics, since this dataset does not have gears.

Comparing the synthetic and original (real) signals for the conditions present in the dataset (normal and BPFO), the similarity of the fundamental characteristics is noted, ensuring that the model learns the behavior of the signal in relation to each operating condition of the equipment.

\subsubsection{Case 2 - Gear Fault}
\begin{figure}[ht]
  \subfloat[Real - Normal]{
	\begin{minipage}[c][0.65\width]{
	   0.3\textwidth}
	   \centering
       \label{fig:gear_original}
	   \includegraphics[width=1\textwidth]{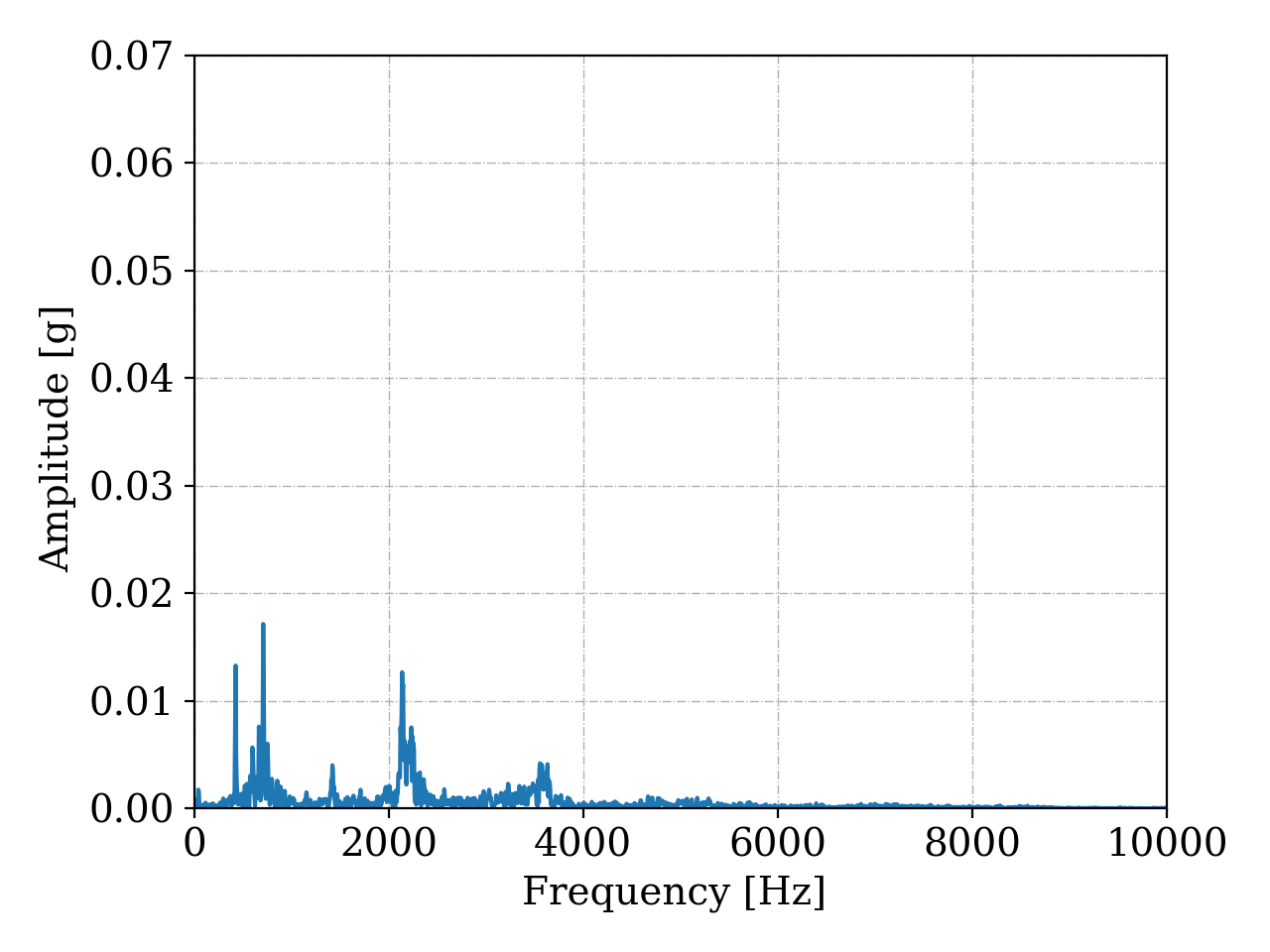}
	\end{minipage}}
 \hfill 
  \subfloat[Real - Gear Fault]{
	\begin{minipage}[c][0.65\width]{
	   0.3\textwidth}
	   \centering
        \label{fig:gear_falha_original}
	   \includegraphics[width=1\textwidth]{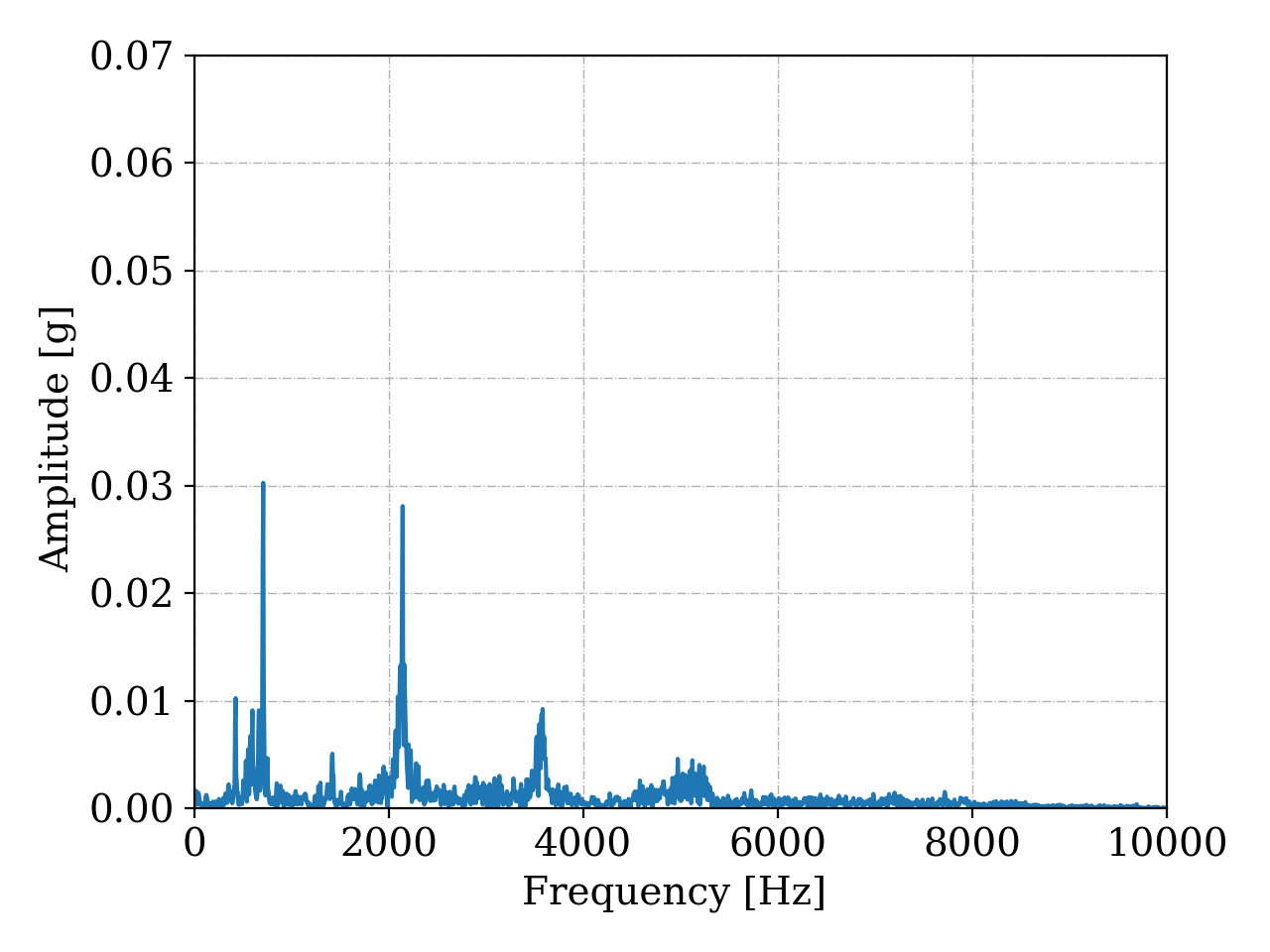}
	\end{minipage}}
 \hfill	
  \subfloat[Synthetic - Normal]{
	\begin{minipage}[c][0.65\width]{
	   0.3\textwidth}
	   \centering
	   \includegraphics[width=1\textwidth]{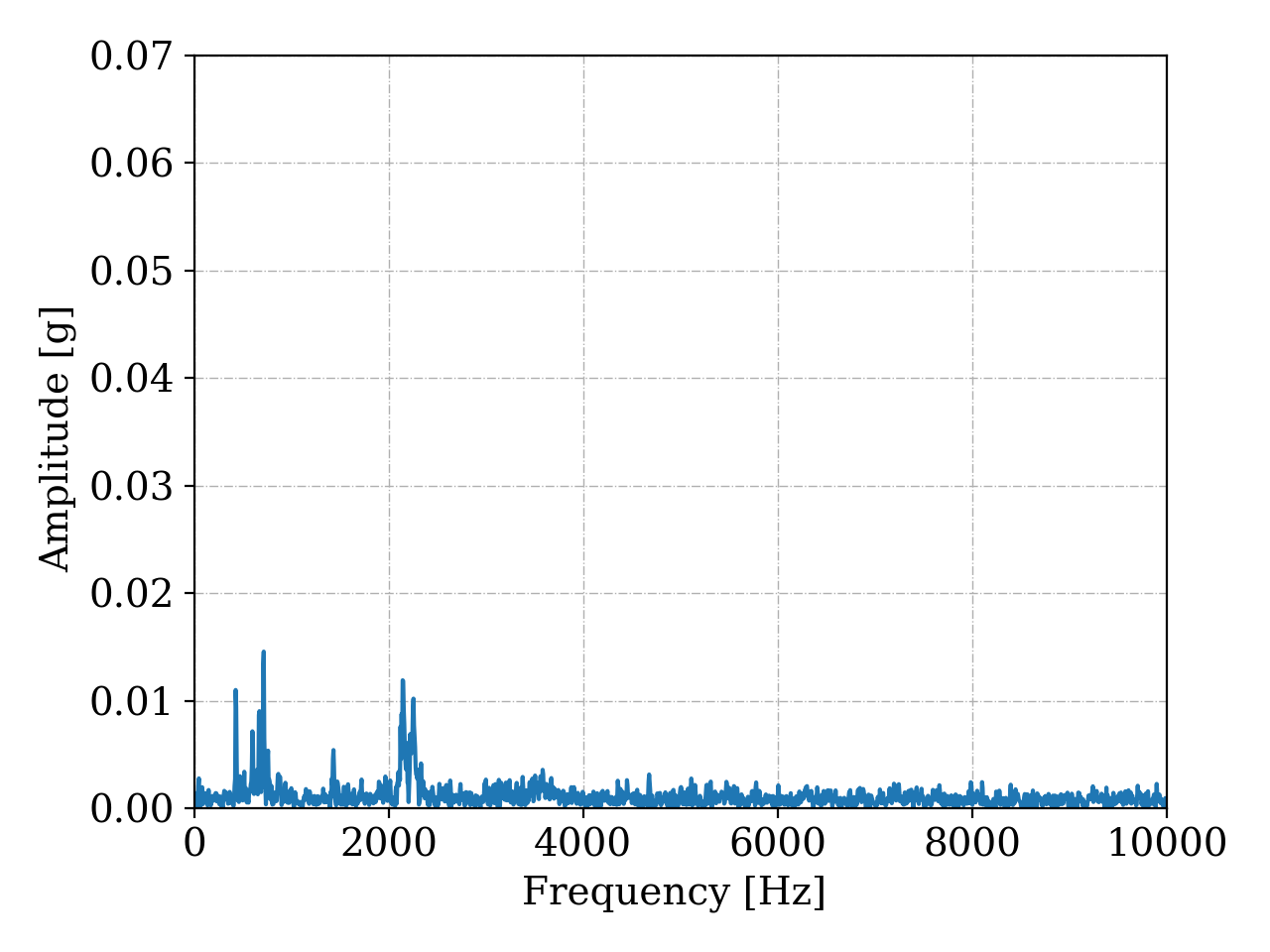}
	\end{minipage}}\\
  \subfloat[Synthetic - BPFO]{
	\begin{minipage}[c][0.65\width]{
	   0.3\textwidth}
	   \centering
       \label{fig:gear_bpfo_syn}
	   \includegraphics[width=1\textwidth]{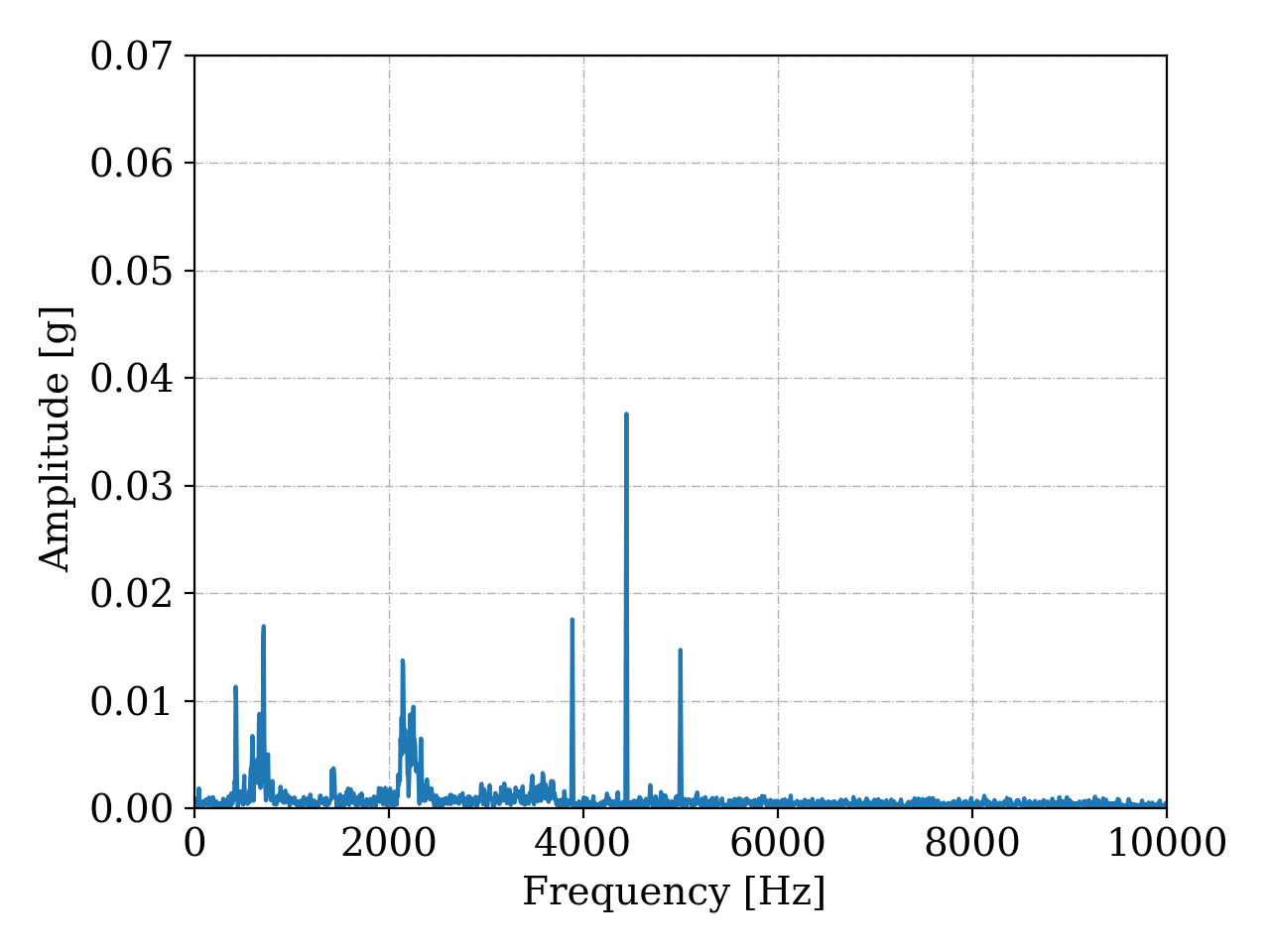}
	\end{minipage}}
 \hfill 	
  \subfloat[Synthetic - BPFI]{
	\begin{minipage}[c][0.65\width]{
	   0.3\textwidth}
	   \centering
       \label{fig:gear_bpfi_syn}
	   \includegraphics[width=1\textwidth]{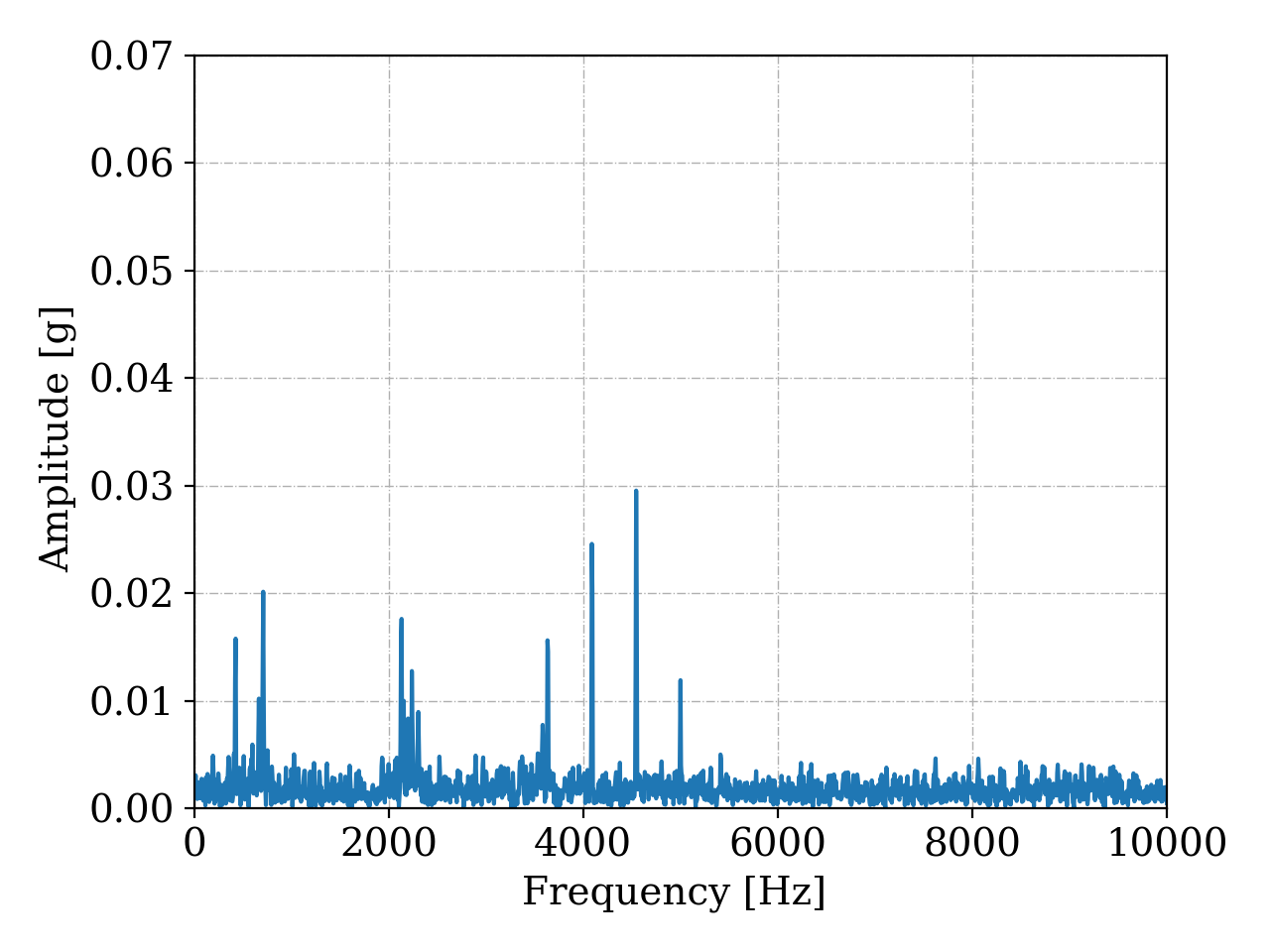}
	\end{minipage}}
  \hfill 	
  \subfloat[Synthetic - Unbalance]{
	\begin{minipage}[c][0.65\width]{
	   0.3\textwidth}
	   \centering
       \label{fig:gear_unbalance_syn}
	   \includegraphics[width=1\textwidth]{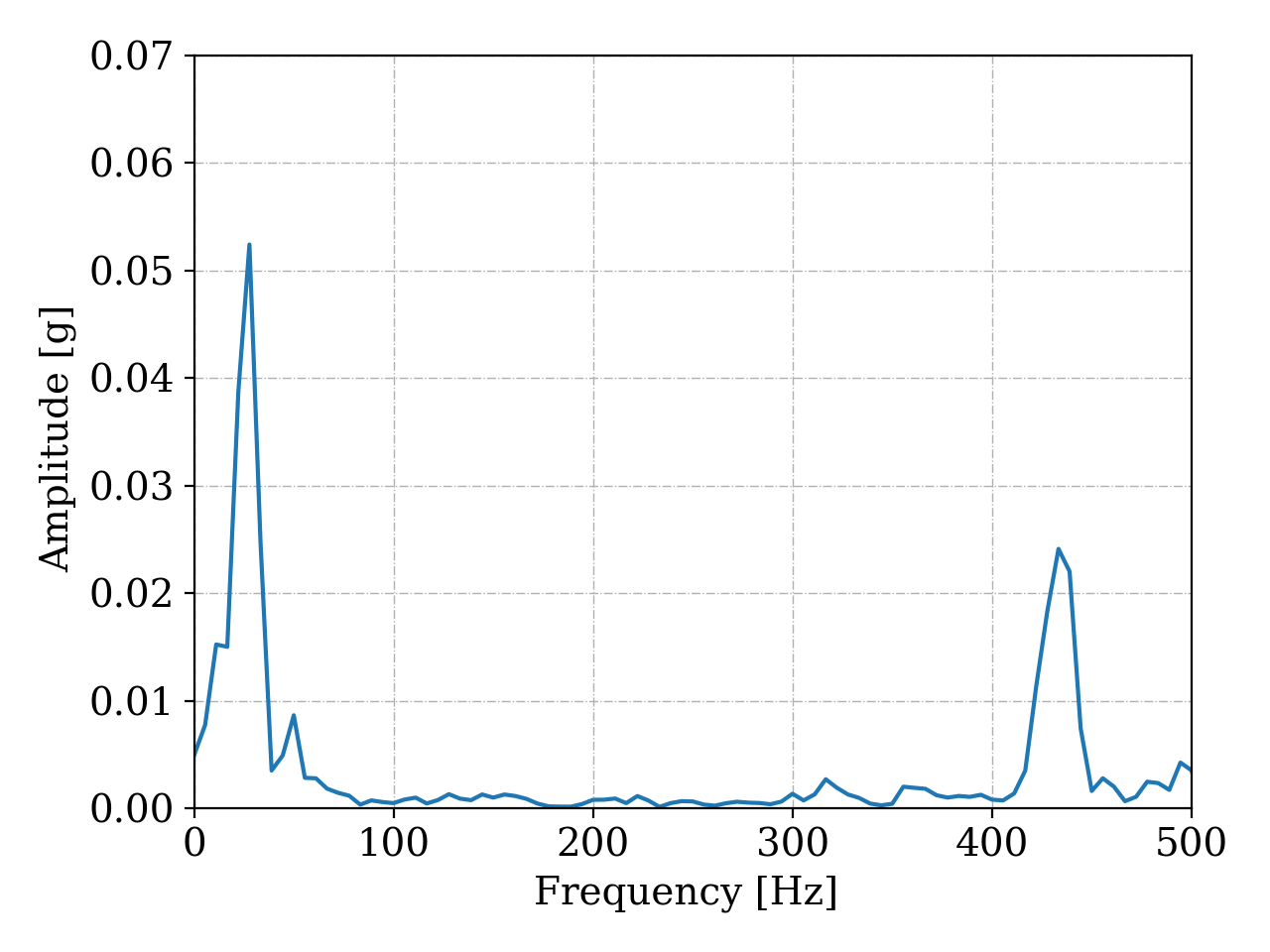}
	\end{minipage}}\\
   \hfill 	
  \subfloat[Synthetic - Misalignment]{
	\begin{minipage}[c][0.65\width]{
	   0.3\textwidth}
	   \centering
       \label{fig:gear_mis_syn}
	   \includegraphics[width=1\textwidth]{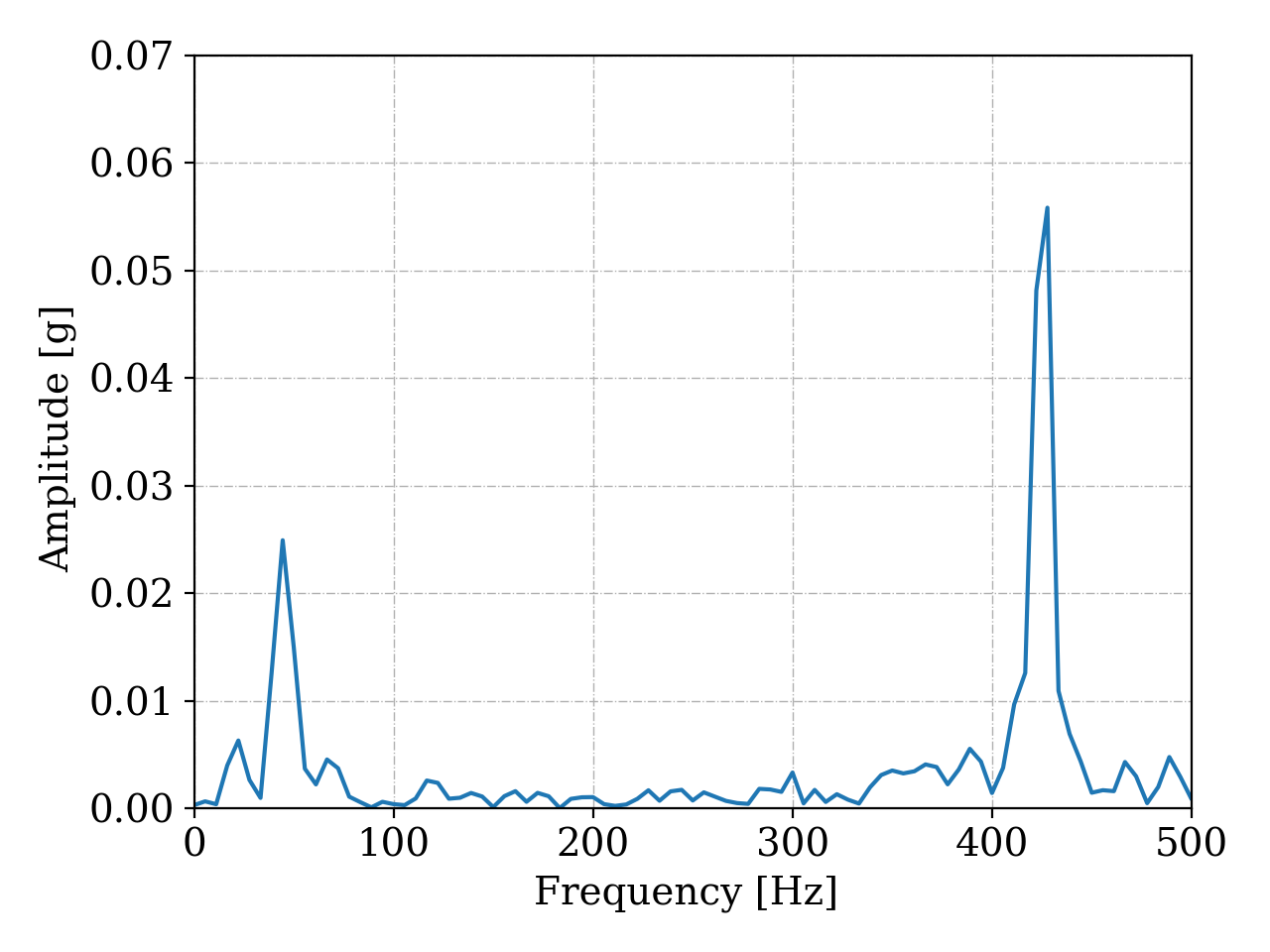}
	\end{minipage}}
   \hfill 	
  \subfloat[Synthetic - Looseness]{
	\begin{minipage}[c][0.65\width]{
	   0.3\textwidth}
	   \centering
       \label{fig:gear_loos_syn}
	   \includegraphics[width=1\textwidth]{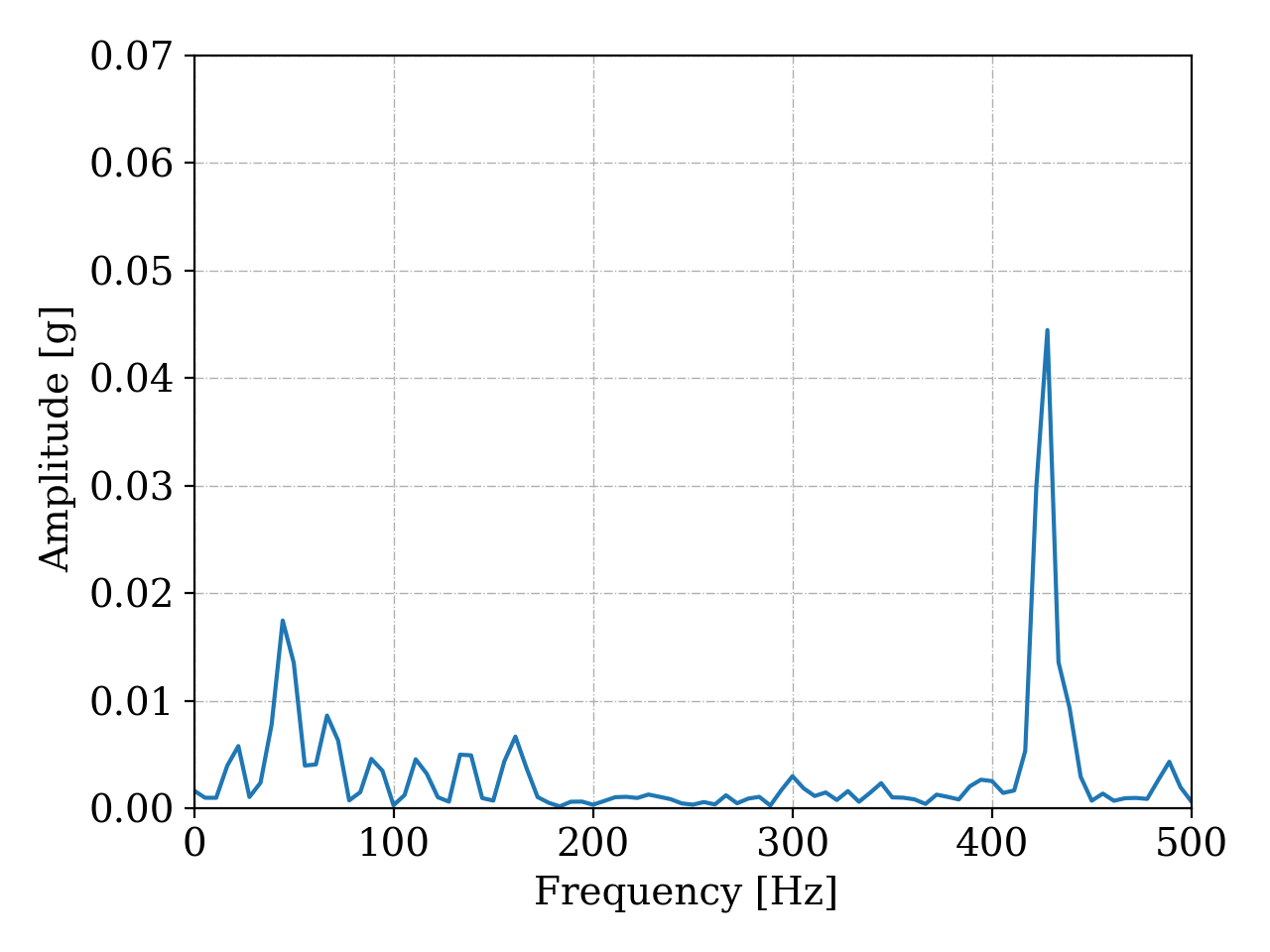}
	\end{minipage}}
   \hfill 	
  \subfloat[Synthetic - Gear Fault]{
	\begin{minipage}[c][0.65\width]{
	   0.3\textwidth}
	   \centering
       \label{fig:gear_gear_syn}
	   \includegraphics[width=1\textwidth]{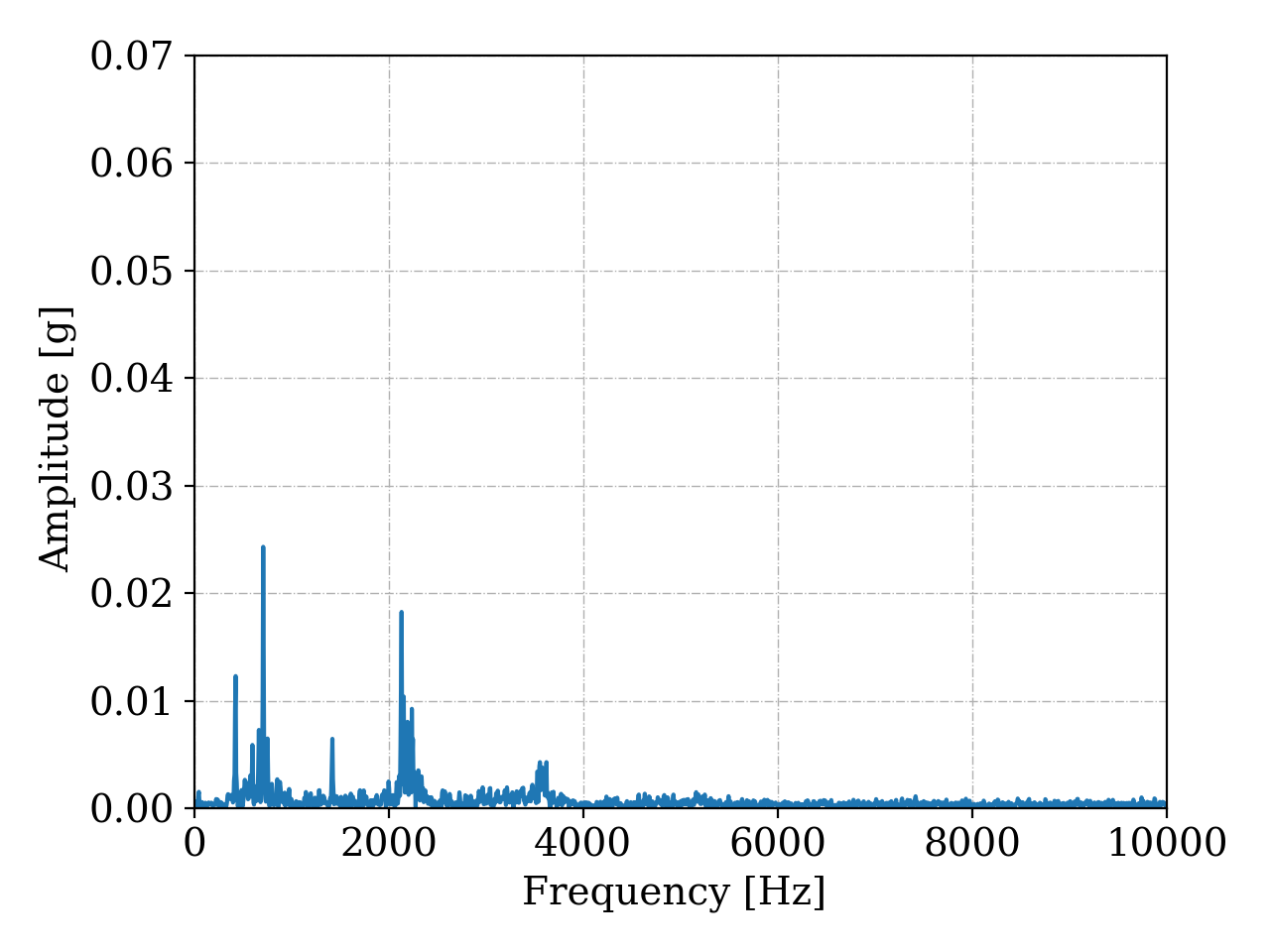}
	\end{minipage}}
    \caption{Examples of vibration signals for Case 2, real and synthetic.}
    \label{fig:sinalGearfaults}	
\end{figure}

In Fig. \ref{fig:sinalGearfaults} the original signals for the two conditions (Normal and Gear Fault) and the synthetic signals for the seven created conditions (Normal, BPFO, BPFI, Unbalance, Misalignment, Looseness and Gear Fault) from Case 2 are presented.

\renewcommand{\baselinestretch}{3} 
\renewcommand{\baselinestretch}{1.5} 

The real normal condition signal, Fig.\ref{fig:gear_original} shows evidence of 1 and 2 x GMF, which is normal when monitoring gearboxes. On the other hand, the original faulty signal, Fig.\ref{fig:gear_falha_original} already presents, in addition to 1 and 2 x GMF, harmonics in 3 and 4 x GMF, and side bands, characterizing the appearance and progression of the fault.

Regarding the synthetic signals, the same analysis performed for Case 1 is valid for Fig.\ref{fig:sinalGearfaults}, with the exception of Fig.\ref{fig:gear_gear_syn} which has GMF equal to 711 hz, being the real frequency of the gearbox. It can be noted that the behavior of the real and synthetic signals for each condition is similar, as proposed by the approach.

\subsubsection{Case 3 - Mechanical Faults}
\begin{figure}
  \subfloat[Real - Normal]{
	\begin{minipage}[c][0.65\width]{
	   0.3\textwidth}
	   \centering
       \label{fig:bancada_original}
	   \includegraphics[width=1\textwidth]{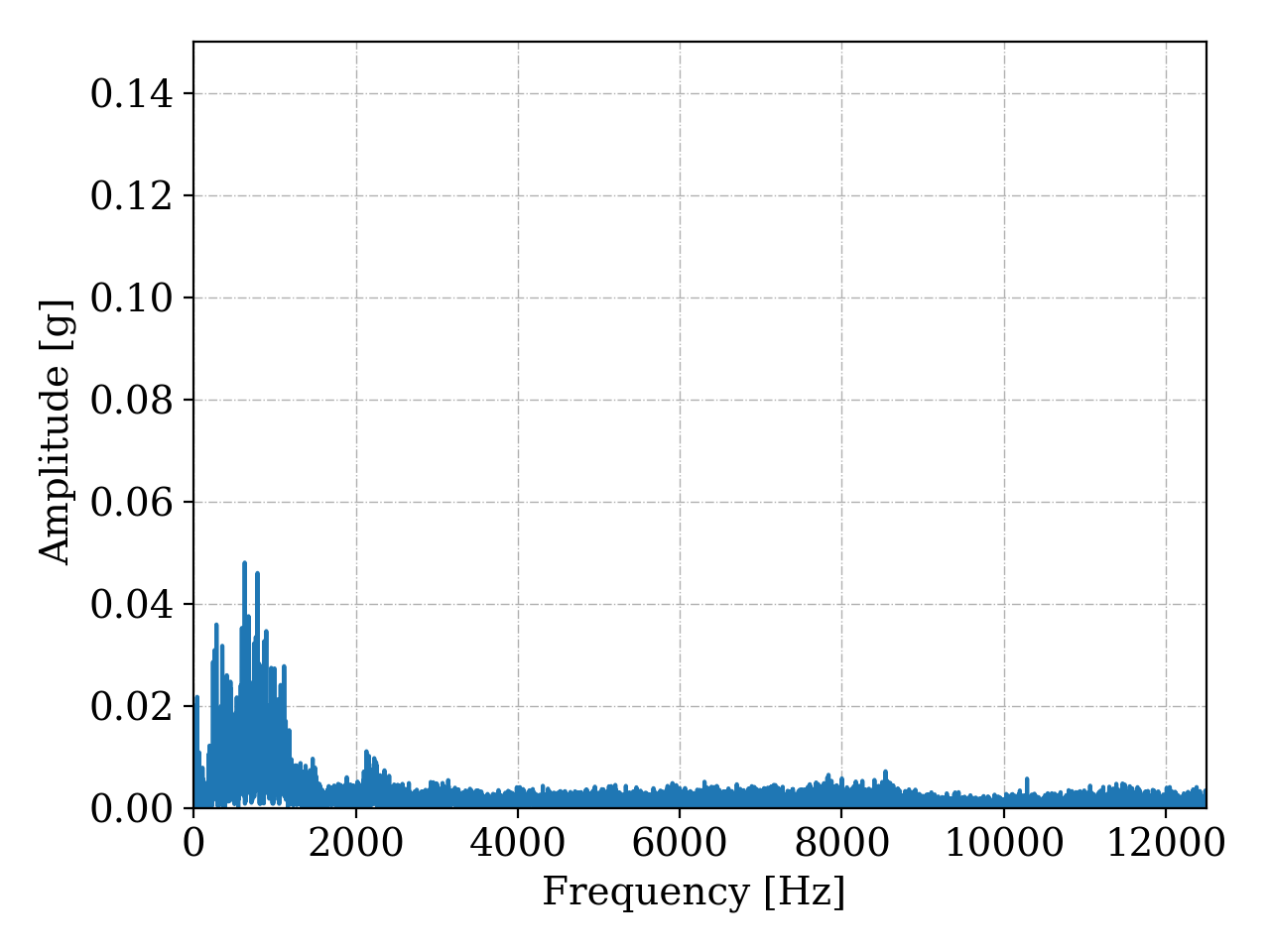}
	\end{minipage}}
 \hfill 
   \subfloat[Real - Unbalance]{
	\begin{minipage}[c][0.65\width]{
	   0.3\textwidth}
	   \centering
       \label{fig:bancada_original}
	   \includegraphics[width=1\textwidth]{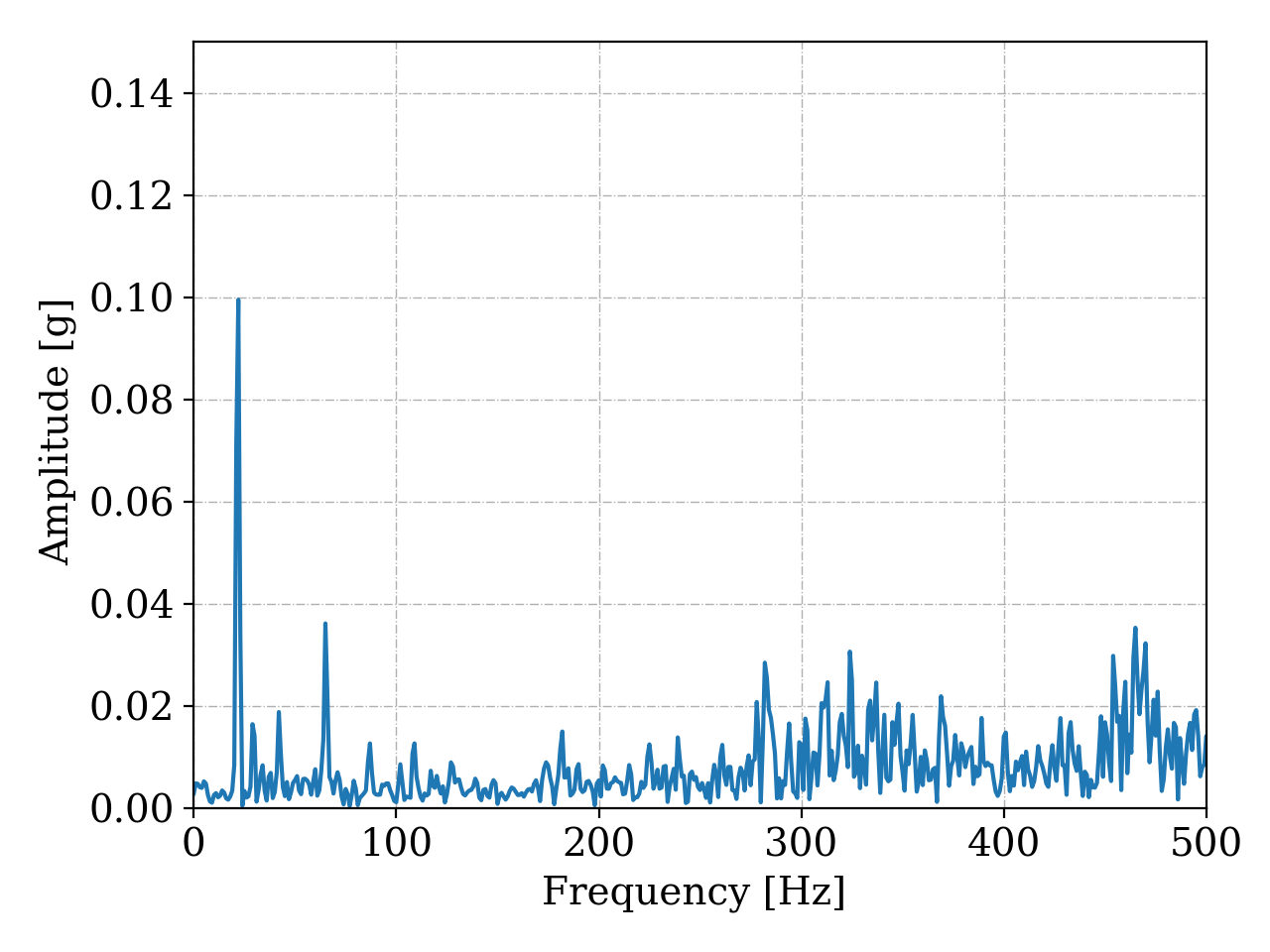}
	\end{minipage}}
 \hfill 
  \subfloat[Real - Misalignment]{
	\begin{minipage}[c][0.65\width]{
	   0.3\textwidth}
	   \centering
       \label{fig:bancada_original}
	   \includegraphics[width=1\textwidth]{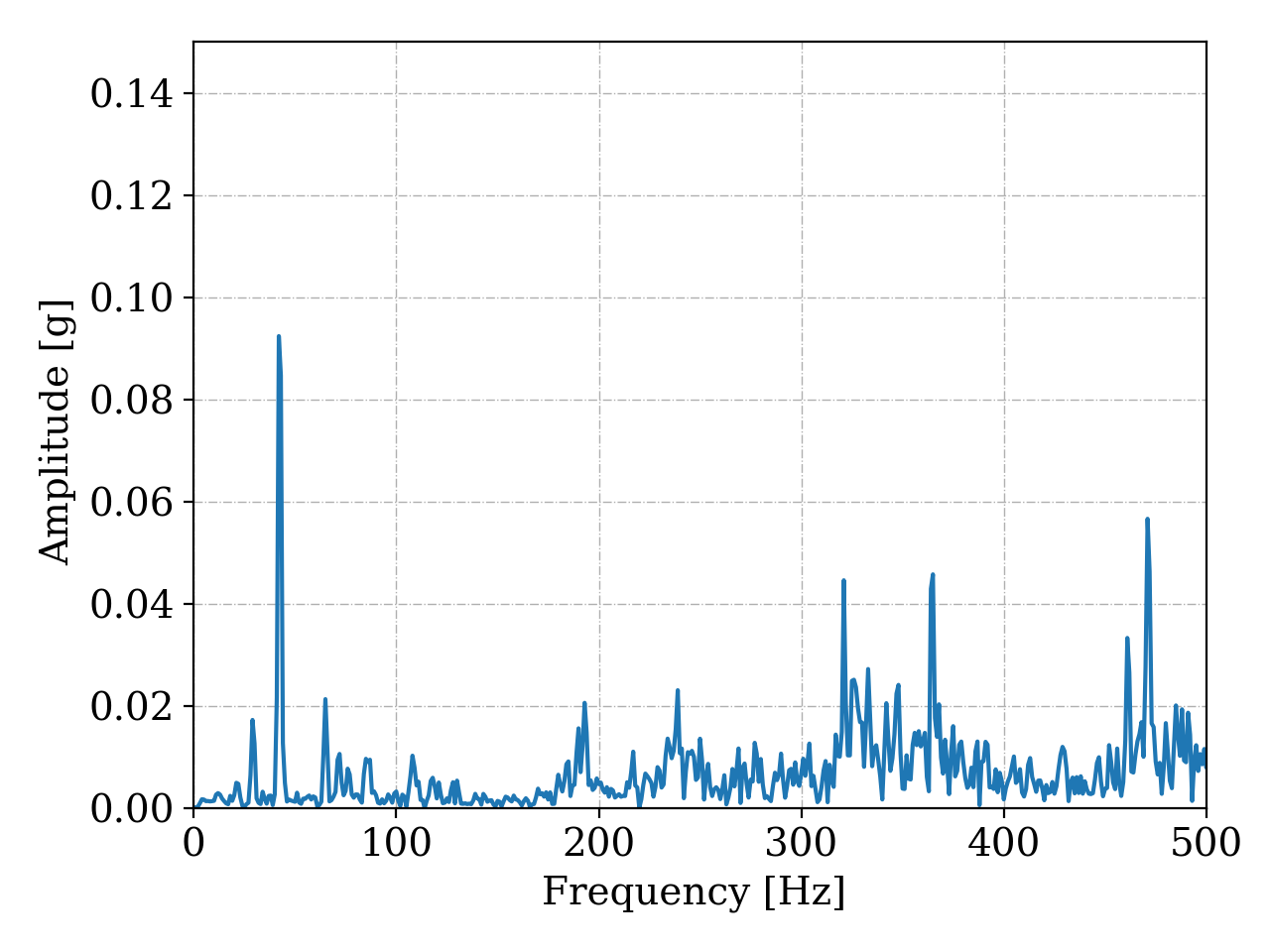}
	\end{minipage}}
 \hfill 
  \subfloat[Real - Looseness]{
	\begin{minipage}[c][0.65\width]{
	   0.3\textwidth}
	   \centering
        \label{fig:bancada_falha_original}
	   \includegraphics[width=1\textwidth]{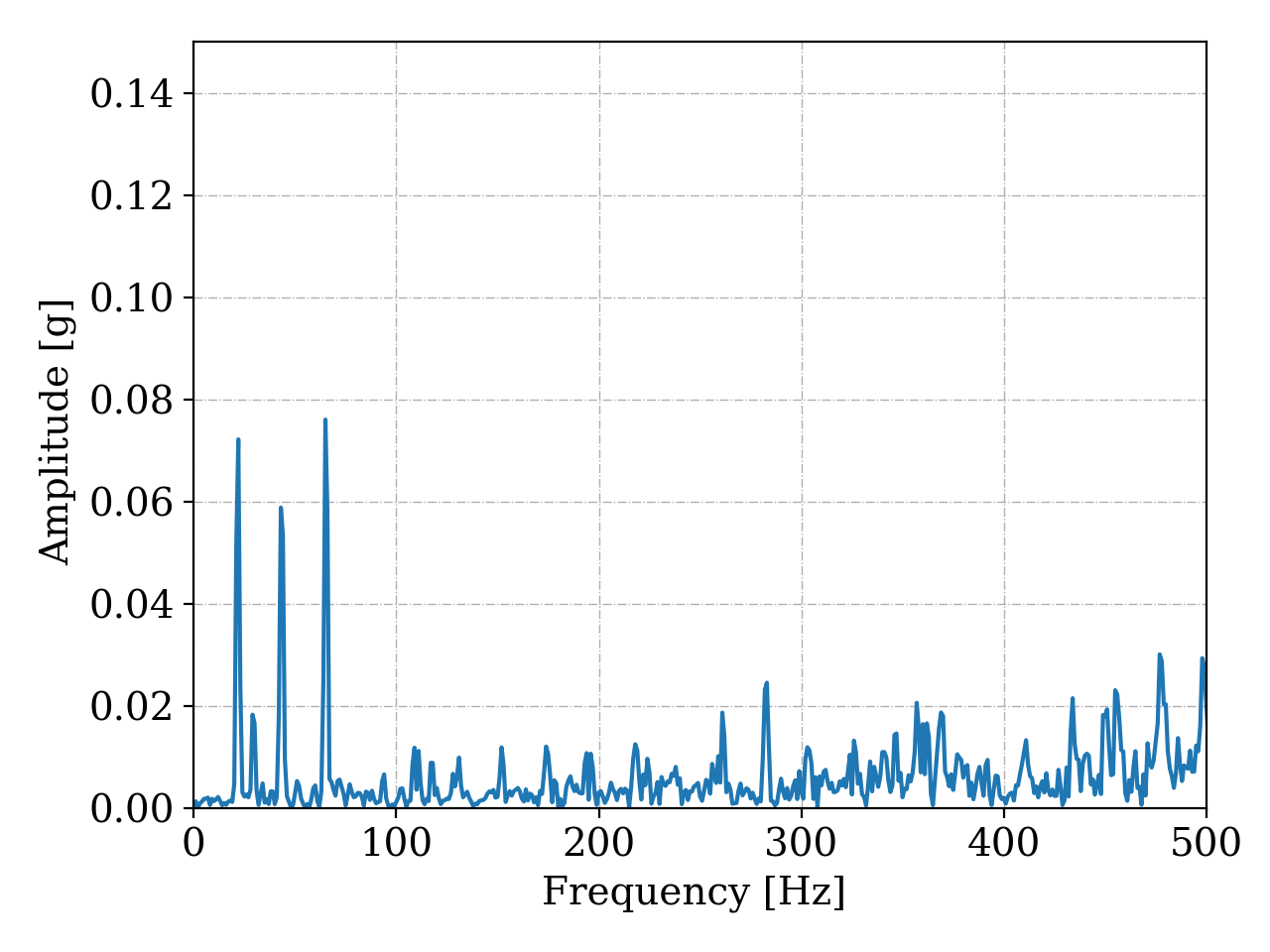}
	\end{minipage}}
 \hfill	
  \subfloat[Synthetic - Normal]{
	\begin{minipage}[c][0.65\width]{
	   0.3\textwidth}
	   \centering
	   \includegraphics[width=1\textwidth]{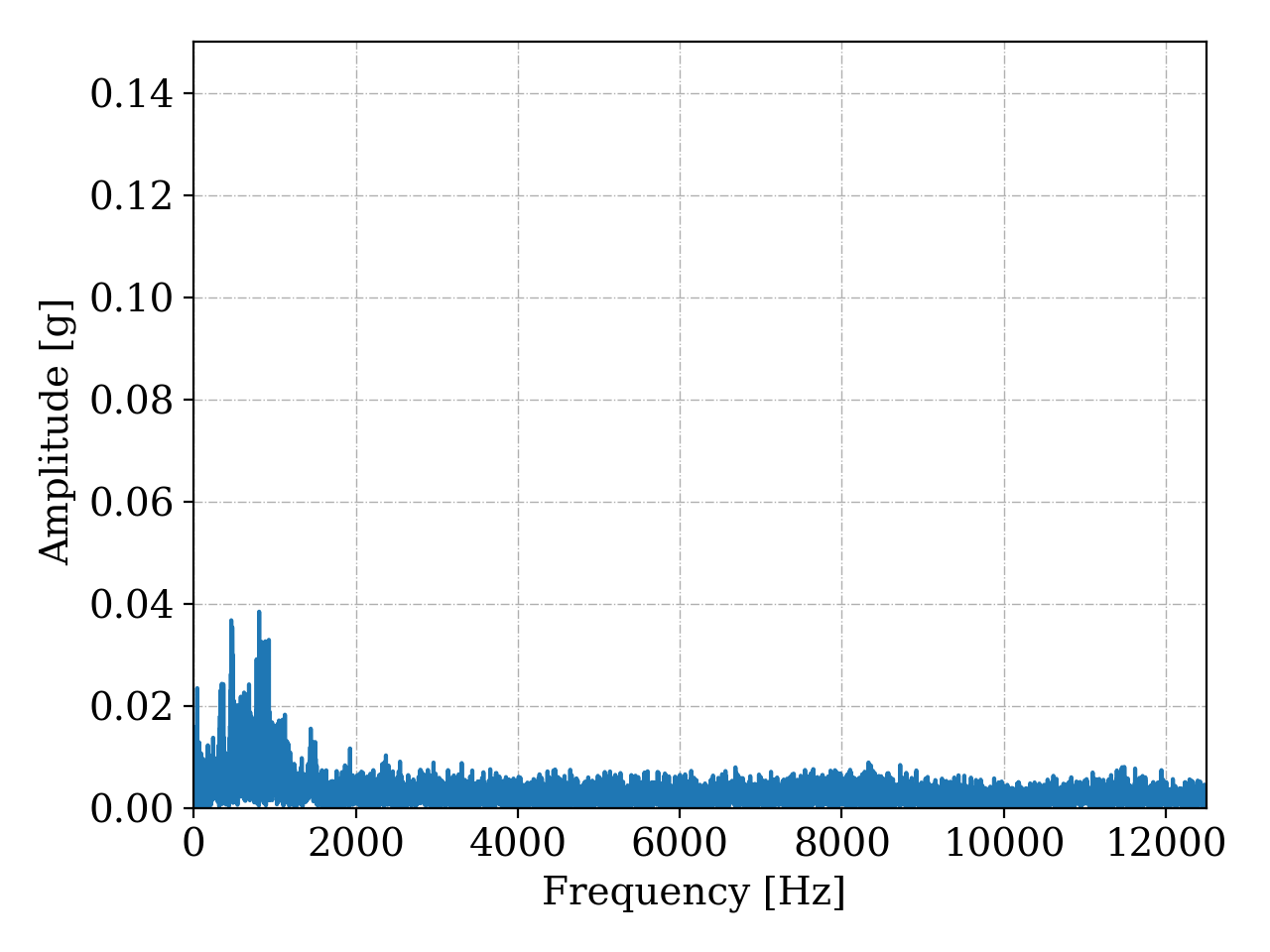}
	\end{minipage}}
 \hfill 
  \subfloat[Synthetic - BPFO]{
	\begin{minipage}[c][0.65\width]{
	   0.3\textwidth}
	   \centering
       \label{fig:bancada_bpfo_syn}
	   \includegraphics[width=1\textwidth]{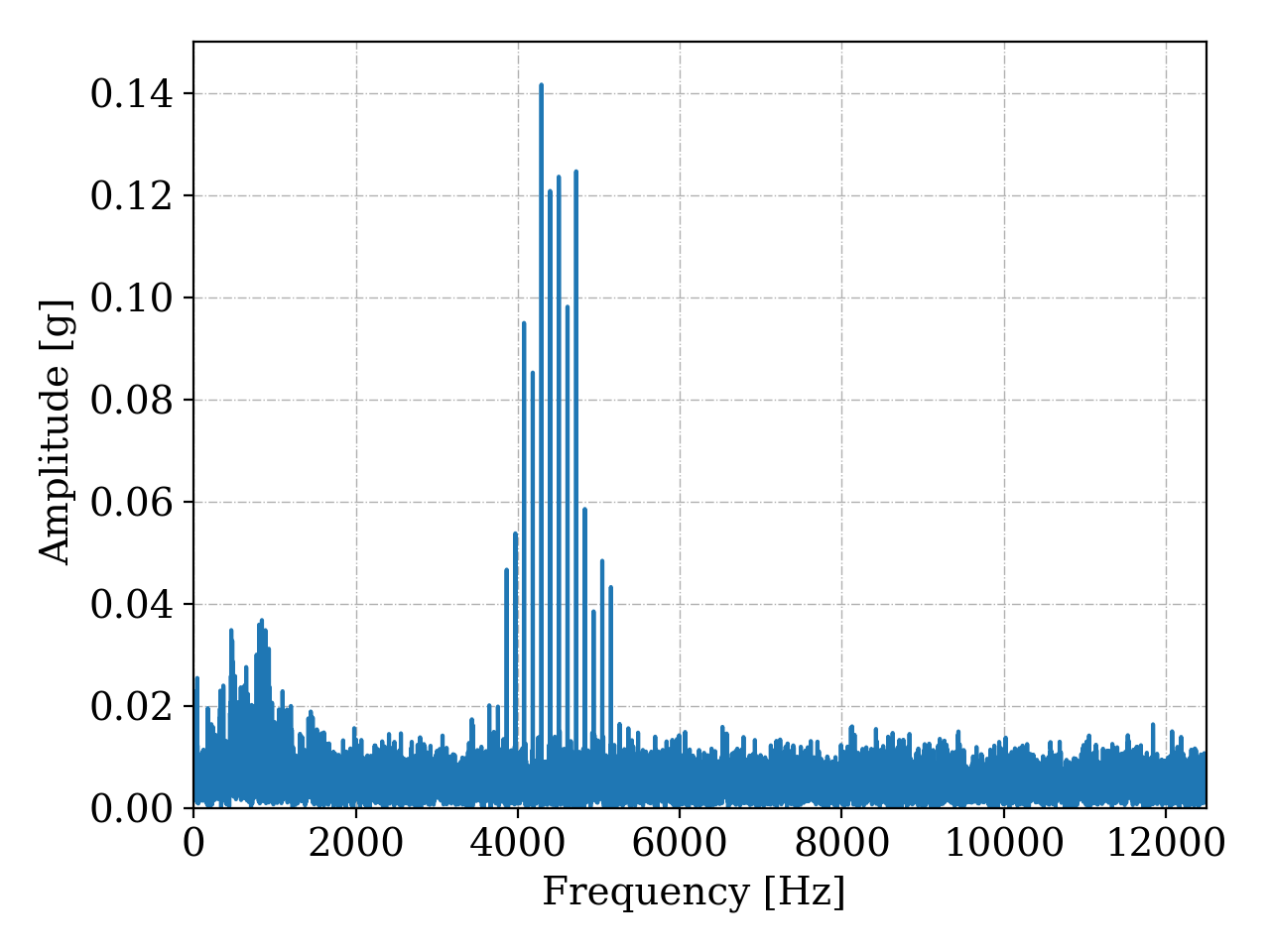}
	\end{minipage}}
 \hfill 	
  \subfloat[Synthetic - BPFI]{
	\begin{minipage}[c][0.65\width]{
	   0.3\textwidth}
	   \centering
       \label{fig:bancada_bpfi_syn}
	   \includegraphics[width=1\textwidth]{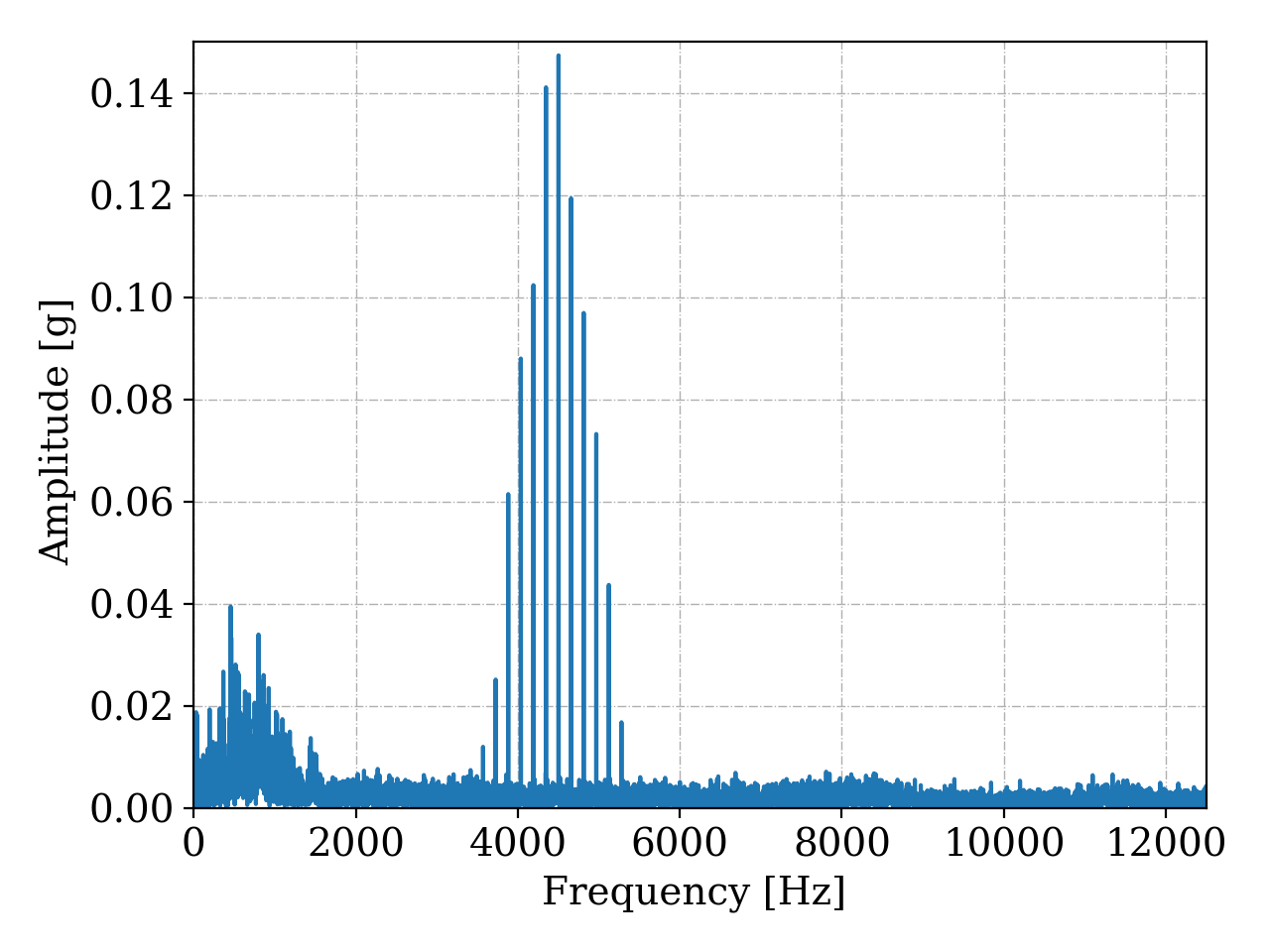}
	\end{minipage}}
  \hfill 	
  \subfloat[Synthetic - Unbalance]{
	\begin{minipage}[c][0.65\width]{
	   0.3\textwidth}
	   \centering
       \label{fig:bancada_unbalance_syn}
	   \includegraphics[width=1\textwidth]{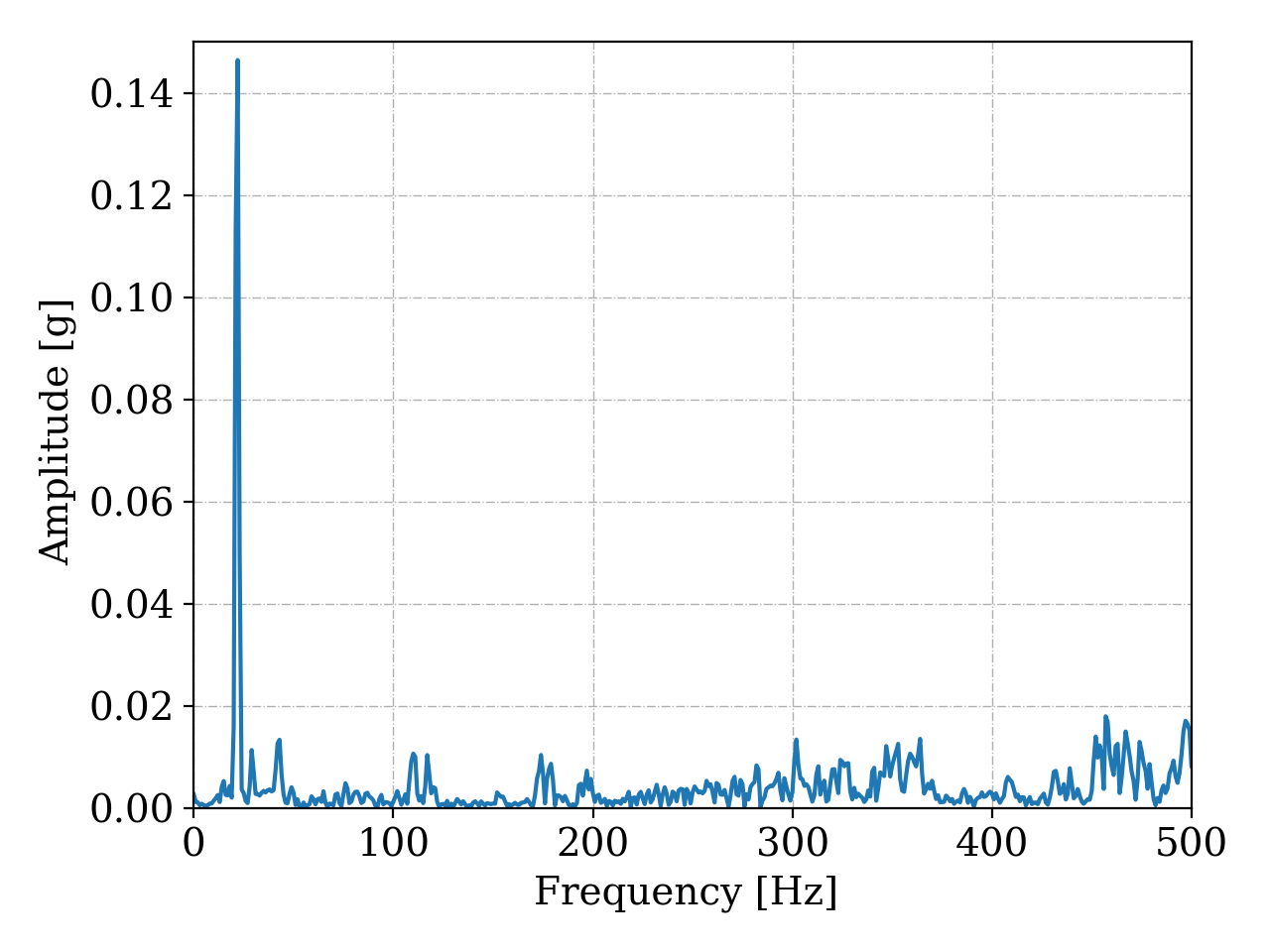}
	\end{minipage}}
   \hfill 	
  \subfloat[Synthetic - Misalignment]{
	\begin{minipage}[c][0.65\width]{
	   0.3\textwidth}
	   \centering
       \label{fig:bancada_mis_syn}
	   \includegraphics[width=1\textwidth]{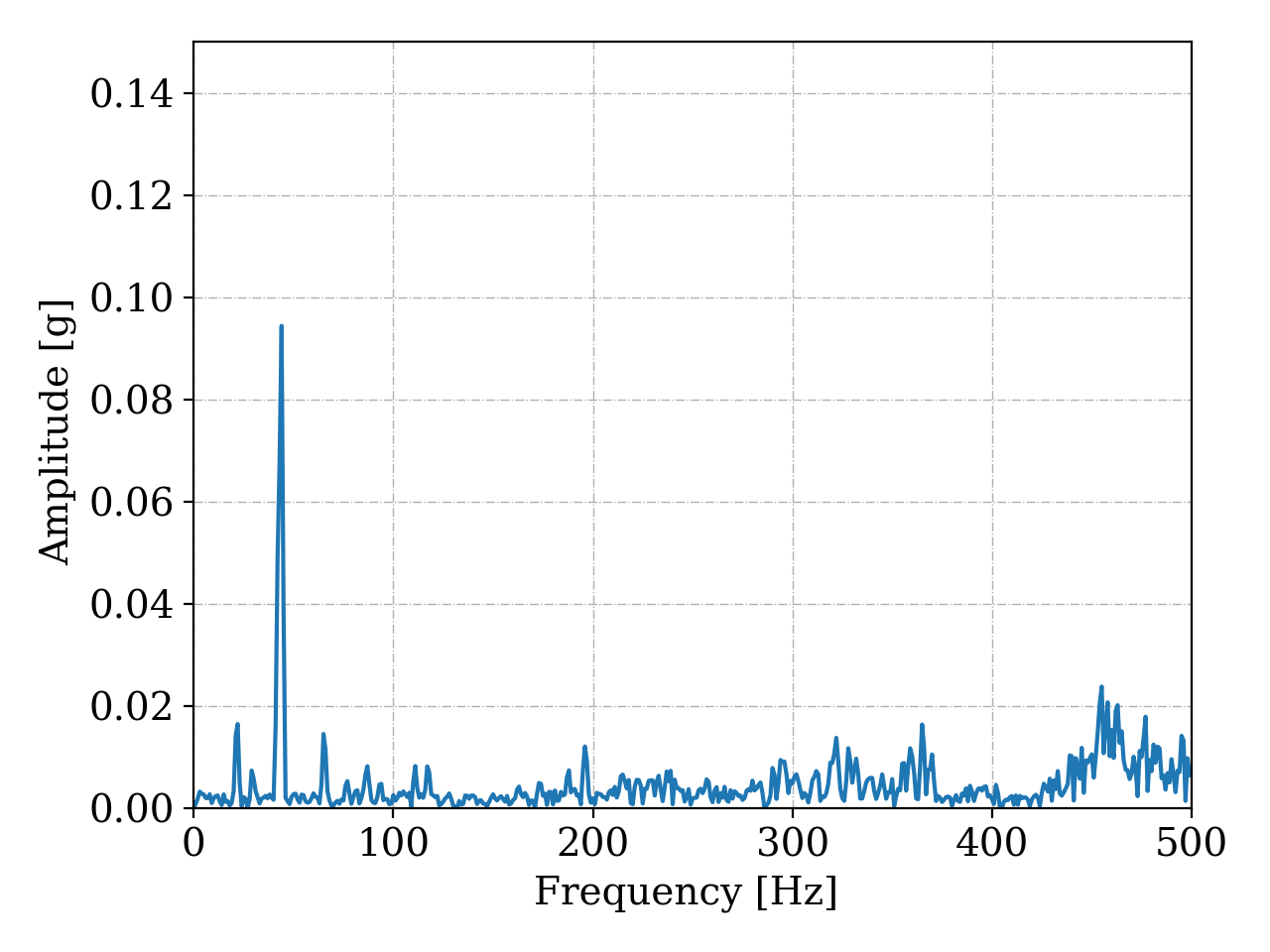}
	\end{minipage}}\\
   \hfill 	
  \subfloat[Synthetic - Looseness]{
	\begin{minipage}[c][0.5\width]{
	   0.5\textwidth}
	   \centering
       \label{fig:bancada_loos_syn}
	   \includegraphics[width=.63\textwidth]{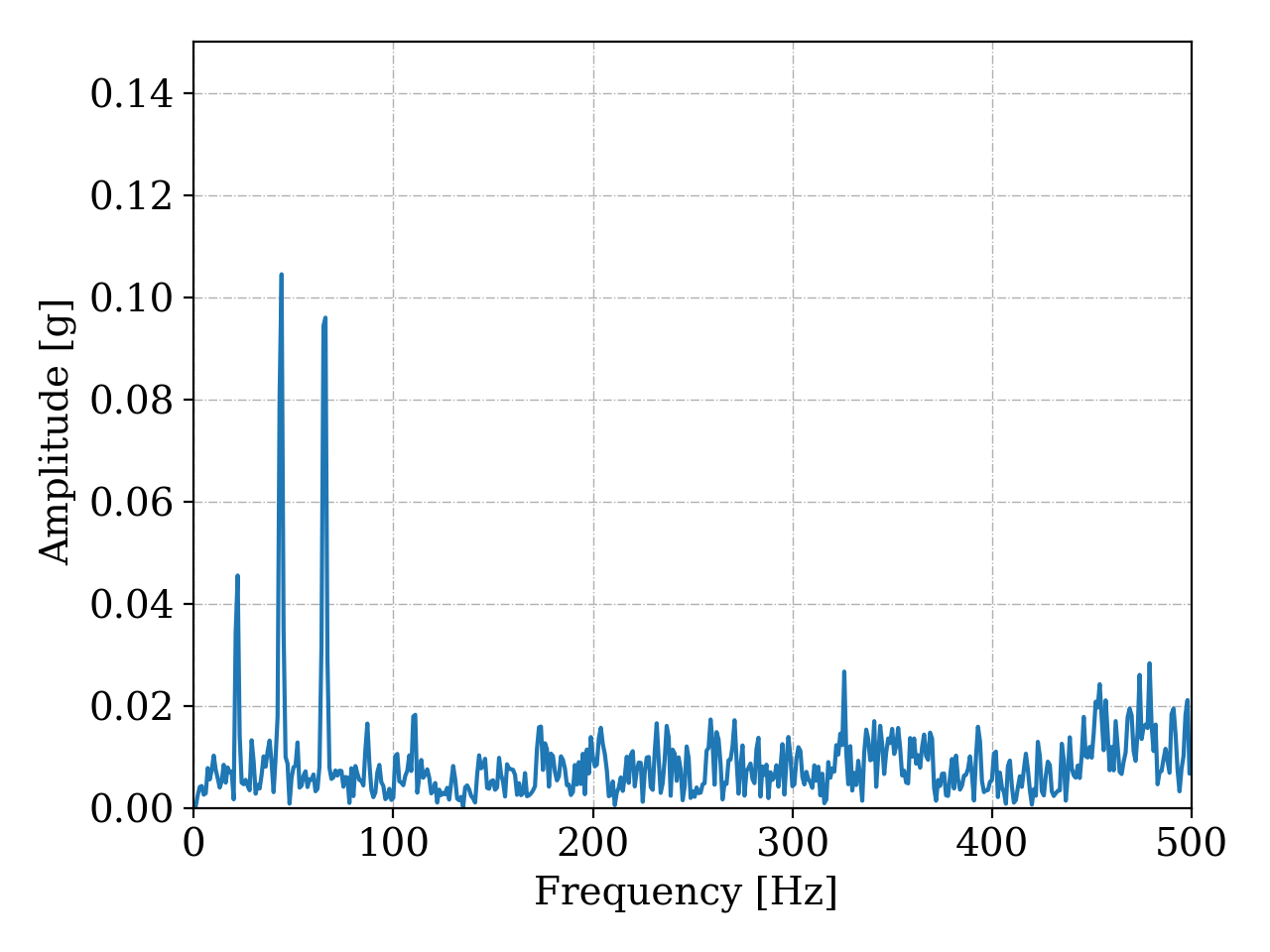}
	\end{minipage}}
   \hfill 	
  \subfloat[Synthetic - Gear Fault]{
	\begin{minipage}[c][0.5\width]{
	   0.5\textwidth}
	   \centering
       \label{fig:bancada_gear_syn}
	   \includegraphics[width=.63\textwidth]{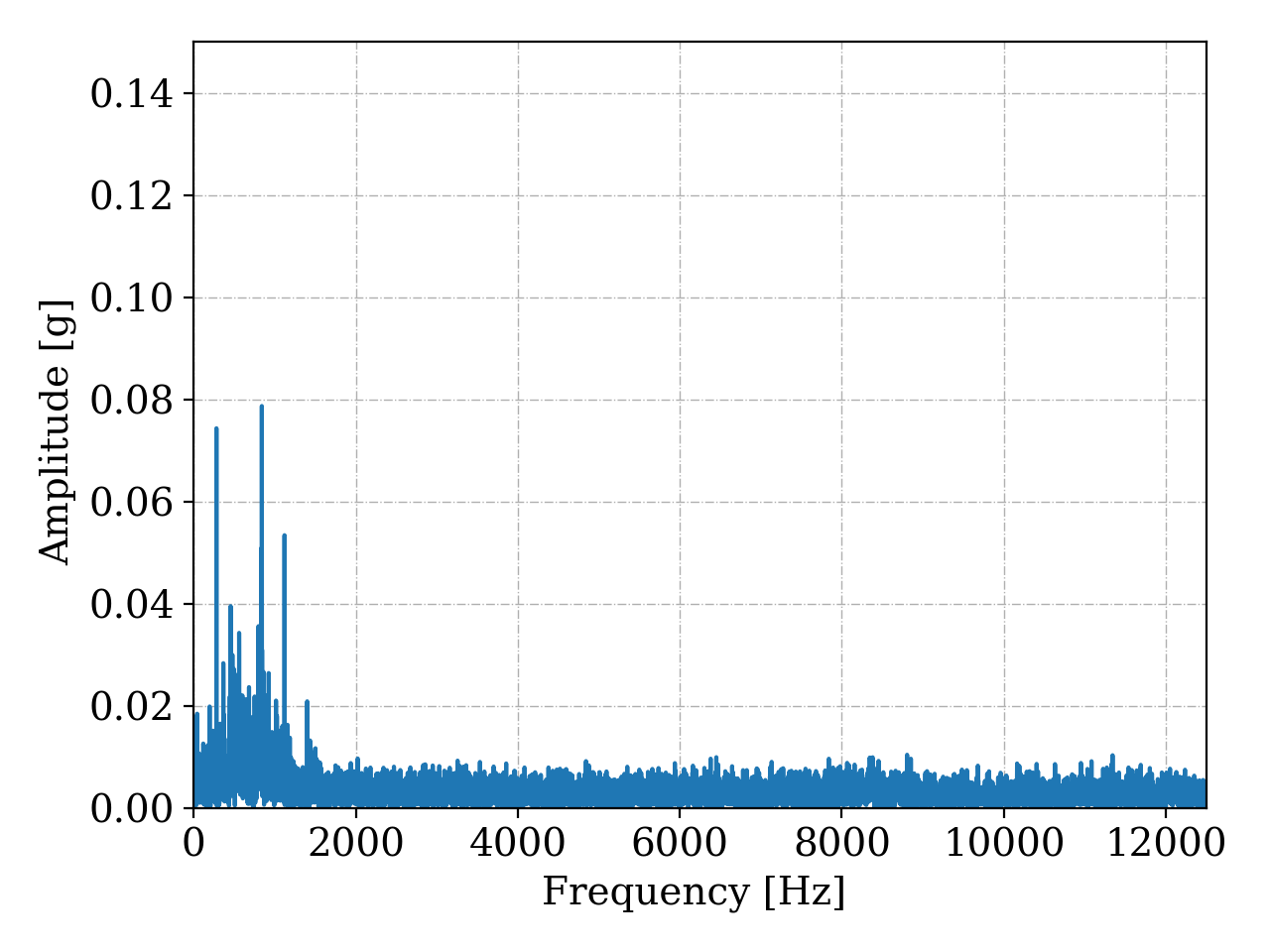}
	\end{minipage}}
    \caption{Examples of vibration signals for Case 3, real and synthetic.}
    \label{fig:sinalbancadafaults}	
\end{figure}

In Fig. \ref{fig:sinalbancadafaults} the original signals for the four conditions (Normal, Unbalance, Misalignment and Looseness) and the synthetic signals for the seven created conditions (Normal, BPFO, BPFI, Unbalance, Misalignment, Looseness and Gear Fault) from Case 3 are presented.

\renewcommand{\baselinestretch}{3} 
\renewcommand{\baselinestretch}{1.5} 

Note, in Fig. \ref{fig:sinalbancadafaults}, that even for the normal condition, there is evidence of highlighted frequencies up to 2000 hz, which may be related to lack of precision in the adjustment, assembly and lubrication of the bearings, improper adjustment (tensioning) and/or wear of the pulleys, noise generated by the frequency inverter etc. The condition was purposely chosen to simulate an industrial situation. It is known that, not all companies have enough specialized technicians to carry out an assembly and adjustment within the World Class Maintenance (WCM) standards. Also, there is a lack of specialized equipment to perform laser alignment, dynamic balancing etc. Moreover, due to the need for quick intervention to release the equipment for operation, maintenance cannot be performed with the desired precision. Because of these problems, the vibration signal can present a series of frequencies and/or noises, which make it difficult for the predictive maintenance specialist to analyze the main condition.

Even in a scenario where the signal is heavily polluted, technicians are usually able to perform an accurate diagnosis, based on field experience. Thus, it was decided to keep the bench in the presented condition, to make the test closer to the industrial reality, instead of completely adjusting the bench to eliminate noise and other sources of excitation that are not related to the defects. Allowing, to assess whether the proposed methodology is capable of imitating human behavior, accurately identifying the operating conditions of the asset.

Again, analysis performed for Case 1 and 2 in relation to synthetic data is valid for Fig. \ref{fig:sinalbancadafaults}, with the exception of Fig. \ref{fig:bancada_bpfo_syn}  and \ref{fig:bancada_bpfi_syn} being the fundamental frequency corresponding to the damaged element, BPFO = 107.09 hz and BPFI = 155.7 hz.

As in the other cases, it can be noted that the described fault characteristics are evident in all the synthetically generated signals and in the original signals. This shows that the data augmentation process was able to increase the diversity of the dataset without losing its fundamental characteristics.

\subsection{Number of total samples for training}

To evaluate the influence of the total number of signals in the training group, 10 sets were created containing: 1050, 2100, 3150, 4200, 5250, 6300, 7350, 8400, 9450, 10500 signals.

\begin{figure*}[ht]
  \centering
  \includegraphics[scale=0.55]{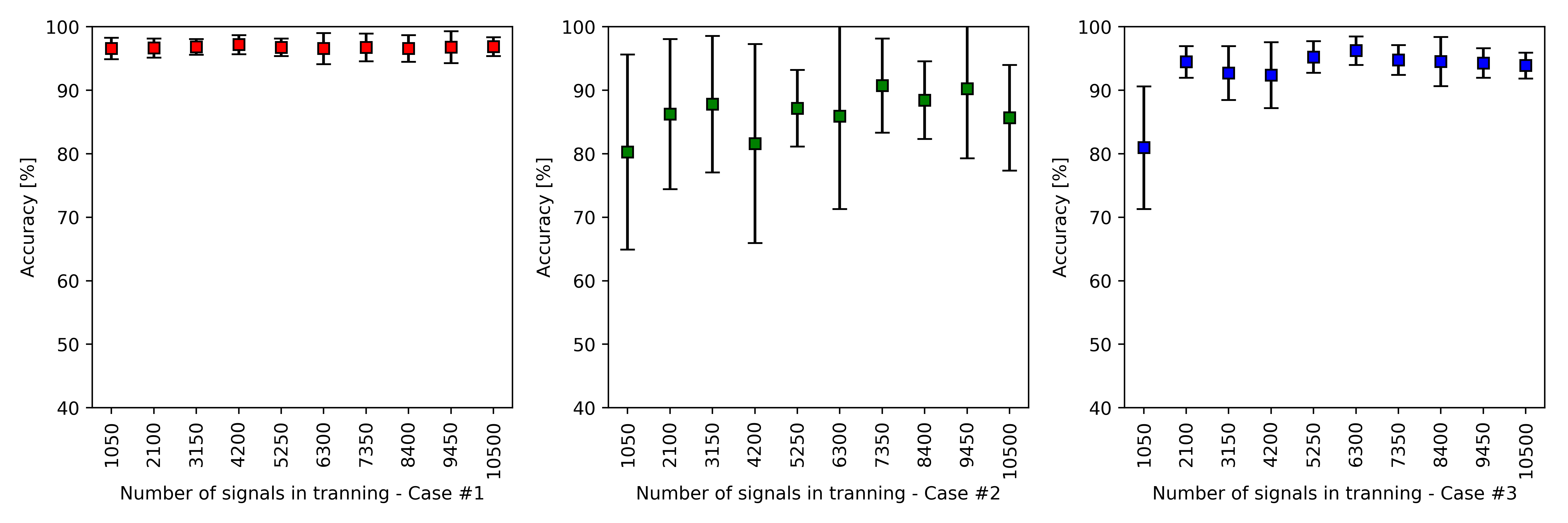}
  \caption{Results for each case varying the total amount of signals in the training set.}
  \label{fig:qta_sinais_MAX}
\end{figure*}

As the objective was not to evaluate the efficiency of the model, but to compare the result obtained in relation to the amount of signals in the training group, 30 original signals were selected to perform the data augmentation step, and the results for accuracy and standard deviation are shown in Fig.\ref{fig:qta_sinais_MAX}.

Analyzing Fig.\ref{fig:qta_sinais_MAX}, it can be noted that there is no significant variation in relation to the amount of data used, with the exception of Case 3 in the set of 1050 signals. Such variation may have occurred due to the greater complexity of the signal present in the dataset, requiring a greater amount of signals in training for learning. Note that there was no significant gain in the accuracy and stability of the model (less standard deviation) in increasing the number of signals above 5250 signals, and therefore, the amount was selected for the other analyses.

\subsection{Number of real samples for training}

Based on the previous analysis, 5250 signals were used in the training, aiming to balance the computational cost and the efficiency of the model.

Unlike the previous analysis, the number of real signals to generate the total amount of training was varied. The goal is to analyze whether, through zero or few data acquisitions, it is possible to train a model robust enough for fault diagnosis.

\begin{figure*}[ht]
  \centering
  \includegraphics[scale=0.52]{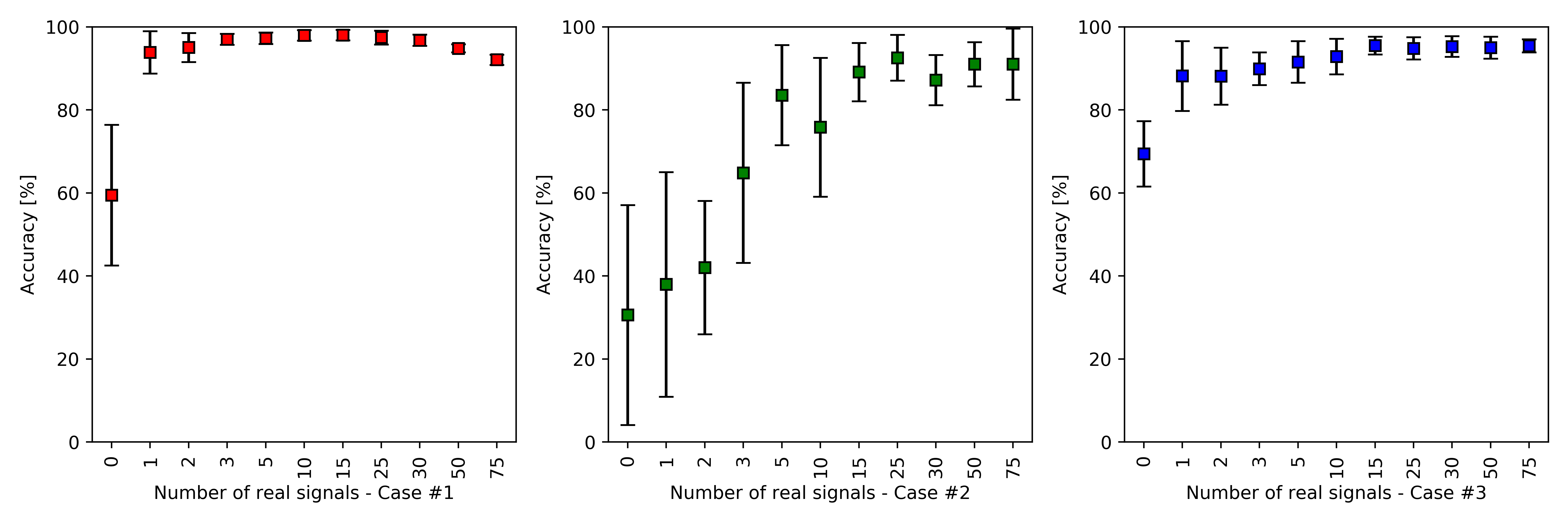}
  \caption{Results for each case varying the amount of real signals used to create the training set.}
  \label{fig:qta_sinais_com_zero}
\end{figure*}

11 conditions were tested, using 0, 1, 2, 3, 5, 10, 15, 25, 30, 50 and 75 real signals to generate the 5250 signals in the training. The values were selected so that the number of times the signals were augmented resulted in integer values, with no need to discard excess signals. As Case 2 has only 104 real signals per condition, the range was set from 1 to 75 signals. The results obtained for accuracy and standard deviation are shown in Fig.\ref{fig:qta_sinais_com_zero}.

Unlike the other conditions, in the first test, all signals for the training group were generated synthetically, without performing the combination with the real signal (zero number of real signal). It can be seen in Fig.\ref{fig:qta_sinais_com_zero}, that with zero real signal the lowest performing test was achieved. This result was expected, since the signals were created, basically containing the machine rotation frequency, white noise and random frequencies with amplitude close to the noise. This characteristic does not represent the dynamic behavior of the machine, resulting in the lowest value obtained. Therefore, it can be concluded that using the chosen strategy, it is not possible to train a model using only synthetic data without the combination with the real signal. It is worth mentioning that, despite not being the focus of the work, other strategies, such as more complex modeling, can be tested to overcome this problem.

Analyzing Fig.\ref{fig:qta_sinais_com_zero}, it can also be seen that using up to two real signals to create the entire training dataset, the average accuracy is significantly lower than the others, and the model has a large standard deviation. This occurs because if one of the selected signals presents some noise or excitation not related to the dynamic behavior of the machine, such pattern will be replicated for at least 50\% of the training set signals, leading the model to error.

The increase in the number of signals tends to contribute to the improvement in the performance of the model, although it does not mean a linearly increasing behavior. It is noted that, in general, there is a stabilization of accuracy and standard deviation after 15 signals. In view of the computational and acquisition cost, and the results obtained with the increase in the amount of real data, it is suggested to be between 15 and 30 signals, a good working range for FaultD-XAI.

The maximum accuracy value reached in Case 1 was 98.1\% and standard deviation of 1.2\% with 15 signals. For Case 2, the maximum accuracy value was 92.5\% and standard deviation of 5.5\% with 25 signals. Case 3 obtained the highest accuracy value with 15 signals, being 95.5\% and standard deviation of 2.1\%. Smaller values of standard deviation were obtained for other amounts of signals, however none of them had a difference greater than 0.5\% in relation to those presented.

The differences in accuracy and standard deviation obtained between each case are in accordance with what was expected in the study. Case 1 is a bearings remaining useful life test, where data were collected continuously, without the bench being dismantled to insert the defects. As there were no changes in the test bench, variations in the dynamic behavior of the machine are not expected, other than those caused by the wear of the components (since it is isolated). This makes the fault identification simpler for the model.

On the other hand, in Case 2 and 3, the benches were dismantled to insert the faults. Therefore, small changes in relation to the real signal used as a reference for generating the synthetic data may occur during assembly and adjustment, generating a reduction in accuracy. This condition represents a situation where the system was trained with a machine signal in a similar condition, but not identical to the current one (which is a transfer learning problem definition).

The two analyzes are extremely interesting, and show the applicability of the system in the industrial scenario. In Case 1, it addresses the monitoring of a machine, where the system was installed, trained, and over time was able to identify a fault in the asset. In Case 2 and 3, the model was initially installed and trained. Hypothetically, the machine underwent some corrective maintenance, and the same model, without being retrained, continued to monitor the asset, and was able to identify the faults present. The fact that there is no need to retrain the model for small changes in the asset or reestablish the operating condition through corrective maintenance, overcomes a major problem in applications that is retraining. It is worth mentioning that for major interventions or changes in projects that modify the dynamic behavior of the machine, the model will tend to reduce its accuracy if not retraining. 

The confusion matrix for the best values obtained in Case 1 is presented, containing the number of signals per class and the standard deviation for the 10 tests, in addition to the value in percentage, Table~\ref{tab:confusionmatrix_1}. Where in the column are the True Labels and in the row are the Predicted Labels. 

\renewcommand{\baselinestretch}{1.3} 
\begin{table*}[ht]
   \caption{}
   \caption*{Confusion Matrix - Case 1}
   \label{tab:confusionmatrix_1}
   \resizebox{\textwidth}{!}{\begin{tabular}{cccccccc}
     \toprule
    \textbf{Case 1} & Normal & BPFO & BPFI & Unbalance & Misalignment & Looseness & Gear Fault  \\
     \midrule
     \multirow{2}{*}{Normal}  
        & \multicolumn{1}{c}{\textcolor{green}{52.01 \% (0.51\%)}} & \multicolumn{1}{c}{\textcolor{red}{0.72 \% (0.41\%)}} & \multicolumn{1}{c}{{0 \% (0\%)}} 
        & \multicolumn{1}{c}{{0 \% (0\%) }} & \multicolumn{1}{c}{\textcolor{red}{0.41 \% (0.51\%) }} &
         \multicolumn{1}{c}{\textcolor{red}{0.10 \% (0.10\%)}}
        & \multicolumn{1}{c}{{0 \% (0\%)}}  \\
        & \multicolumn{1}{c}{\textcolor{green}{504 (5) }} & \multicolumn{1}{c}{\textcolor{red}{7 (4)}} & \multicolumn{1}{c}{{0 (0) }} & \multicolumn{1}{c}{{0 (0)}} & \multicolumn{1}{c}{\textcolor{red}{4 (5)}} & \multicolumn{1}{c}{\textcolor{red}{1 (1) }} & \multicolumn{1}{c}{{0 (0)}} \\

     \multirow{2}{*}{BPFO}  
        & \multicolumn{1}{c}{\textcolor{red}{0.62 \% (0.92\%)}} & \multicolumn{1}{c}{\textcolor{green}{46.03 \% (0.92\%)}} & \multicolumn{1}{c}{{0 \% (0\%)}} 
        & \multicolumn{1}{c}{{0 \% (0\%) }} & \multicolumn{1}{c}{{0 \% (0\%) }} &
         \multicolumn{1}{c}{\textcolor{red}{0.10 \% (0.10\%)}}
        & \multicolumn{1}{c}{{0 \% (0\%)}}  \\
        & \multicolumn{1}{c}{\textcolor{red}{6 (9) }} & \multicolumn{1}{c}{\textcolor{green}{446 (9)}} & \multicolumn{1}{c}{{0 (0) }} & \multicolumn{1}{c}{{0 (0)}} & \multicolumn{1}{c}{{0 (0)}} & \multicolumn{1}{c}{\textcolor{red}{1 (1) }} & \multicolumn{1}{c}{{0 (0)}} \\

     \multirow{2}{*}{BPFI}  
        & \multicolumn{1}{c}{{0 \% (0\%)}} & \multicolumn{1}{c}{{0 \% (0\%)}} & \multicolumn{1}{c}{\textcolor{green}{0 \% (0\%)}} 
        & \multicolumn{1}{c}{{0 \% (0\%) }} & \multicolumn{1}{c}{{0 \% (0\%) }} &
         \multicolumn{1}{c}{{0 \% (0\%)}}
        & \multicolumn{1}{c}{{0 \% (0\%)}}  \\
        & \multicolumn{1}{c}{{0 (0) }} & \multicolumn{1}{c}{{0 (0)}} & \multicolumn{1}{c}{\textcolor{green}{0 (0) }} & \multicolumn{1}{c}{{0 (0)}} & \multicolumn{1}{c}{{0 (0)}} & \multicolumn{1}{c}{{0 (0) }} & \multicolumn{1}{c}{{0 (0)}} \\

     \multirow{2}{*}{Unbalance}  
        & \multicolumn{1}{c}{{0 \% (0\%)}} & \multicolumn{1}{c}{{0 \% (0\%)}} & \multicolumn{1}{c}{{0 \% (0\%)}} 
        & \multicolumn{1}{c}{\textcolor{green}{0 \% (0\%) }} & \multicolumn{1}{c}{{0 \% (0\%) }} &
         \multicolumn{1}{c}{{0 \% (0\%)}}
        & \multicolumn{1}{c}{{0 \% (0\%)}}  \\
        & \multicolumn{1}{c}{{0 (0) }} & \multicolumn{1}{c}{{0 (0)}} & \multicolumn{1}{c}{{0 (0) }} & \multicolumn{1}{c}{\textcolor{green}{0 (0)}} & \multicolumn{1}{c}{{0 (0)}} & \multicolumn{1}{c}{{0 (0) }} & \multicolumn{1}{c}{{0 (0)}} \\

     \multirow{2}{*}{Misalignment}  
        & \multicolumn{1}{c}{{0 \% (0\%)}} & \multicolumn{1}{c}{{0 \% (0\%)}} & \multicolumn{1}{c}{{0 \% (0\%)}} 
        & \multicolumn{1}{c}{{0 \% (0\%) }} & \multicolumn{1}{c}{\textcolor{green}{0 \% (0\%) }} &
         \multicolumn{1}{c}{{0 \% (0\%)}}
        & \multicolumn{1}{c}{{0 \% (0\%)}}  \\
        & \multicolumn{1}{c}{{0 (0) }} & \multicolumn{1}{c}{{0 (0)}} & \multicolumn{1}{c}{{0 (0) }} & \multicolumn{1}{c}{{0 (0)}} & \multicolumn{1}{c}{\textcolor{green}{0 (0)}} & \multicolumn{1}{c}{{0 (0) }} & \multicolumn{1}{c}{{0 (0)}} \\

     \multirow{2}{*}{Looseness}  
        & \multicolumn{1}{c}{{0 \% (0\%)}} & \multicolumn{1}{c}{{0 \% (0\%)}} & \multicolumn{1}{c}{{0 \% (0\%)}} 
        & \multicolumn{1}{c}{{0 \% (0\%) }} & \multicolumn{1}{c}{{0 \% (0\%) }} &
         \multicolumn{1}{c}{\textcolor{green}{0 \% (0\%)}}
        & \multicolumn{1}{c}{{0 \% (0\%)}}  \\
        & \multicolumn{1}{c}{{0 (0) }} & \multicolumn{1}{c}{{0 (0)}} & \multicolumn{1}{c}{{0 (0) }} & \multicolumn{1}{c}{{0 (0)}} & \multicolumn{1}{c}{{0 (0)}} & \multicolumn{1}{c}{\textcolor{green}{0 (0) }} & \multicolumn{1}{c}{{0 (0)}} \\

     \multirow{2}{*}{Gear Fault}  
        & \multicolumn{1}{c}{{0 \% (0\%)}} & \multicolumn{1}{c}{{0 \% (0\%)}} & \multicolumn{1}{c}{{0 \% (0\%)}} 
        & \multicolumn{1}{c}{{0 \% (0\%) }} & \multicolumn{1}{c}{{0 \% (0\%) }} &
         \multicolumn{1}{c}{{0 \% (0\%)}}
        & \multicolumn{1}{c}{\textcolor{green}{0 \% (0\%)}}  \\
        & \multicolumn{1}{c}{{0 (0) }} & \multicolumn{1}{c}{{0 (0)}} & \multicolumn{1}{c}{{0 (0) }} & \multicolumn{1}{c}{{0 (0)}} & \multicolumn{1}{c}{{0 (0)}} & \multicolumn{1}{c}{{0 (0) }} & \multicolumn{1}{c}{\textcolor{green}{0 (0)}} \\

     \bottomrule    
   \end{tabular}}
\end{table*}
\renewcommand{\baselinestretch}{1.5} 

In Table~\ref{tab:confusionmatrix_1}, it can be seen that the model confused normal signals with BPFO, misalignment and looseness. On the other hand, the faulted signals (BPFO) were confused with the normal condition and looseness only. The confusion between normal and BPFO may still be acceptable, due to the run to failure test feature and manual label. Therefore, signals referring to the beginning of the fault may present similar characteristics. The other classifications were model errors. Analyzing the result only by class, the normal signals were classified correctly 97.67\% of the time, and faulty signals (BPFO) 98.45\%. The confusion matrix for Case 2 is shown in Table~\ref{tab:confusionmatrix_2}. In Case 2, Table~\ref{tab:confusionmatrix_2}, normal condition and gear fault were confused with unbalance, misalignment, looseness and gear defect. As the gear fault characteristics are different from the errors presented, it can be concluded that there is a model error, despite the low amount. Analyzing by class, normal signals were classified correctly 91.14\% of the time, and faulty signals 92.67\%. The confusion matrix for Case 3 is shown in Table~\ref{tab:confusionmatrix_3}.

\renewcommand{\baselinestretch}{1.3} 
\begin{table*}[ht]
   \caption{}
   \caption*{Confusion Matrix - Case 2}
   \label{tab:confusionmatrix_2}
   \resizebox{\textwidth}{!}{\begin{tabular}{cccccccc}
     \toprule
    \textbf{Case 2} & Normal & BPFO & BPFI & Unbalance & Misalignment & Looseness & Gear Fault  \\
     \midrule
     \multirow{2}{*}{Normal}  
        & \multicolumn{1}{c}{\textcolor{green}{7.90 \% (0.98\%)}} & \multicolumn{1}{c}{{0 \% (0\%)}} & \multicolumn{1}{c}{{0 \% (0\%)}} 
        & \multicolumn{1}{c}{\textcolor{red}{0.11 \% (0.11\%) }} & \multicolumn{1}{c}{\textcolor{red}{0.11 \% (0.11\%) }} &
         \multicolumn{1}{c}{\textcolor{red}{0.11 \% (0.11\%)}}
        & \multicolumn{1}{c}{\textcolor{red}{0.44 \% (0.76\%)}}  \\
        & \multicolumn{1}{c}{\textcolor{green}{72 (9) }} & \multicolumn{1}{c}{{0 (0)}} & \multicolumn{1}{c}{{0 (0) }} & \multicolumn{1}{c}{\textcolor{red}{1 (1)}} & \multicolumn{1}{c}{\textcolor{red}{1 (1)}} & \multicolumn{1}{c}{\textcolor{red}{1 (1) }} & \multicolumn{1}{c}{\textcolor{red}{4 (7)}} \\

     \multirow{2}{*}{BPFO}  
        & \multicolumn{1}{c}{{0 \% (0\%)}} & \multicolumn{1}{c}{\textcolor{green}{0 \% (0\%)}} & \multicolumn{1}{c}{{0 \% (0\%)}} 
        & \multicolumn{1}{c}{{0 \% (0\%) }} & \multicolumn{1}{c}{{0 \% (0\%) }} &
         \multicolumn{1}{c}{{0 \% (0\%)}}
        & \multicolumn{1}{c}{{0 \% (0\%)}}  \\
        & \multicolumn{1}{c}{{0 (0) }} & \multicolumn{1}{c}{\textcolor{green}{0 (0)}} & \multicolumn{1}{c}{{0 (0) }} & \multicolumn{1}{c}{{0 (0)}} & \multicolumn{1}{c}{{0 (0)}} & \multicolumn{1}{c}{{0 (0) }} & \multicolumn{1}{c}{{0 (0)}} \\

     \multirow{2}{*}{BPFI}  
        & \multicolumn{1}{c}{{0 \% (0\%)}} & \multicolumn{1}{c}{{0 \% (0\%)}} & \multicolumn{1}{c}{\textcolor{green}{0 \% (0\%)}} 
        & \multicolumn{1}{c}{{0 \% (0\%) }} & \multicolumn{1}{c}{{0 \% (0\%) }} &
         \multicolumn{1}{c}{{0 \% (0\%)}}
        & \multicolumn{1}{c}{{0 \% (0\%)}}  \\
        & \multicolumn{1}{c}{{0 (0) }} & \multicolumn{1}{c}{{0 (0)}} & \multicolumn{1}{c}{\textcolor{green}{0 (0) }} & \multicolumn{1}{c}{{0 (0)}} & \multicolumn{1}{c}{{0 (0)}} & \multicolumn{1}{c}{{0 (0) }} & \multicolumn{1}{c}{{0 (0)}} \\

     \multirow{2}{*}{Unbalance}  
        & \multicolumn{1}{c}{{0 \% (0\%)}} & \multicolumn{1}{c}{{0 \% (0\%)}} & \multicolumn{1}{c}{{0 \% (0\%)}} 
        & \multicolumn{1}{c}{\textcolor{green}{0 \% (0\%) }} & \multicolumn{1}{c}{{0 \% (0\%) }} &
         \multicolumn{1}{c}{{0 \% (0\%)}}
        & \multicolumn{1}{c}{{0 \% (0\%)}}  \\
        & \multicolumn{1}{c}{{0 (0) }} & \multicolumn{1}{c}{{0 (0)}} & \multicolumn{1}{c}{{0 (0) }} & \multicolumn{1}{c}{\textcolor{green}{0 (0)}} & \multicolumn{1}{c}{{0 (0)}} & \multicolumn{1}{c}{{0 (0) }} & \multicolumn{1}{c}{{0 (0)}} \\

     \multirow{2}{*}{Misalignment}  
        & \multicolumn{1}{c}{{0 \% (0\%)}} & \multicolumn{1}{c}{{0 \% (0\%)}} & \multicolumn{1}{c}{{0 \% (0\%)}} 
        & \multicolumn{1}{c}{{0 \% (0\%) }} & \multicolumn{1}{c}{\textcolor{green}{0 \% (0\%) }} &
         \multicolumn{1}{c}{{0 \% (0\%)}}
        & \multicolumn{1}{c}{{0 \% (0\%)}}  \\
        & \multicolumn{1}{c}{{0 (0) }} & \multicolumn{1}{c}{{0 (0)}} & \multicolumn{1}{c}{{0 (0) }} & \multicolumn{1}{c}{{0 (0)}} & \multicolumn{1}{c}{\textcolor{green}{0 (0)}} & \multicolumn{1}{c}{{0 (0) }} & \multicolumn{1}{c}{{0 (0)}} \\

     \multirow{2}{*}{Looseness}  
        & \multicolumn{1}{c}{{0 \% (0\%)}} & \multicolumn{1}{c}{{0 \% (0\%)}} & \multicolumn{1}{c}{{0 \% (0\%)}} 
        & \multicolumn{1}{c}{{0 \% (0\%) }} & \multicolumn{1}{c}{{0 \% (0\%) }} &
         \multicolumn{1}{c}{\textcolor{green}{0 \% (0\%)}}
        & \multicolumn{1}{c}{{0 \% (0\%)}}  \\
        & \multicolumn{1}{c}{{0 (0) }} & \multicolumn{1}{c}{{0 (0)}} & \multicolumn{1}{c}{{0 (0) }} & \multicolumn{1}{c}{{0 (0)}} & \multicolumn{1}{c}{{0 (0)}} & \multicolumn{1}{c}{\textcolor{green}{0 (0) }} & \multicolumn{1}{c}{{0 (0)}} \\

     \multirow{2}{*}{Gear Fault}  
        & \multicolumn{1}{c}{\textcolor{red}{5.49 \% (3.29\%)}} & \multicolumn{1}{c}{{0 \% (0\%)}} & \multicolumn{1}{c}{{0 \% (0\%)}} 
        & \multicolumn{1}{c}{\textcolor{red}{0.88 \% (2.19\%) }} & \multicolumn{1}{c}{\textcolor{red}{0.22 \% (0.32\%) }} &
         \multicolumn{1}{c}{\textcolor{red}{0.11 \% (0.11\%)}}
        & \multicolumn{1}{c}{\textcolor{green}{84.63 \% (7.68\%)}}  \\
        & \multicolumn{1}{c}{\textcolor{red}{50 (30) }} & \multicolumn{1}{c}{{0 (0)}} & \multicolumn{1}{c}{{0 (0) }} & \multicolumn{1}{c}{\textcolor{red}{8 (20)}} & \multicolumn{1}{c}{\textcolor{red}{2 (3)}} & \multicolumn{1}{c}{\textcolor{red}{1 (1) }} & \multicolumn{1}{c}{\textcolor{green}{771 (70)}} \\

     \bottomrule    
   \end{tabular}}
\end{table*}
\renewcommand{\baselinestretch}{1.5} 

\renewcommand{\baselinestretch}{1.3} 
\begin{table*}[ht]
   \caption{}
   \caption*{Confusion Matrix - Case 3}
   \label{tab:confusionmatrix_3}
   \resizebox{\textwidth}{!}{\begin{tabular}{cccccccc}
     \toprule
    \textbf{Case 3} & Normal & BPFO & BPFI & Unbalance & Misalignment & Looseness & Gear Fault  \\
     \midrule
     \multirow{2}{*}{Normal}  
        & \multicolumn{1}{c}{\textcolor{green}{24.54 \% (0.17\%)}} & \multicolumn{1}{c}{\textcolor{red}{0.04 \% (0.09\%)}} & \multicolumn{1}{c}{\textcolor{red}{0.02 \% (0.02\%)}} 
        & \multicolumn{1}{c}{\textcolor{red}{0.06 \% (0\%) }} & \multicolumn{1}{c}{\textcolor{red}{0.04 \% (0\%) }} &
         \multicolumn{1}{c}{\textcolor{red}{0.02 \% (0.02\%)}}
        & \multicolumn{1}{c}{\textcolor{red}{0.06 \% (0.09\%)}}  \\
        & \multicolumn{1}{c}{\textcolor{green}{1223 (9) }} & \multicolumn{1}{c}{\textcolor{red}{2 (5)}} & \multicolumn{1}{c}{\textcolor{red}{1 (1) }} & \multicolumn{1}{c}{\textcolor{red}{3 (0)}} & \multicolumn{1}{c}{\textcolor{red}{2 (0)}} & \multicolumn{1}{c}{\textcolor{red}{1 (1) }} & \multicolumn{1}{c}{\textcolor{red}{3 (5)}} \\

     \multirow{2}{*}{BPFO}  
        & \multicolumn{1}{c}{{0 \% (0\%)}} & \multicolumn{1}{c}{\textcolor{green}{0 \% (0\%)}} & \multicolumn{1}{c}{{0 \% (0\%)}} 
        & \multicolumn{1}{c}{{0 \% (0\%) }} & \multicolumn{1}{c}{{0 \% (0\%) }} &
         \multicolumn{1}{c}{{0 \% (0\%)}}
        & \multicolumn{1}{c}{{0 \% (0\%)}}  \\
        & \multicolumn{1}{c}{{0 (0) }} & \multicolumn{1}{c}{\textcolor{green}{0 (0)}} & \multicolumn{1}{c}{{0 (0) }} & \multicolumn{1}{c}{{0 (0)}} & \multicolumn{1}{c}{{0 (0)}} & \multicolumn{1}{c}{{0 (0) }} & \multicolumn{1}{c}{{0 (0)}} \\

     \multirow{2}{*}{BPFI}  
        & \multicolumn{1}{c}{{0 \% (0\%)}} & \multicolumn{1}{c}{{0 \% (0\%)}} & \multicolumn{1}{c}{\textcolor{green}{0 \% (0\%)}} 
        & \multicolumn{1}{c}{{0 \% (0\%) }} & \multicolumn{1}{c}{{0 \% (0\%) }} &
         \multicolumn{1}{c}{{0 \% (0\%)}}
        & \multicolumn{1}{c}{{0 \% (0\%)}}  \\
        & \multicolumn{1}{c}{{0 (0) }} & \multicolumn{1}{c}{{0 (0)}} & \multicolumn{1}{c}{\textcolor{green}{0 (0) }} & \multicolumn{1}{c}{{0 (0)}} & \multicolumn{1}{c}{{0 (0)}} & \multicolumn{1}{c}{{0 (0) }} & \multicolumn{1}{c}{{0 (0)}} \\

     \multirow{2}{*}{Unbalance}  
        & \multicolumn{1}{c}{\textcolor{red}{0.08 \% (0.04\%)}} & \multicolumn{1}{c}{\textcolor{red}{0.02 \% (0.02\%)}} & \multicolumn{1}{c}{{0 \% (0\%)}} 
        & \multicolumn{1}{c}{\textcolor{green}{24.82 \% (0.39\%) }} & \multicolumn{1}{c}{{0 \% (0\%) }} &
         \multicolumn{1}{c}{\textcolor{red}{0.16 \% (0.39\%)}}
        & \multicolumn{1}{c}{{0 \% (0\%)}}  \\
        & \multicolumn{1}{c}{\textcolor{red}{4 (2) }} & \multicolumn{1}{c}{\textcolor{red}{1 (1)}} & \multicolumn{1}{c}{{0 (0) }} & \multicolumn{1}{c}{\textcolor{green}{1247 (20)}} & \multicolumn{1}{c}{{0 (0)}} & \multicolumn{1}{c}{\textcolor{red}{8 (20) }} & \multicolumn{1}{c}{{0 (0)}} \\

     \multirow{2}{*}{Misalignment}  
        & \multicolumn{1}{c}{\textcolor{red}{0.62 \% (0.65\%)}} & \multicolumn{1}{c}{{0 \% (0\%)}} & \multicolumn{1}{c}{{0 \% (0\%)}} 
        & \multicolumn{1}{c}{\textcolor{red}{0.02 \% (0\%) }} & \multicolumn{1}{c}{\textcolor{green}{22.71 \% (2.46\%) }} &
         \multicolumn{1}{c}{\textcolor{red}{1.71 \% (2.40\%)}}
        & \multicolumn{1}{c}{\textcolor{red}{0.02 \% (0\%)}}  \\
        & \multicolumn{1}{c}{\textcolor{red}{31 (33) }} & \multicolumn{1}{c}{{0 (0)}} & \multicolumn{1}{c}{{0 (0) }} & \multicolumn{1}{c}{\textcolor{red}{1 (0)}} & \multicolumn{1}{c}{\textcolor{green}{1141 (124)}} & \multicolumn{1}{c}{\textcolor{red}{86 (121) }} & \multicolumn{1}{c}{\textcolor{red}{1 (0)}} \\

     \multirow{2}{*}{Looseness}  
        & \multicolumn{1}{c}{{0 \% (0\%)}} & \multicolumn{1}{c}{{0 \% (0\%)}} & \multicolumn{1}{c}{{0 \% (0\%)}} 
        & \multicolumn{1}{c}{\textcolor{red}{1.53 \% (0.85\%) }} & \multicolumn{1}{c}{\textcolor{red}{0.10 \% (0.13\%) }} &
         \multicolumn{1}{c}{\textcolor{green}{23.44 \% (0.75\%)}}
        & \multicolumn{1}{c}{{0 \% (0\%)}}  \\
        & \multicolumn{1}{c}{{0 (0) }} & \multicolumn{1}{c}{{0 (0)}} & \multicolumn{1}{c}{{0 (0) }} & \multicolumn{1}{c}{\textcolor{red}{77 (43)}} & \multicolumn{1}{c}{\textcolor{red}{5 (7)}} & \multicolumn{1}{c}{\textcolor{green}{1178 (38) }} & \multicolumn{1}{c}{{0 (0)}} \\

     \multirow{2}{*}{Gear Fault}  
        & \multicolumn{1}{c}{{0 \% (0\%)}} & \multicolumn{1}{c}{{0 \% (0\%)}} & \multicolumn{1}{c}{{0 \% (0\%)}} 
        & \multicolumn{1}{c}{{0 \% (0\%) }} & \multicolumn{1}{c}{{0 \% (0\%) }} &
         \multicolumn{1}{c}{{0 \% (0\%)}}
        & \multicolumn{1}{c}{\textcolor{green}{0 \% (0\%)}}  \\
        & \multicolumn{1}{c}{{0 (0) }} & \multicolumn{1}{c}{{0 (0)}} & \multicolumn{1}{c}{{0 (0) }} & \multicolumn{1}{c}{{0 (0)}} & \multicolumn{1}{c}{{0 (0)}} & \multicolumn{1}{c}{{0 (0) }} & \multicolumn{1}{c}{\textcolor{green}{0 (0)}} \\

     \bottomrule    
   \end{tabular}}
\end{table*}
\renewcommand{\baselinestretch}{1.5} 

Table~\ref{tab:confusionmatrix_3} presents the results obtained for Case 3. The normal signals were confused with all types of faults, despite the low amount. The unbalance signals were confused with normal condition, BPFO and looseness. Misalignment was confused with normal condition, unbalance, looseness and gear fault. Finally, looseness was confused only with unbalance and misalignment. Among the classes, the normal condition was the one with the highest accuracy rate with 99.04\%. The misalignment showed the highest error, resulting in 90.56\%. The unbalance showed 98.97\% and looseness 93.49\%. The results show the greatest difficulty in classifying misalignment and looseness, which was expected, since some harmonics (such as: 2x and 3x) can be highlighted in both condition, making the analysis difficult.

\subsection{Supervised training with real signals only}

A comparative analysis was performed to verify the difference in accuracy and robustness of the model using FaultD-XAI and with supervised training using only real data.

In supervised training, all samples have labels and the signals used for each class refer to real fault conditions. Although in general, it presents excellent results, it is necessary to have samples of all fault conditions and they need to be previously labeled. Assumptions that often make industrial applications unfeasible. The real signals were divided into 70\% training and 30\% testing, and as in the proposed approach, they were executed 10 times due to the randomness of the model. All available signals were used. The accuracy and standard deviation for each test were: 98.1\% (0.53\%), 98.9\% (0.86\%) and 99.8\% (0.11\%), for Case 1, 2 and 3 respectively.

The result obtained for Case 1 it was the same using the two types of training (supervised with real signals and FaultD-XAI). The only difference was lower the standard deviation (0.53\% compared to 1.2\% in the proposed methodology). It is believed that the results were similar due to the lower complexity of the dataset, and consequently ease of identification of the fault. Moreover, the faults have more specific pattern compared to the other cases.  

In Cases 2 and 3, supervised training with real data obtained higher results and lower standard deviation than FaultD-XAI, as expected. In Case 2 the precision was improved from 92.5\% to 98.9\% and the standard deviation decreased from 5.5\% to 0.86\%. In Case 3, the accuracy was 95.5\% with FaultD-XAI becoming 99.8\% using only the real signals, and the standard deviation went from 2.1\% to 0.11\%.

The results confirm the expected superiority of supervised training with only real data. On the other hand, it is worth mentioning that obtaining real and labeled data of all faults can be a complex and time-consuming task in the industrial applications, problems that are overcome by FaultD-XAI.

\subsection{Explainable Artificial Intelligence (XAI)}

Understanding how the AI model arrived at the end result is critical to its application in the industry. The models need a minimum of interpretability, so that users can trust in the predictions.

In addition to allowing us to understand the most relevant parameters used by the model for the final classification, the analysis also allows us to evaluate the learning of the model.

The model's interpretability makes it possible to understand which were the most relevant features used for the prediction, and also to evaluate the model's learning.

\begin{figure}[H]
  \subfloat[Normal]{
	\begin{minipage}[c]{
	   0.5\textwidth}
	   \centering
       \label{fig:xai_rolamento_a}
       \includegraphics[width=1.1\linewidth]{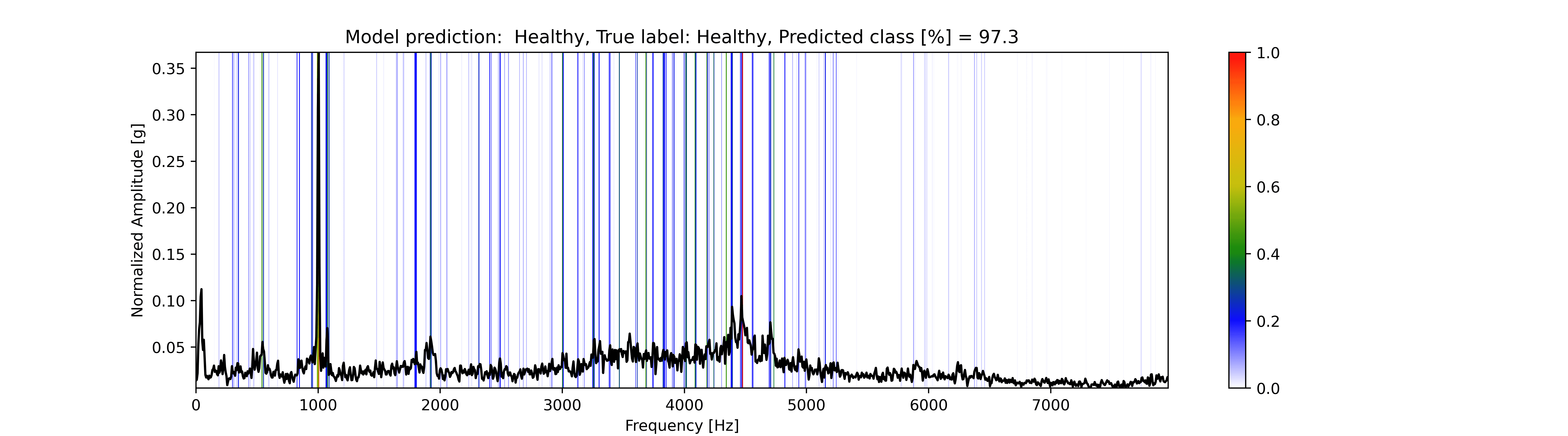}
	\end{minipage}}
 \hfill 
   \subfloat[BPFO]{
	\begin{minipage}[c]{
	   0.5\textwidth}
	   \centering
       \label{fig:xai_rolamento_b}
	   \includegraphics[width=1.1\textwidth]{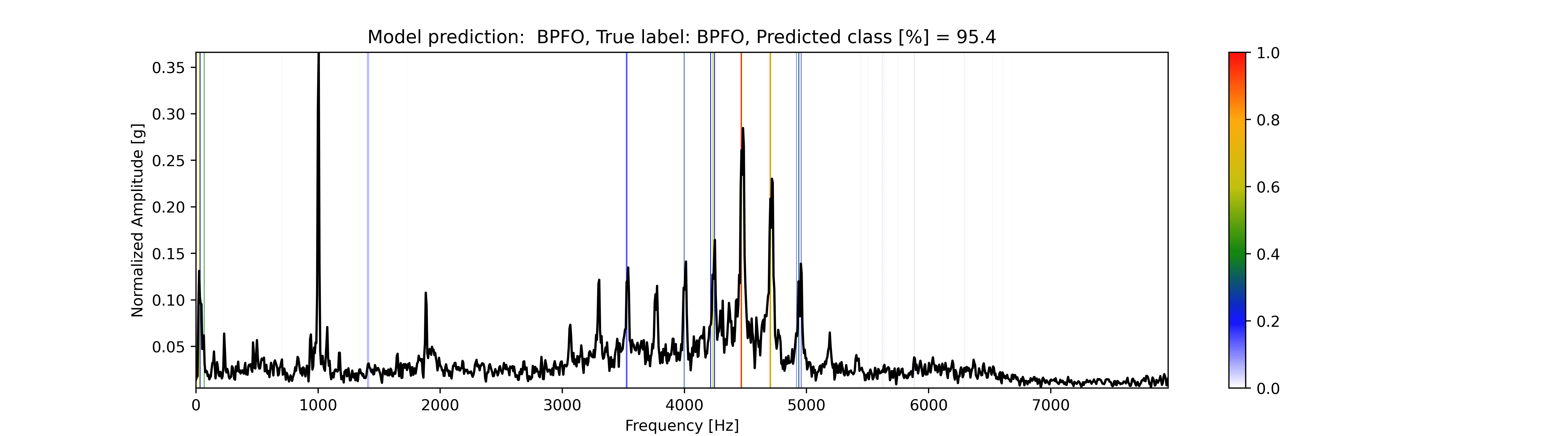}
	\end{minipage}}
    \caption{XAI analysis - Case 1}
    \label{fig:xai_rolamento}	
\end{figure}

The gradients used by the Grad-CAM method after the classification of the sample by the 1D CNN are evaluated. The heat map with the most relevant frequencies for the prediction are presented. A random sample was selected for each type of classification in each case, to exemplify the most relevant features, and validate if they are in accordance with the available literature on faults in vibration analysis of rotating machines. The results obtained for Case 1 are shown in Fig.\ref{fig:xai_rolamento}. 

\renewcommand{\baselinestretch}{1} 
\renewcommand{\baselinestretch}{1.5} 

\begin{figure}[H]
  \subfloat[Normal]{
	\begin{minipage}[c]{
	   0.5\textwidth}
	   \centering
       \label{fig:xai_redutor_a}
       \includegraphics[width=1.1\linewidth]{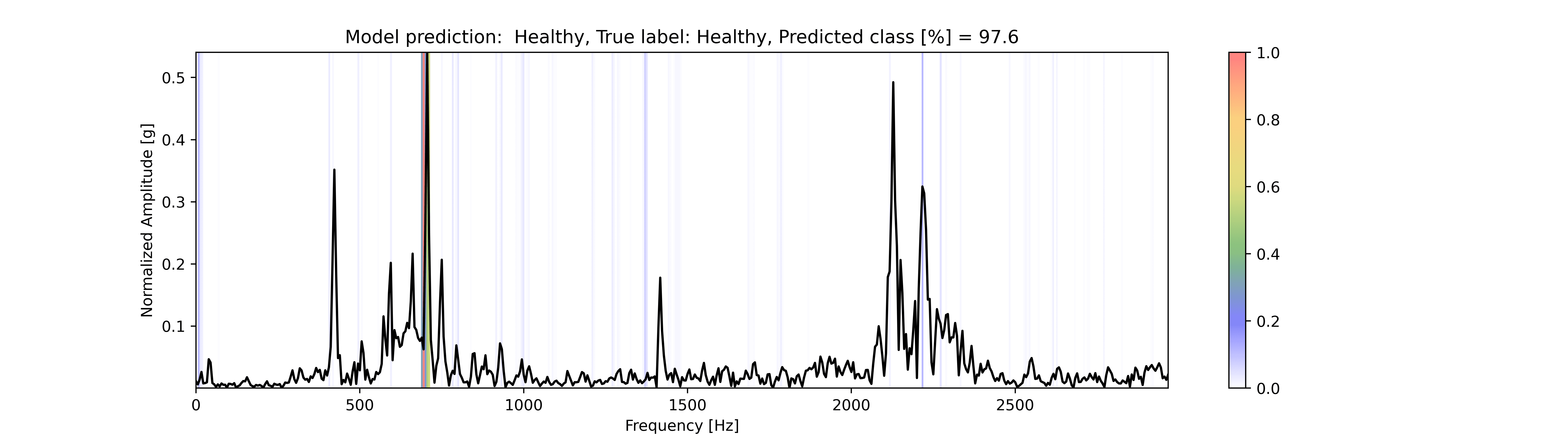}
	\end{minipage}}
 \hfill 
   \subfloat[Gear Fault]{
	\begin{minipage}[c]{
	   0.5\textwidth}
	   \centering
       \label{fig:xai_redutor_b}
	   \includegraphics[width=1.1\textwidth]{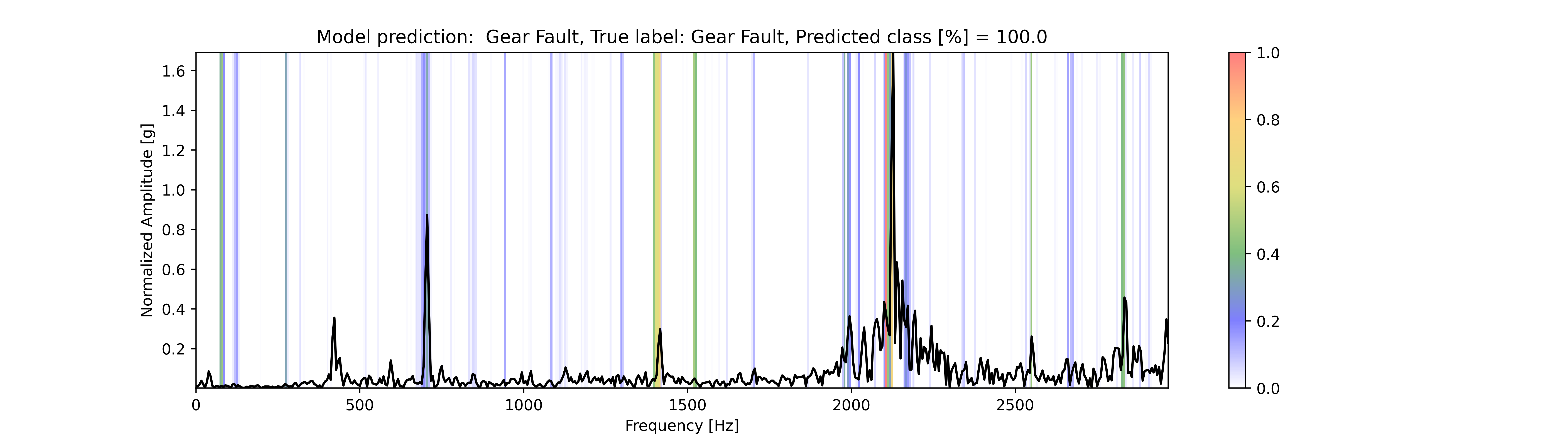}
	\end{minipage}}
    \caption{XAI analysis - Case 2}
    \label{fig:xai_redutor}	
\end{figure}

Analyzing Fig.\ref{fig:xai_rolamento_a}, it is verified that the normal signal uses several frequencies of the signal, not having a specific frequency or region. This was to be expected, as no faults are present, the entire signal referring to a normal machine condition. In the faulty condition, Fig.\ref{fig:xai_rolamento_b}, the high frequency regions are excited, since it is a bearing fault. Consequently, the model uses the frequencies associated with the BPFO as the most relevant for indicating the defect. The analysis corresponds to a manual analysis performed by a rotating machinery specialist. The results obtained for Case 2 are shown in Fig.\ref{fig:xai_redutor}.

\renewcommand{\baselinestretch}{1} 
\renewcommand{\baselinestretch}{1.5}

Case 2, Fig.\ref{fig:xai_redutor_a}, the model uses 1xGMF as a reference for determining the no-fault condition. It is normal to see peaks in GMF, even when the system does not present a fault, as identified by the method. On the other hand, when sidebands start to appear and harmonics appear in greater evidence, it is quite likely that a gear fault is happening. Note that in Fig.\ref{fig:xai_redutor_b}, the method identified the harmonics of the GMF and its sidebands as the most relevant frequencies, again validating the manual analysis of an expert.

\begin{figure}[H]
  \subfloat[Normal]{
	\begin{minipage}[c]{
	   0.5\textwidth}
	   \centering
       \label{fig:xai_bancada_a}
       \includegraphics[width=1.1\linewidth]{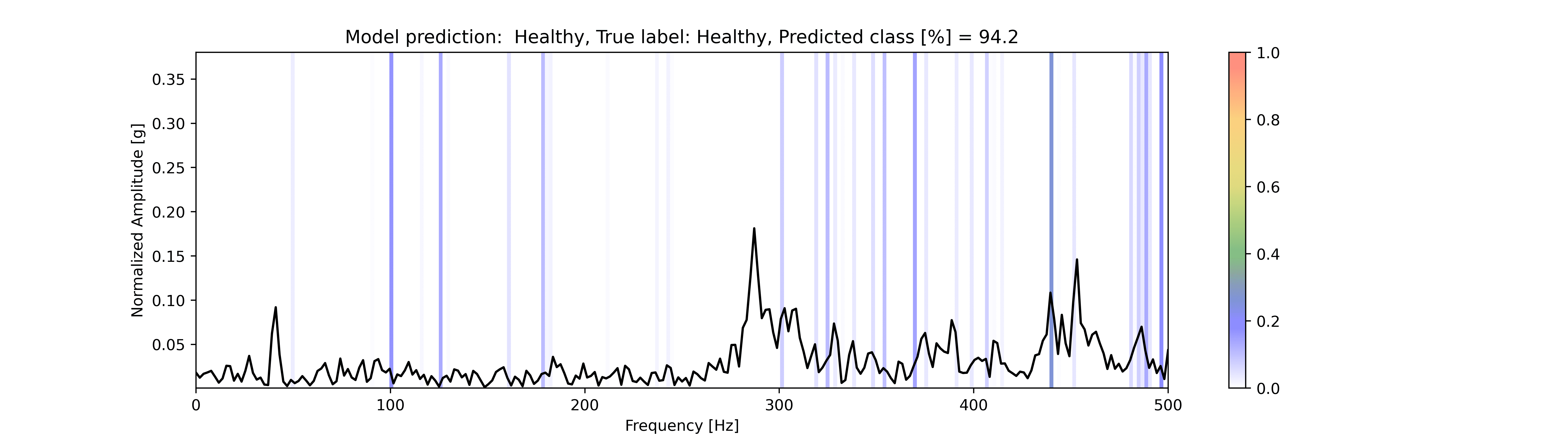}
	\end{minipage}}
 \hfill 
   \subfloat[Unbalance]{
	\begin{minipage}[c]{
	   0.5\textwidth}
	   \centering
       \label{fig:xai_bancada_b}
	   \includegraphics[width=1.1\textwidth]{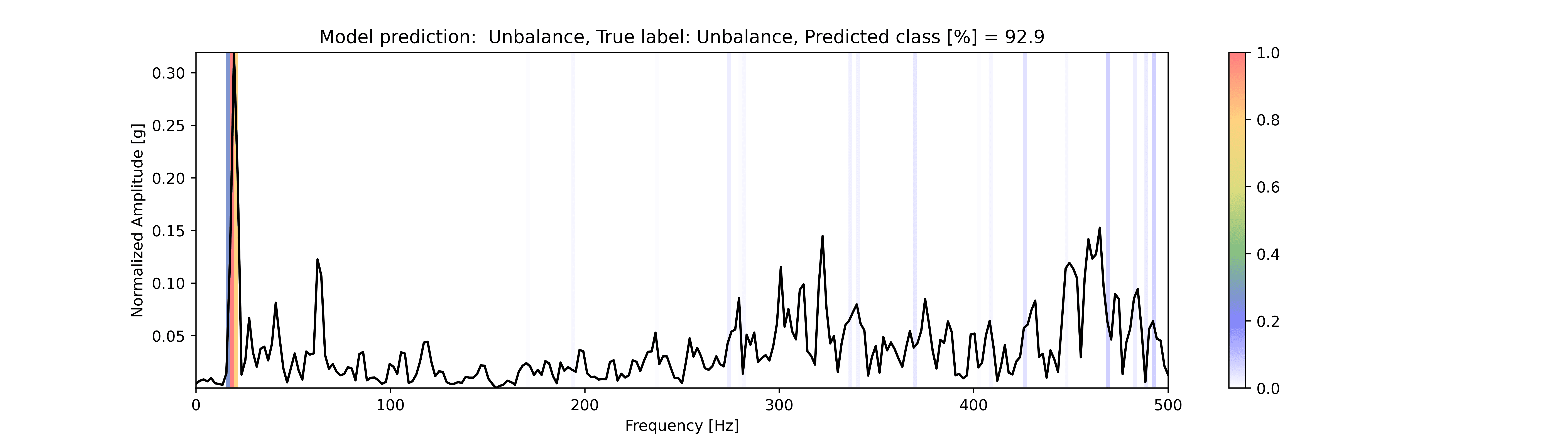}
	\end{minipage}}
  \hfill 
   \subfloat[Misalignment]{
	\begin{minipage}[c]{
	   0.5\textwidth}
	   \centering
       \label{fig:xai_bancada_c}
	   \includegraphics[width=1.1\textwidth]{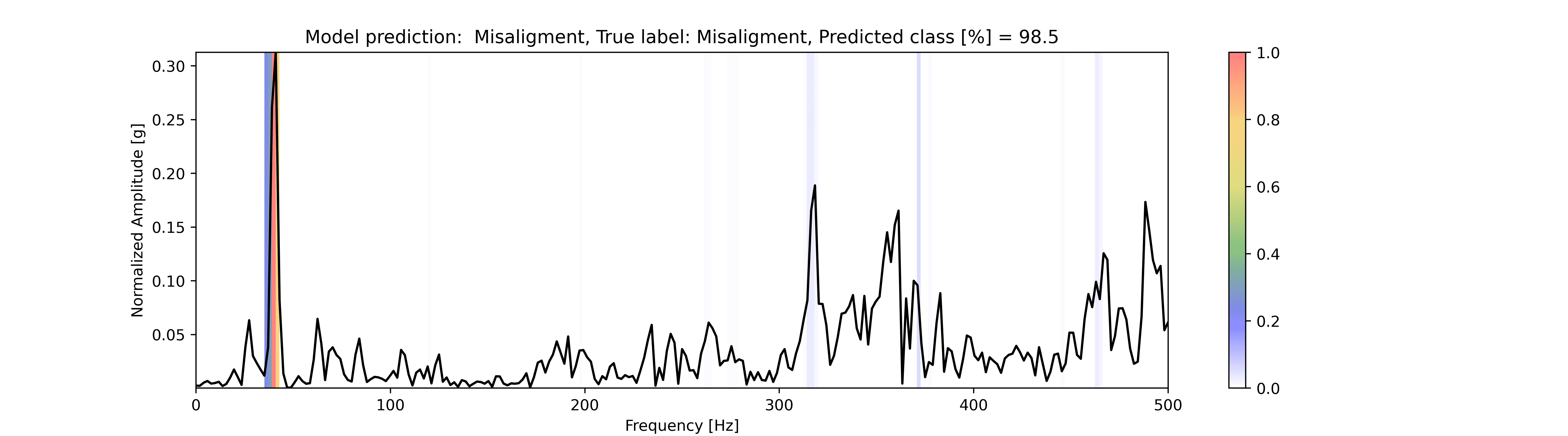}
	\end{minipage}}
  \hfill 
   \subfloat[Looseness]{
	\begin{minipage}[c]{
	   0.5\textwidth}
	   \centering
       \label{fig:xai_bancada_d}
	   \includegraphics[width=1.1\textwidth]{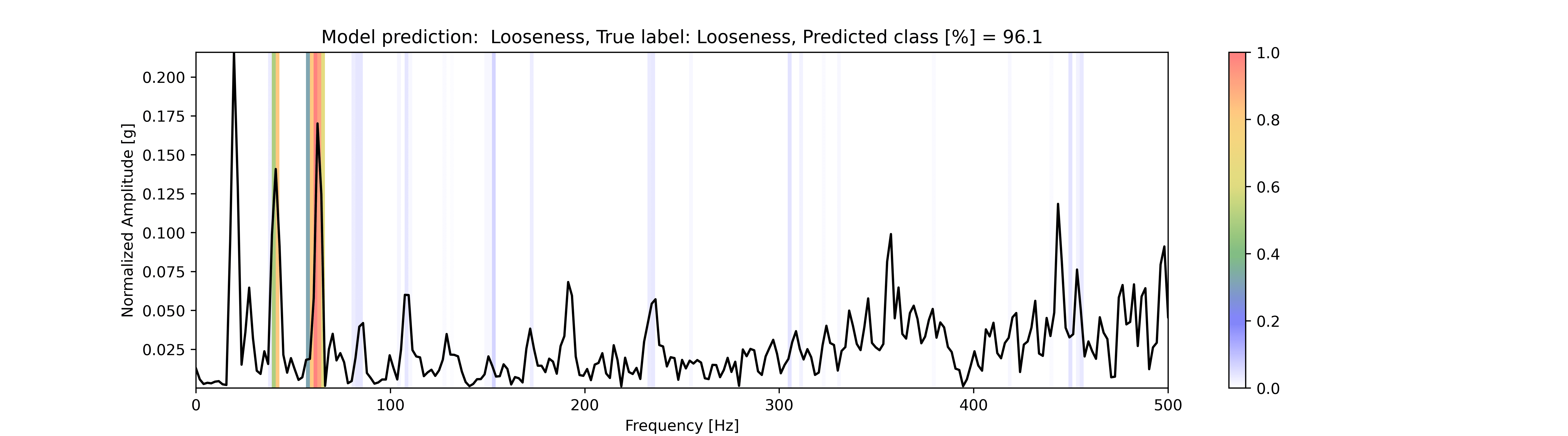}
	\end{minipage}}
    \caption{XAI analysis - Case 3}
    \label{fig:xai_bancada}	
\end{figure}

The results obtained for Case 3 are shown in Fig.\ref{fig:xai_bancada}. The signals are zoomed in the 0 to 500 hz range for easy viewing. For the normal condition, Fig.\ref{fig:xai_bancada_a}, the model did not identify a specific frequency as expected. On the other hand, in Fig.\ref{fig:xai_bancada_b}, there is unbalance, where normally 1 x fr is dominant in the signal. Once again, the model presented the same analysis by an expert, indicating 1 x fr as the most relevant frequency. In Fig.\ref{fig:xai_bancada_c}, the most relevant frequency is 2 x fr, which is exactly one of the misalignment characteristics, where 2 x fr is greater than 1 x fr, and there may or may not be harmonics of 3x, 4x, 5x etc. The mechanical looseness can present itself in some ways, being one of them the excitation of the fr harmonics. In Fig.\ref{fig:xai_bancada_d} it is noted that the model used the rotation harmonics as a reference, with emphasis on 3 x fr, again confirming the analysis based on human knowledge.

\renewcommand{\baselinestretch}{1} 
\renewcommand{\baselinestretch}{1.5} 

\section{Conclusions}

This paper presents a new generic and interpretable approach to classifying faults in rotating machinery based on transfer learning from augmented synthetic data to real rotating machinery. The six-stage scheme is adopted: i) Data Acquisition; ii) Signal Generation; iii) Data Augmentation; iv) Signal Processing; v) Fault Diagnosis; vi) Explainable Artificial Intelligence (XAI). From the acquired vibration signals, the synthetic faults are generated and augmented. Subsequently, they are transformed to the frequency domain, and used to train the AI model. After training, the model is able to classify real fault signals, without them having been used in training. Finally, explainability allows analyzing the predictions obtained. 

To the best of our knowledge, the proposed FaultD-XAI is the first work that combined synthetic data with data augmentation to avoid the need of real data by exploiting XAI and transfer learning for fault diagnosis in rotating machinery. The main contributions of FaultD-XAI are: i) a new transfer learning classification approach based on a synthetic dataset, without the need to have signals of real fault conditions; ii) possibility of interpreting the way in which the final result was obtained by the model, supporting decision making (a new contribution to the study of XAI in fault diagnosis); iii) a generic and simple way to generate synthetic fault data for training, based on the knowledge available in vibration analysis, without the need for complex models; iv) possibility of generating varied training datasets of different sizes (data augmentation - targeting deep learning applications); v) new dataset, publicly available, to study failures such as: unbalance, misalignment and looseness; vi) possibility of application in different types of faults; vii) faster deployment of the model in production; viii) industrial application on real world datasets.

The results show that it is possible to train a model for fault diagnosis in rotating machinery, without having labeled data for all faults, overcoming this major problem in industrial applications. In addition, FaultD-XAI makes it possible to identify the most relevant features used by the model to perform the prediction, overcoming another major problem, which is to trust the result obtained by black-box models.

In summary, the best results (accuracy and standard deviation) reached in Case 1, 2 and 3 were: 98.1\% (1.2\%), 92.5\% (5.5\%) and 95.5\% (2.1\%), respectively. Using the traditional supervised training method and real data, the results obtained were: 98.1\% (0.53\%), 98.9\% (0.86\%) and 99.8\% (0.11\%), for Case 1, 2 and 3 respectively. Therefore, comparing both results, it can be concluded that FaultD-XAI presents a promissory proposal for fault diagnosis. Moreover, the model's explainability analysis shows that it was able to learn the most relevant characteristics of each condition, and that FaultD-XAI is capable of supporting the expert's final decision-making.

As FaultD-XAI does not require labeled data of all possible machine operating conditions, and only knowledge currently available on fault diagnosis through vibration analysis, the methodology has many possible engineering applications. Future work will explore FaultD-XAI to remaining useful life estimation of rotating machinery components \cite{susto2014machine}, using synthetic data, transfer learning and XAI.

\section*{Acknowledgement}
\addcontentsline{toc}{section}{Acknowledgement}

The authors gratefully acknowledge the Brazilian research funding agencies CNPq (National Council for Scientific and Technological Development) and CAPES (Federal Agency for the Support and Improvement of Higher Education) for their financial support of this work.

\addcontentsline{toc}{section}{References}
\bibliography{mybibfile}

\end{document}